\documentclass[twoside,11pt]{article}

\usepackage[accepted,specialissue]{melba}
\usepackage{amsmath,amsfonts}
\usepackage{boldline}
\usepackage{multirow}
\usepackage{tabularx}

\melbaid{2024:020}  
\doi{https://doi.org/10.59275/j.melba.2024-g93a}
\melbaauthors{Woodland et al.}  
\volume{2}
\firstpageno{2006}  
\melbayear{2024}  
\datesubmitted{12/2023}  
\datepublished{10/2024}  

\melbaspecialissue{UNSURE2023 special issue}
\melbaspecialissueeditors{Christian Baumgartner, Adrian Dalca, Raghav Mehta, Chen Qin, Carole Sudre, William (Sandy) Wells}

\ShortHeadings{Dimensionality Reduction and Nearest Neighbors for Improved OOD Detection}{Woodland et al.}

\title{Dimensionality Reduction and Nearest Neighbors for Improving Out-of-Distribution Detection in Medical Image Segmentation}

\author{\name McKell Woodland\orcid{0000-0003-0995-3695}\email mewoodland@mdanderson.org \\
        \addr The University of Texas MD Anderson Cancer Center, Houston, TX, USA \\
        \addr Rice University, Houston, TX, USA
        \vspace{-.6em}
        \AND \name Nihil Patel\orcid{0000-0002-5382-1983}
        \vspace{-.5em}
        \AND \name Austin Castelo\orcid{0009-0009-3015-3570}
        \vspace{-.5em}
        \AND \name Mais Al Taie\orcid{0009-0000-9298-6644}
        \vspace{-.5em}
        \AND \name Mohamed Eltaher\orcid{0009-0008-2078-4718}
        \vspace{-.5em}
        \AND \name Joshua P. Yung\orcid{0000-0003-3752-5997}
        \vspace{-.5em}
        \AND \name Tucker J. Netherton\orcid{0000-0003-1583-7121}
        \vspace{-.5em}
        \AND \name Tiffany L. Calderone\orcid{0000-0003-4404-5342}
        \vspace{-.3em}
        \AND \name Jessica I. Sanchez\orcid{0009-0006-1042-5661}\\
        \addr The University of Texas MD Anderson Cancer Center, Houston, TX, USA
        \vspace{-.6em}
        \AND \name Darrel W. Cleere
        \vspace{-.5em}
        \AND \name Ahmed Elsaiey
        \vspace{-.5em}
        \AND \name Nakul Gupta\orcid{0000-0002-2413-2193}
        \vspace{-.3em}
        \AND \name David Victor\orcid{0000-0003-1414-3128}\\
        \addr Houston Methodist Hospital, Houston, TX, USA
        \vspace{-.3em}
        \AND \name Laura Beretta\orcid{0000-0002-2054-684X}\\
        \addr The University of Texas MD Anderson Cancer Center, Houston, TX, USA
        \vspace{-.3em}
        \AND \name Ankit B. Patel\orcid{0000-0001-9678-496X}\\
        \addr Baylor College of Medicine, Houston, TX, USA \\
        \addr Rice University, Houston, TX, USA
        \vspace{-.3em}
        \AND \name Kristy K. Brock\orcid{0000-0001-9364-5040} \email kkbrock@mdanderson.org \\
        \addr The University of Texas MD Anderson Cancer Center, Houston, TX, USA
}

\begin{document}

\maketitle

\vspace{-3em}

\begin{abstract}%
Clinically deployed deep learning-based segmentation models are known to fail on data outside of their training distributions.
While clinicians review the segmentations, these models tend to perform well in most instances, which could exacerbate automation bias.
Therefore, detecting out-of-distribution images at inference is critical to warn the clinicians that the model likely failed. 
This work applied the Mahalanobis distance (MD) post hoc to the bottleneck features of four Swin UNETR and nnU-net models that segmented the liver on T1-weighted magnetic resonance imaging and computed tomography.
By reducing the dimensions of the bottleneck features with either principal component analysis or uniform manifold approximation and projection, images the models failed on were detected with high performance and minimal computational load.
In addition, this work explored a non-parametric alternative to the MD, a k-th nearest neighbors distance (KNN).
KNN drastically improved scalability and performance over MD when both were applied to raw and average-pooled bottleneck features.
Our code is available at \url{https://github.com/mckellwoodland/dimen\_reduce\_mahal}.
\end{abstract}


\section{Introduction}

Liver malignancy is one of the leading causes of cancer death worldwide \citep{Ferlay2021}, with mortality rates increasing more rapidly than all other cancers within the United States \citep{Ryerson2016}.
Radiotherapy is a non-invasive treatment for advanced liver cancer that leverages ionizing radiation to treat tumors \citep{Wenqi2020}.
Precise delineation of treatment targets and surrounding anatomical structures is critical to the success of radiotherapy.
Manual segmentation of these structures is time-intensive \citep{multi2011}, leading to delays that have been correlated with lower survival rates \citep{Chen2008} and incompatibility with techniques that require frequent imaging to account for anatomical changes \citep{Sheng2020}, such as magnetic resonance imaging (MRI)-guided adaptive radiotherapy \citep{Otazo2021}.
In addition, manual segmentation is subject to human variability and inconsistencies \citep{Nelms2012}, which can lead to a lower quality of radiotherapy \citep{Saarnak2000}.
These limitations have prompted expansive research into automated segmentation methods.

Deep learning (DL) algorithms constitute the current state-of-the-art for medical imaging segmentation, with research spanning many anatomical regions and imaging modalities \citep{Cardenas2019}.
However, DL models struggle to generalize to information that was not present while the model was being trained \citep{Zech2018}.
This problem is exacerbated in the medical field, where collecting large-scale, annotated, and diverse training datasets is challenging due to the cost of labeling, rare cases, and patient privacy.
Even models with high performance during external validation may fail when presented with novel information after clinical deployment.
This can be demonstrated by the work of \cite{Anderson2021}.
On test data, 96\% of their DL-based liver segmentations were deemed clinically acceptable, with most of their segmentations being preferred over manual segmentations. 
The two images the model performed poorly on contained information not present during training – namely, the presence of ascites and a stent.

Automated segmentations are typically manually evaluated and corrected, if need be, by a clinician before they are used in patient treatment.
The main concern with human evaluation is automation bias, where physicians may become too reliant on model output. 
To protect against automation bias, it is critical to warn clinicians of potential segmentation model failure.
Identifying model inputs that will lead to poor model performance is referred to as out-of-distribution (OOD) detection \citep{Yang2024}.
This study focuses on post-hoc OOD detection or methods that can be applied after model training in order to develop warning systems for models already in clinical deployment.

Mahalanobis distance (MD) is a commonly used post-hoc OOD detection method that computes the distance between a test image and a Gaussian distribution fitted to training images \citep{Lee2018}.
Given the inherent high dimensionality of images, the distance is typically applied to features extracted from the network being analyzed.
MD has achieved state-of-the-art performance in natural imaging when applied directly to classifier features \citep{Fort2021}.
However, features from medical imaging segmentation models are an order of magnitude larger than classifier features, necessitating further dimensionality reduction to ensure computational feasibility.
While average pooling has been used conventionally to reduce feature dimensionality \citep{Lee2018,Gonzalez2021}, no prior studies have examined the best way to prepare features for the MD calculation.
Other open areas of research in regards to the application of MD to medical imaging segmentation models include the validity of the Gaussian assumption, which features from the model should be utilized \citep{GONZALEZ2022}, how to combine features best if multiple features are utilized, and how to utilize MD with multi-class segmentation networks.

We aim to improve the performance and scalability of feature-based OOD detection in medical imaging segmentation.
Our main contributions are two-fold.
First, we propose using principal component analysis (PCA) and uniform manifold approximation and projection (UMAP) to prepare features for the MD calculation.
We demonstrate that these methods outperform average pooling across four liver segmentation models.
Second, we propose using a k-th nearest neighbor (KNN) distance \citep{Sun2021} as a distribution-agnostic replacement for MD for medical imaging segmentation models.
Our results show a drastic improvement of KNN over MD on raw and average pooled features, questioning the validity of the Gaussian assumption for segmentation model features.

This work was first published in Lecture Notes of Computer Science volume 14291, pages 147–156 by Springer Nature \citep{Woodland2023}.
It was extended to include validation of the dimensionality reduction techniques for three additional liver segmentation models (including extensions to computed tomography and the nnU-net architecture).
Furthermore, the extension includes a novel analysis of the KNN distance as a replacement for MD. 
Finally, the extension provides greater context into how MD and KNN fit into the larger OOD detection field by comparing their performance to standard methods.

\section{Related Works}

Traditional OOD detection aims to identify and reject model input whose true label deviates semantically from the label distribution observed during the model's training phase \citep{Yang2024}.
Our work follows an alternative definition of OOD detection common in many safety-critical applications: identifying and rejecting model input that falls outside the model's generalization capacity \citep{pleiss2019neural,Yang2024}.
Distribution in this context refers to a theoretical statistical distribution where data drawn from it is within the scope of the model under consideration.
A comprehensive review of OOD detection approaches was recently compiled by \cite{Yang2024}.
Our review focuses on feature- and output-based OOD detection methods that can be applied post hoc to segmentation models and training-based uncertainty estimation approaches.

\cite{Lee2018} utilized density estimation for OOD detection by calculating the MD between test features and class-conditional Gaussian distributions fit to training features.
\cite{Fort2021} achieved state-of-the-art OOD detection performance on standard vision benchmarks by applying the MD to features extracted from large-scale, pre-trained vision transformers.
\cite{Gonzalez2021} applied the MD to medical imaging segmentation architectures by fitting a Gaussian distribution to the bottleneck features of an nnU-Net architecture \citep{Isensee2020}.
\cite{Sun2022} replaced MD with a k-th nearest neighbor distance to relax the Gaussian assumption on the feature space.
\cite{Ghosal2024} proposed Subspace Nearest Neighbor (SNN), a k-th nearest neighbor approach that reduces the dimensionality of the calculation by masking out irrelevant features.
\cite{Karimi2023} found that the Euclidean distance between the spectral features of a test image and its nearest neighbor in the training dataset achieved the best OOD detection performance on medical imaging segmentation tasks.
\cite{Sastry2020} measured the deviation of test images from training images by applying high-order Gram matrices to all neural network layers.
Our work builds upon past research by improving the performance of MD through dimensionality reduction and comparing MD to KNN in a medical imaging feature space.

\cite{Hendrycks2017} proposed Maximum Softmax Probability (MSP) as an OOD detection baseline, where OOD samples were identified by their prediction probabilities.
The intuition behind this approach is that models should express more confidence on ID samples than on OOD samples.
However, in practice, neural networks tend to express high confidence, even on OOD samples \citep{Nguyen2015}.
\cite{Guo2017} calibrated model confidence, or aligned prediction probabilities with true correctness likelihoods, by scaling model logits with a single parameter (temperature scaling).
\cite{Liang2018} utilized temperature scaling and small input perturbations to improve the performance of MSP (ODIN).
\cite{Liu2020} furthered performance by replacing the softmax function with an energy function.
\cite{Sun2021} outperformed previous methods by truncating hidden activations (ReAct).
In this work, we compare MD and KNN to MSP, temperature scaling, and energy scoring.

OOD detection can also be performed using prediction uncertainties.
While Bayesian neural networks provide probability distributions over weights, they are intractable for uncertainty estimation as they require significant modifications to neural network architectures and are computationally prohibitive.
MC Dropout, which computes prediction uncertainties by combining multiple stochastic passes through a network at inference, was proposed by \cite{Gal2016} as a Bayesian approximation to Gaussian processes.
\cite{Gal207} further introduced concrete dropout, which improved uncertainty calibration by allowing for automated tuning of the dropout probability.
\cite{Teye2018} approximated Bayesian inference with batch normalization.
\cite{Lakshminarayanan2017} took a frequentist approach by ensembling neural networks to improve predictive uncertainty.
While ensembling is computationally expensive, it performs well in medical imaging segmentation literature \citep{Jungo2020,Mehrtash2020,Adams2023}.
To reduce the computational complexity of ensembling, \cite{Wen2020} proposed BatchEnsemble, which enables weight sharing.
In this work, we compare MD and KNN to MC Dropout and ensembling.

\section{Methods}

\subsection{Segmentation}

We used six liver segmentation models (Table \ref{tab:models}) that were based on six datasets (Table \ref{tab:data}) for OOD detection analysis.
The first model was the Swin UNETR \citep{Tang2022} from \cite{Woodland2023} (hereafter called MRI UNETR).
This model was trained on 337 T1-weighted liver MRI exams from the Duke Liver (DLDS) \citep{Macdonald2020,MacDonald2023}, Abdominal Multi-Organ Segmentation (AMOS) \citep{Ji2022data,Ji2022}, and Combined Healthy Abdominal Organ Segmentation (CHAOS) \citep{Kavur2019data,kavur2019,Kavur2021} datasets (collectively called MRI$_{\text{Tr}}$) and tested on 27 T1-weighted liver MRIs from The University of Texas MD Anderson Cancer Center (called MRI$_{\text{Te}}$).
All MD Anderson images were retrospectively acquired under an internal review board (IRB)-approved protocol. 

\begin{table}[h]
    \centering
    \caption{Description of the segmentation models.}
    \label{tab:models}
    \begin{tabular}{|l|c|c|l|}
    \cline{1-4}
    \textbf{Name} & \textbf{Train} & \textbf{Test} & \parbox{8cm}{\centering\textbf{Description}} \\
    \cline{1-4}
    MRI UNETR & MRI$_{\text{Tr}}$ & MRI$_{\text{Te}}$ & UNETR from \cite{Woodland2023} \\
    \cline{1-4}
    \multirow{2}{*}{MRI Dropout} & \multirow{2}{*}{MRI$_{\text{Tr}}$} & \multirow{2}{*}{MRI$_{\text{Te}}$} & \multirow{2}{*}{\parbox{8cm}{UNETR with dropout enabled and predictions combined}} \\
    &&&\\
    \cline{1-4}
    MRI Ensemble & MRI$_{\text{Tr}}$ & MRI$_{\text{Te}}$ & Ensemble of 5 UNETRs \\
    \cline{1-4}
    MRI+ UNETR & MRI$+_{\text{Tr}}$ & MRI$+_{\text{Te}}$ & UNETR from \cite{Patel2024} \\
    \cline{1-4}
    MRI+ nnU-net & MRI$+_{\text{Tr}}$ & MRI$+_{\text{Te}}$ & nnU-net from \cite{Patel2024} \\
    \cline{1-4}
    CT nnU-net & CT$_{\text{Tr}}$ & CT$_{\text{Te}}$ & nnU-net from MD Anderson \\
    \cline{1-4}
    \end{tabular}
\end{table}

The next two models were created to enable comparison of MC Dropout and ensembling to MD and KNN.
The second model, called MRI Dropout, was a Swin UNETR trained on MRI$_{\text{Tr}}$ with a 20\% dropout rate and tested on MRI$_{\text{Te}}$.
Enabling dropout added dropout layers into many architecture components, including after the positional embedding, the window-based multi-head self-attention module, and the multi-layer perceptron.
The segmentation map for this model was generated by averaging the predictions obtained from enabling dropout during inference and conducting five forward passes per image.
The third model, called MRI Ensemble, was an ensemble of five Swin UNETRs trained on MRI$_{\text{Tr}}$. 
The predictions from each of these five models were averaged to generate a final segmentation map.

\begin{table}[h]
    \centering
    \caption{Description of the datasets.}
    \label{tab:data}
    \begin{tabular}{|l|l|}
        \cline{1-2}
        \textbf{Name} & \parbox{13cm}{\centering \textbf{Description}} \\
        \hline
         MRI$_{\text{Tr}}$ &  337 T1-weighted abdominal MRIs from DLDS, AMOS, and CHAOS datasets \\
         \hline
         MRI$_{\text{Te}}$ & 27 T1-weighted abdominal MRIs from MD Anderson\\
         \hline
         \multirow{2}{*}{MRI+$_{\text{Tr}}$} & \multirow{2}{*}{\parbox{13cm}{371 T1-weighted abdominal MRIs from MD Anderson and curated DLDS, AMOS, CHAOS, ATLAS datasets}} \\
         & \\
         \hline
         MRI+$_{\text{Te}}$ & 352 T1-weighted abdominal MRIs from Houston Methodist \\
         \hline
         CT$_{\text{Tr}}$ & 2,840 abdominal CTs from MD Anderson \\
         \hline
         CT$_{\text{Te}}$ & 248 abdominal CTs from MD Anderson and BTCV challenge \\
         \hline
    \end{tabular}
\end{table}

The encoders of MRI UNETR, MRI Dropout, and MRI Ensemble models were pre-trained using self-distilled masked imaging (SMIT) \citep{Jiang2022} with 3,610 unlabeled head and neck computed tomography scans (CTs) from the Beyond the Cranial Vault (BTCV) Segmentation Challenge dataset \citep{Landman2015}. 
The official Swin UNETR codebase, built on top of the Medical Open Network for AI (MONAI) \citep{Monai2021}, was utilized for the pre-trained weights and training.
Models were trained with default parameters for 1,000 epochs with the default batch size of 1.
Each model was trained on a single node of a Kubernetes cluster containing eight A100 graphic processing units (GPUs) with 40 gigabytes (GB) of memory. 
A total of 100 GB of memory was requested from the cluster.
Final model weights were selected according to the weights with the highest validation Dice similarity coefficient (DSC).

The rest of the liver segmentation models were previously trained at MD Anderson and were utilized for post-hoc OOD detection analysis.
The fourth model, called MRI+ UNETR, builds upon MRI UNETR by expanding and curating the training and testing datasets.
MRI+ UNETR was trained on 48 scans from the AMOS dataset, 172 scans from the DLDS dataset, 38 scans from the CHAOS dataset, 44 scans from the Tumor and Liver Automatic Segmentation (ATLAS) dataset \citep{Quinton2023}, and 69 scans from MD Anderson, for a total of 371 T1-weighted liver MRIs (collectively named MRI+$_{\text{Tr}}$).
352 scans from 71 patients with hepatocellular carcinoma collected from Houston Methodist Hospital (called MRI+$_{\text{Te}}$) were used for evaluation.
Inclusion criteria for MRI+$_{\text{Tr}}$ and MRI+$_{\text{Te}}$ necessitated the entire liver to be visible, no prior liver surgery, and sufficient image quality to ensure the boundary of the liver was identifiable without pre-existing contours.
The fifth model, named MRI+ nnU-net, was an nnU-net trained and tested on MRI$_{\text{Tr}}$ 
 and MRI+$_{\text{Te}}$ that was included to enable a comparison between the nnU-net and Swin UNETR architectures.
For more information on the MRI+ models, please refer to \cite{Patel2024}.

The final model, named CT nnU-net, was an nnU-net trained on 2,840 internally obtained abdominal computed tomography (CT) scans (CT$_{\text{Tr}}$) and tested on 248 CT scans (CT$_{\text{Te}}$).
It was included to expand our analysis to computed tomography.
The training scans varied in the presence of and phase of contrast (portal-venous and arterial phases), states of liver disease and histology, presence of artifacts (including ablation needles, stents, and post-resection clips), and therapy stage (planning, intra-operative, and post-operative).
30 of the test scans came from the BTCV challenge \citep{Landman2015}, while the rest were acquired internally from MD Anderson.
Images from an ongoing liver ablation clinical trial\footnote{\url{https://www.clinicaltrials.gov/study/NCT04083378}} and \cite{Anderson2021} were used in both the training and testing phases of the segmentation model.

Segmentation performance was evaluated with DSC, maximum Hausdorff distance (HD), and Normalized Surface Dice (NSD) with a threshold of 2 millimeters.
One-sided paired \emph{t}-tests were conducted with a significance level of $\alpha$=.05 to determine the significance of performance improvements.

\subsection{OOD Detection}

To evaluate the detection of images that a segmentation model will perform poorly on, each model’s test data was split into in-distribution (ID) and OOD categories based on that specific model's performance.
An image was labeled ID for a model if the image had an associated DSC of at least 95\%.
Accordingly, an image was labeled OOD if it had a DSC under 95\%.
If there were not at least two ID images, the threshold was lowered to 80\%.
While we consider 80\% DSC to be acceptable for clinical deployment, we prefer a 95\% DSC as these contours are unlikely to require any editing in the clinical process.
In practice, the threshold should be determined by the individual use case.
For robustness, experiments were computed for 95\%, 80\%, and median value thresholds.
Furthermore, DSCs were plotted against OOD scores to visually demonstrate how the results would change if the threshold changed.

Performance was measured with the area under the receiver operating characteristic curve (AUROC), the area under the precision-recall curve (AUPRC), the false positive rate at a 90\% true positive rate (FPR90), and the amount of time in seconds it took to compute the OOD scores, with OOD considered the positive class.
Averages and standard deviations (SDs) were reported across five OOD score calculations, with each score being calculated on the entire test dataset with a different NumPy random seed.
\emph{t}-tests were performed with a significance level $\alpha=0.10$ to determine the significance of configuration improvements.
All calculations were performed on a node of a Kubernetes cluster with 16 central processing units (CPUs) and a requested 256 megabytes of memory, with a 256 GB limit.

\subsubsection{Mahalanobis distance}

The Mahalanobis distance $D$ measures the distance between a point $x$ and a distribution with mean $\mu$ and covariance matrix $\Sigma$, $D^{2}=(x-\mu)^{T} \Sigma^{-1} (x-\mu)$ \citep{Mahalanobis1936}. 
\cite{Lee2018} first proposed using the MD for OOD detection to calculate the distance between test images embedded by a classifier and a Gaussian distribution fit to class-conditional embeddings of the training images. 
Similarly, \cite{Gonzalez2021} used the MD for OOD detection in segmentation networks by extracting embeddings from the encoder of an nnU-Net. 
As distances in high dimensions are subject to the curse of dimensionality, both sets of authors decreased the dimensionality of the embeddings through average pooling.
\cite{Lee2018} suggested pooling the features such that the height and width dimensions of the features become singular.
\cite{Gonzalez2021} pooled features with a kernel size and stride of 2 until the feature dimensionality fell below 10,000.

The MD was calculated on features extracted from the bottlenecks of the liver segmentation models.
For the UNETRs, the projected features \texttt{x} were saved from the \texttt{Trans\_Unetr} class.
For the nnU-nets, the features \texttt{skips} were saved from the \texttt{PlainConvUNet} class for each sliding window and subsequently concatenated.
As the nnU-nets automatically cropped their inputs, the size of the bottleneck feature dimension representing the number of concatenated sliding windows could not be standardized.
To account for this, average pooling was applied across the concatenated embeddings such that the size of the dimension representing the number of concatenated windows was made singular.
The size of the bottleneck features for the UNETR models was standardized by resizing model inputs to (256, 128, 128) prior to feature extraction.
After extraction, all features were flattened to prepare them for distance calculations.
Note that nnU-net uses instance normalization in the encoding process, whereas UNETR uses layer normalization.

Gaussian distributions were fitted on the raw training embeddings.
For each test embedding, the MD between the embedding and the corresponding Gaussian distribution was calculated and used as the OOD score.
Covariance matrices were estimated empirically with maximum likelihood estimation.

\subsubsection{K-Nearest Neighbors}

When \cite{Lee2018} introduced MD for OOD detection, the authors showed that the posterior distribution of a softmax classifier can be modeled by a generative classifier defined by Gaussian Discriminant Analysis, thereby demonstrating that classifier embeddings can be Gaussian-distributed.
However, this does not guarantee the embeddings are Gaussian-distributed, as demonstrated by classifier embeddings failing normality tests \citep{Sun2022}.
Additionally, this analysis has not been extended to segmentation networks.

\cite{Sun2022} first proposed using a k-th nearest neighbor distance as a non-parametric alternative to the MD.
In their work, KNN improved overall performance over MD across five OOD detection benchmark datasets, though performance improvements were dataset-dependent.
In this work, we propose KNN for medical imaging segmentation networks.
We define the k-th nearest neighbor distance to be the Euclidean distance between a test embedding $f(x)$ and its k-th nearest training embedding $f(z_k)$ where $f$ is a trained U-Net encoder $${||f(x)-f(z_k)||}_2.$$
This k-th nearest neighbor distance serves as the OOD score.
A hyperparameter search was performed over $k$  such that $k\in\{2,4,8,16,32,64,128,256\}$.
KNN was performed on the features that were extracted for MD.

\subsubsection{Dimensionality Reduction}

\begin{figure}
    \centering
    \includegraphics[width=0.6\textwidth]{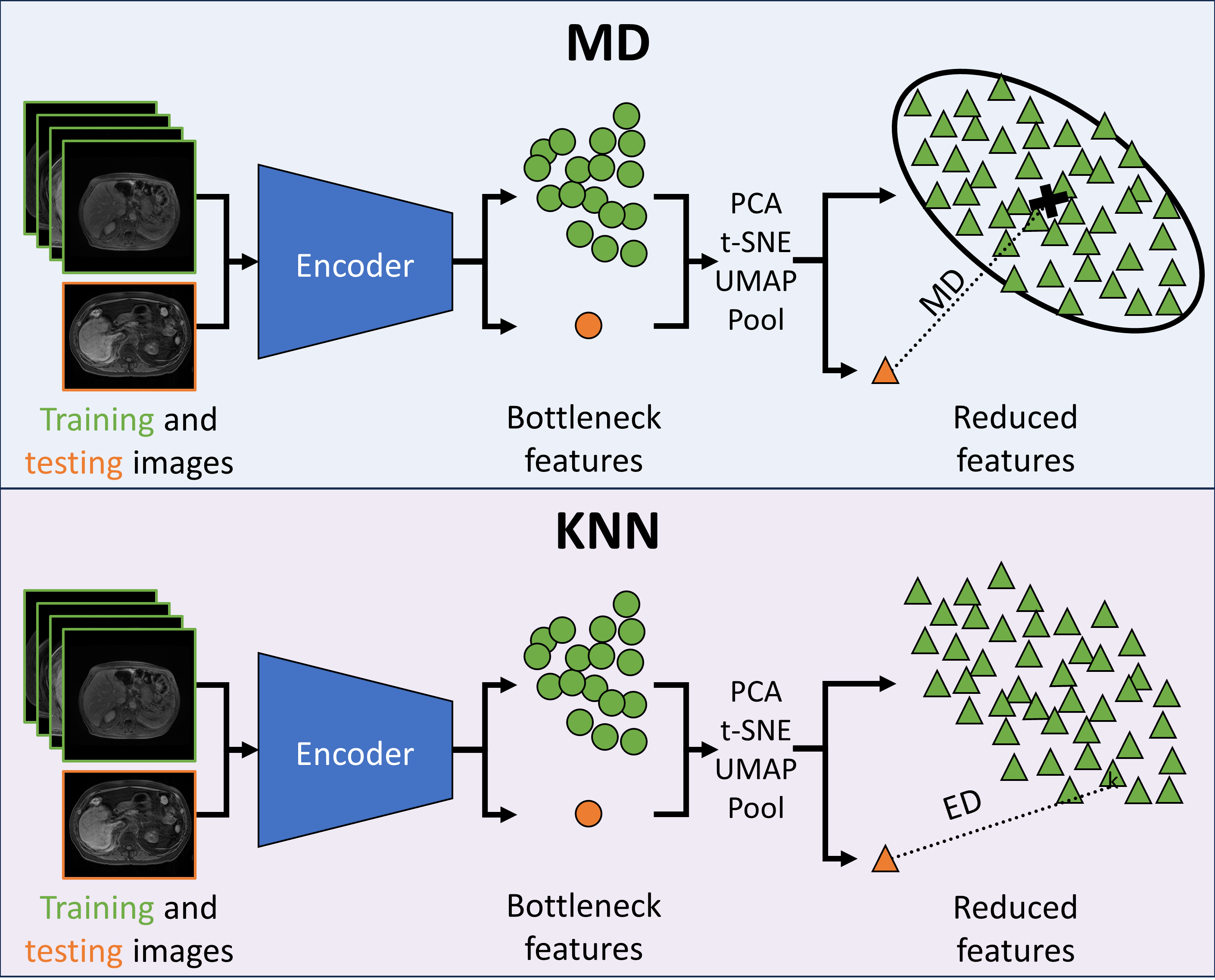}
    \caption{MD and KNN pipelines with dimensionality-reduced features using either PCA, t-SNE, UMAP, or average pooling (Pool).
    The encoder is a trained encoder from a U-Net architecture. 
    k is the k-th nearest neighbor.}
    \label{fig:methods}
\end{figure}

As distances in extremely high-dimensional spaces often lose meaning \citep{Aggarwal2001}, experiments were performed on the effect of decreasing the size of the bottleneck features using average pooling, PCA, UMAP \citep{Leland2020}, and t-distributed stochastic neighbor embeddings (t-SNE) \citep{VanDerMaaten2008}.
For average pooling, features were pooled in both 2- and 3-dimensions with kernel size $k$ and stride $s$ for $(k,s)\in\{(2,1),(2,2),(3,1),(3,2),(4,1)\}$. 
For PCA, each embedding was flattened and standardized. 
For both PCA and UMAP, a hyperparameter search was performed over the number of components $n$ such that $n\in\{2,4,8,16,32,64,128,256\}$. 
The PyTorch, scikit-learn, and UMAP Python packages were used for the dimensionality reduction \citep{Leland2020}.
Outside of the hyperparameter searches mentioned above, default parameters were used.
A visual representation of integrating dimensionality reduction with MD and KNN is available in Figure \ref{fig:methods}.

The efficacy of PCA, t-SNE, and UMAP for OOD detection was examined qualitatively by plotting the features from the MRI UNETR reduced to two dimensions.
Features generated by the MRI$_{\text{Tr}}$ and MRI$_{\text{Te}}$ datasets (split into ID and OOD by 95\% DSC) were plotted and subsequently compared.
Additionally, MRI$_{\text{Te}}$ embeddings were plotted by DSC, and MRI$_{\text{Tr}}$ embeddings were plotted by source (DLDS, AMOS, and CHAOS).

\subsubsection{Comparison Methods}

We used MSP \citep{Hendrycks2017} as a post-hoc detection baseline.
In addition, we evaluated two proposed improvements to MSP: temperature scaling \citep{Guo2017} and energy scoring \citep{Liu2020}.
To provide non-post-hoc detection baselines, MD and KNN were further compared with MC Dropout \citep{Gal2016} and ensembling \citep{Lakshminarayanan2017} on the MRI$_{\text{Te}}$ dataset.

To undertake the OOD detection task using MSP, the logits pertaining to the foreground and background classes were stacked, followed by a softmax.
The maximum probability for each voxel was then calculated across the foreground and background classes.
The average of these maximum probabilities for the entire image was then computed.
We subtracted this average from 1 to get the final OOD score for MSP.
For temperature scaling, logits were divided evenly by $T\in\{2,3,4,5, 10, 100, 1000\}$ before the softmax was applied.
For energy scoring, the OOD score was calculated as \begin{equation*}
E(x;f)=-T*\log \sum_{i=1}^k \frac{f_i(x)}{T}
\end{equation*} 
for image $x$, $i$-th logit $f_i$, and scaling parameter $T\in\{1,2, 3, 4, 5, 10,100,1000\}$.
For MC Dropout, voxel-wise standard deviations were calculated across the predictions from the five forward passes of MRI Dropout with dropout enabled.
The average of these standard deviations was used as the MC Dropout OOD score.
Similarly, the average of voxel-wise standard deviations calculated across the predictions from the five members of MRI Ensemble was used as the ensembling OOD score.

To further evaluate the performance of the OOD detection methods on the poor segmentation performance task, Pearson correlation coefficients (PCCs) were computed between the OOD scores of the best-performing configuration of each OOD detection method and the DSC, HD, and NSD segmentation metrics with a significance level of $\alpha=.10$.
These relationships were further explored qualitatively by plotting the OOD scores against DSCs.

\section{Results}

\subsection{Segmentation}

\begin{figure}[h]
    \centering
    \includegraphics[width=\textwidth]{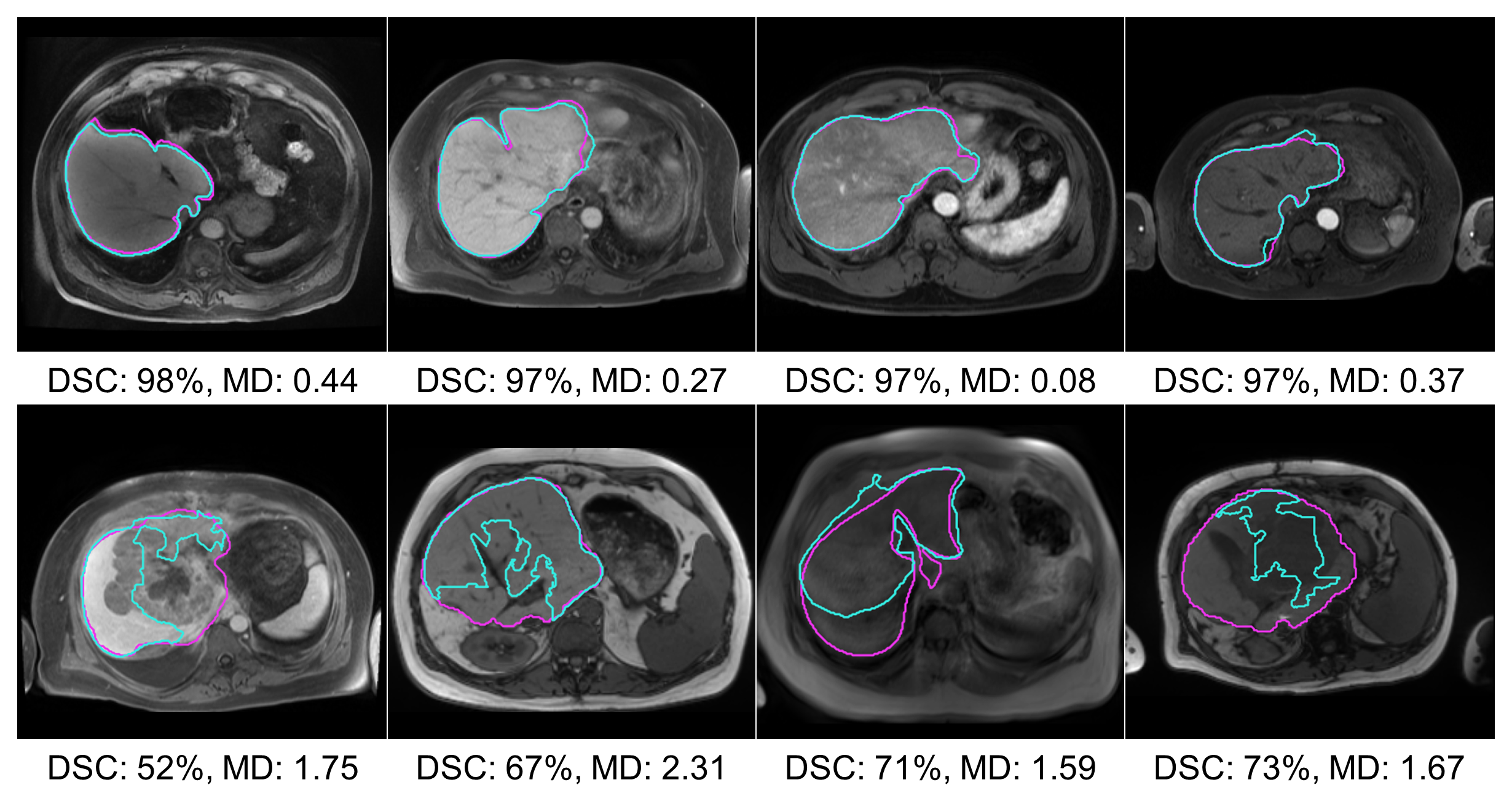}
    \caption{Segmentations with high (top) and low (bottom) DSCs along with their corresponding MDs, calculated in conjunction with PCA with two components.
    Pink is the ground truth segmentation; 
    teal is the MRI UNETR segmentation.}
    \label{fig:dsc_md}
\end{figure}

The segmentation performance of the MRI Dropout and MRI Ensemble models improved on that of the MRI UNETR by averaging the predictions (paired \emph{t}-tests for HD, $p = .013$ Dropout, $p = .020$ Ensemble; Table \ref{tab:seg}).
Thresholds for OOD detection were set at 95\% DSC for all models except the MRI+ UNETR.
This model only produced one DSC over 95\%, so the threshold was lowered to 80\%.
13 images determined to be OOD were shared across the MRI UNETR, MRI Dropout, and MRI Ensemble models.
MRI+ UNETR and MRI+ nnU-net performed similarly on MRI+$_{\text{Te}}$, with MRI+ UNETR achieving a lower HD and MRI+ nnU-net achieving a higher NSD (paired \emph{t}-tests, $p<.001$ all tests).
Figure \ref{fig:dsc_md} displays visual examples of the segmentation quality of the MRI UNETR.

\begin{table}[h]
    \centering
    \caption{Average ($\pm$SD) segmentation performances.
    \# OOD refers to the number of test images determined to be OOD.
    Arrows denote whether higher or lower is better.
    Bold highlights the best performance per test dataset, with underlined performances denoting statistical significance.}
    \label{tab:seg}
    \begin{tabular}{|l|c|c|c|c|}
    \hline
    \textbf{Model} & \textbf{DSC} {\small($\pm$SD)} $\uparrow$ & \textbf{HD} {\small($\pm$SD)} $\downarrow$ & \textbf{NSD} {\small($\pm$SD)} $\uparrow$ & \# OOD\\
    \hline
    MRI UNETR & 0.89 {\small ($\pm$0.11)} & 34.38 {\small ($\pm$25.67)} & 0.72 {\small ($\pm$0.22)} & 14\\
    \hline
    MRI Dropout & \textbf{0.91} {\small($\pm$0.09)} & \textbf{23.30} {\small($\pm$24.63)} & 0.75 {\small($\pm$0.20)}& 14\\
    \hline
    MRI Ensemble & \textbf{0.91} {\small ($\pm$0.10)} & 24.23 {\small($\pm$24.94)} & \textbf{\underline{0.76}} {\small ($\pm$0.20)}& 13\\
    \hline
    MRI+ UNETR & \textbf{\underline{0.90}} {\small($\pm$0.04)} & \textbf{\underline{18.10}} {\small($\pm$10.66)}& 0.65 {\small($\pm$0.10)}& 7\\
    \hline
    MRI+ nnU-net & \textbf{\underline{0.90}} {\small($\pm$0.03)}& 31.27 {\small($\pm$25.06)}& \textbf{\underline{0.78}} {\small($\pm$0.08)}& 349 \\
    \hline
    CT nnU-net & \textbf{\underline{0.97}} {\small($\pm$0.01)} & \textbf{\underline{23.10}} {\small($\pm$26.34)} & \textbf{\underline{0.96}} {\small($\pm$0.05)}& 22\\
    \hline
    \end{tabular}
\end{table}

\subsection{OOD Detection}

\subsubsection{Mahalanobis Distance}

The calculation of the MD on raw bottleneck features was computationally intractable (Table \ref{tab:results_md}). 
For the MRI UNETR, the inverse of the covariance matrix took $\sim$2.6 hours to compute. 
Once computed, it took 75.5 GB to store the inverse. 
Once the matrix was in memory, each MD calculation took $\sim$2 seconds. 
The average ($\pm$ SD) MD on MRI$_{\text{Tr}}$ was 1203.02 ($\pm$ 24.66); whereas, the average ($\pm$ SD) MD on MRI$_{\text{Te}}$ was 1.47$\times 10^{9}$ ($\pm 8.66\times 10^{8}$) and $1.52\times 10^{9}(\pm 9.10\times 10^{8})$ for ID and OOD images, respectively.

\begin{table}[h]
    \centering
    \caption{MD-based OOD detection results.
    Results are based on a 95\% DSC threshold for all models except the MRI+ UNETR, where the threshold was set at 80\% DSC.
    Only the best-performing configuration by AUROC is reported for each dimensionality reduction technique (Reduct): no reduction (None), PCA($n_p$) with $n_p$ components, t-SNE, UMAP($n_u$) with $n_u$ components, and $n_d$-dimensional average pooling with kernel size $k$ and stride $s$, Pool$n_d$D($k$,$s$).
    Seconds is the amount of time it took to calculate the test distances.
    The results are averages ($\pm$SD) across 5 runs.
    Arrows denote whether higher or lower is better.
    Bold highlights the best performance per model, with underlined performances denoting statistical significance.
    Appendix \ref{sec:thres_md} contains the results with varied thresholds, and Appendix \ref{sec:hyper_md} contains the full hyperparameter searches.
    }\label{tab:results_md}
    \begin{tabularx}{\textwidth}{|X|l|c|c|c|c|}
    \cline{1-6}
    \textbf{Model} & \textbf{Reduct} & \textbf{AUROC} $\uparrow$ & \textbf{AUPRC} $\uparrow$ & \textbf{FPR90} $\downarrow$ &\textbf{Seconds} $\downarrow$\\
    \cline{1-6}
    \multirow{6}{=}{\parbox{2cm}{MRI \\UNETR}}&None & 0.48 {\small ($\pm$0.00)}& 0.61 {\small ($\pm$0.00)} & 1.00 {\small ($\pm$0.00)} & 9,354.34 {\small ($\pm$48.53)}\\
    \cline{2-6}
    & PCA(256) & \textbf{\underline{0.93}} {\small ($\pm$0.00)} & \textbf{\underline{0.94}} {\small ($\pm$0.00)} & \textbf{\underline{0.23}} {\small ($\pm$0.00)} & \textbf{\underline{2.82}} {\small ($\pm$0.14)} \\
    \cline{2-6}
    & t-SNE & 0.70 {\small ($\pm$0.08)} & 0.72 {\small ($\pm$0.12)} & 0.71 {\small ($\pm$0.14)} & 4.70 {\small ($\pm$0.28)}\\
    \cline{2-6}
    & UMAP(2) & 0.77 {\small ($\pm$0.08)} & 0.79 {\small ($\pm$0.11)} & 0.57 {\small ($\pm$0.08)} & 10.44 {\small ($\pm$0.22)}\\
    \cline{2-6}
    & Pool2D(3,2) & 0.82 {\small ($\pm$0.00)} & 0.86 {\small ($\pm$0.00)} & 0.46 {\small ($\pm$0.00)} & 15.32 {\small ($\pm$8.92)}\\
    \cline{1-6}
    \multirow{6}{=}{\parbox{2cm}{MRI+ \\ UNETR}}& None&0.53 {\small ($\pm$0.00)}&0.03 {\small ($\pm$0.00)}& 0.78 {\small ($\pm$0.00)} & 10,070.78 {\small ($\pm$141.69)}\\
    \cline{2-6}
    & PCA(32) & \textbf{\underline{0.85}} {\small ($\pm$0.01)} & \textbf{\underline{0.13}} {\small ($\pm$0.00)} & \textbf{\underline{0.34}} {\small ($\pm$0.02)} & \textbf{\underline{1.86}} {\small ($\pm$0.32)}\\
    \cline{2-6}
    & t-SNE & 0.66 {\small ($\pm$0.00)} & 0.03 {\small ($\pm$0.00)} & 0.45 {\small ($\pm$0.00)} & 5.77 {\small ($\pm$0.06)}\\
    \cline{2-6}
    & UMAP(2) & 0.68 {\small ($\pm$0.07)} & 0.05 {\small ($\pm$0.01)} & 0.49 {\small ($\pm$0.06)} & 21.29 {\small ($\pm$0.43)}\\
    \cline{2-6}
    & Pool2D(4,1) & 0.64 {\small ($\pm$0.00)} & 0.04 {\small ($\pm$0.00)} & 0.64 {\small ($\pm$0.00)} & 16.08 {\small ($\pm$13.41)}\\
    \cline{1-6}
    \multirow{6}{=}{\parbox{2cm}{MRI+ \\ nnU-net}}&None&0.69 {\small ($\pm$0.00)}&\textbf{\underline{1.00}} {\small ($\pm$0.00)}&0.67 {\small ($\pm$0.00)}&4,125.96 {\small ($\pm$13.12)}\\
    \cline{2-6}
    & PCA(8) &\textbf{\underline{0.96}} {\small ($\pm$0.00)}&\textbf{\underline{1.00}} {\small ($\pm$0.00)}& \textbf{\underline{0.00}} {\small ($\pm$0.00)} & \textbf{\underline{1.12}} {\small ($\pm$0.04)}\\
    \cline{2-6}
    & t-SNE&0.70 {\small ($\pm$0.18)}&\textbf{\underline{1.00}} {\small ($\pm$0.00)}&0.87 {\small ($\pm$0.16)}&4.72 {\small ($\pm$0.11)}\\
    \cline{2-6}
    & UMAP(16) &0.82 {\small ($\pm$0.08)}&\textbf{\underline{1.00}} {\small ($\pm$0.00)}& 0.67 {\small ($\pm$0.03)}&19.07 {\small ($\pm$0.96)}\\
    \cline{2-6}
    & Pool2D(2,1) &0.85 {\small ($\pm$0.00)} &\textbf{\underline{1.00}} {\small ($\pm$0.00)} &0.67 {\small ($\pm$0.00)}&1,579.73 {\small ($\pm$52.41)}\\
    \cline{1-6}
    \multirow{6}{=}{\parbox{2cm}{CT \\ nnU-net}}&None& 0.41 {\small ($\pm$0.00)} & 0.10 {\small ($\pm$0.00)} & 0.91 {\small ($\pm$0.00)} & 5,856.21 {\small ($\pm$63.04)}\\
    \cline{2-6}
    & PCA(32) & 0.56 {\small ($\pm$0.00)} & 0.17 {\small ($\pm$0.00)} & 0.92 {\small ($\pm$0.00)} & \textbf{\underline{8.17}} {\small ($\pm$0.23)}\\
    \cline{2-6}
    & t-SNE& 0.59 {\small ($\pm$0.04)} & 0.20 {\small ($\pm$0.02)} & 0.80 {\small ($\pm$0.02)} & 13.75 {\small ($\pm$0.35)}\\
    \cline{2-6}
    & UMAP(128) & \textbf{\underline{0.68}} {\small ($\pm$0.03)} & \textbf{0.22} {\small ($\pm$0.01)} & \textbf{0.74} {\small ($\pm$0.09)} & 288.42 {\small ($\pm$25.99)} \\
    \cline{2-6}
    & Pool2D(2,2) & 0.59 {\small ($\pm$0.00)} & 0.13 {\small ($\pm$0.00)} & 0.84 {\small ($\pm$0.00)} & 163.84 {\small ($\pm$19.54)}\\
    \cline{1-6}
    \end{tabularx}
\end{table}

\subsubsection{K-Nearest Neighbors}
While the MD calculation was not tractable on raw features, the KNN calculation was (Table \ref{tab:results_knn}). 
The calculation took $\sim$0.02 seconds per image for the MRI data and $\sim$0.08 seconds per image for the liver CT data.
In addition to being more scalable, KNN improved the AUROC over the MD applied to raw features for all models (\emph{t}-tests, $p<.001$ all tests).

\begin{table}[h]
    \centering
    \caption{KNN-based OOD detection of poor performance results.
    Results are based on a 95\% DSC threshold for all models except the MRI+ UNETR, where the threshold was set at 80\% DSC.
    Only the best-performing configuration by AUROC is reported for each dimensionality reduction technique (Reduct): no reduction (None), PCA($n_p$) with $n_p$ components, t-SNE, UMAP($n_u$) with $n_u$ components, and $n_d$-dimensional average pooling with kernel size $k$ and stride $s$, Pool$n_d$D($k$,$s$).
    Seconds is the amount of time it took to calculate the test distances.
    The results are averages ($\pm$SD) across 5 runs.
    Arrows denote whether higher or lower is better.
    Bold highlights the best performance, with underlined performances denoting statistical significance.
    Appendix \ref{sec:thres_knn} contains the results with varied thresholds, and Appendix \ref{sec:hyper_knn} contains the link to the full hyperparameter searches.
    }\label{tab:results_knn}
    \begin{tabularx}{\textwidth}{|X|l|c|c|c|c|c|}
    \cline{1-7}
    \textbf{Model} & \textbf{Reduct} & \textbf{K} & \textbf{AUROC} $\uparrow$ & \textbf{AUPRC} $\uparrow$ & \textbf{FPR90} $\downarrow$ &\textbf{Seconds} $\downarrow$\\
    \cline{1-7}
    \multirow{6}{=}{\parbox{2cm}{MRI \\UNETR}}&None& 256& 0.87 {\small ($\pm$0.00)} & 0.88 {\small($\pm$0.00)} & 0.31 {\small ($\pm$0.00)} & \textbf{\underline{0.78}} {\small ($\pm$0.00)}\\
    \cline{2-7}
    & PCA(2) & 256 & 0.90 {\small ($\pm$0.00)} & 0.92 {\small ($\pm$0.00)} & 0.31 {\small ($\pm$0.00)} & 0.95 {\small ($\pm$0.05)}\\
    \cline{2-7}
    & t-SNE & 256 &  0.77 {\small ($\pm$0.05)} & 0.83 {\small ($\pm$0.04)} & 0.74 {\small ($\pm$0.06)} & 4.48 {\small ($\pm$0.07)} \\
    \cline{2-7}
    & UMAP(32) & 256 & 0.83 {\small ($\pm$0.05)} & 0.85 {\small ($\pm$0.04)} & 0.51 {\small ($\pm$0.08)} & 6.70 {\small ($\pm$0.15)}\\
    \cline{2-7}
    & Pool2D(3,1) & 256 & \textbf{\underline{0.94}} {\small ($\pm$0.00)} & \textbf{\underline{0.95}} {\small ($\pm$0.00)} & \textbf{\underline{0.23}} {\small ($\pm$0.00)} & 0.90 {\small ($\pm$0.00) }\\
    \cline{1-7}
    \multirow{6}{=}{\parbox{2cm}{MRI+ \\ UNETR}}&None& 256 & 0.76 {\small ($\pm$0.00)}& 0.25 {\small ($\pm$0.00)} & 0.59 {\small ($\pm$0.00)}& 9.44 {\small ($\pm$0.16)}\\
    \cline{2-7}
    & PCA(32) & 64 & 0.84 {\small ($\pm$0.01)}& 0.17 {\small ($\pm$0.02)}& 0.40 {\small ($\pm$0.01)}& \textbf{\underline{1.57}} {\small ($\pm$0.06)}\\
    \cline{2-7}
    & t-SNE & 64 & 0.70 {\small ($\pm$0.00)} & 0.04 {\small ($\pm$0.00)} & \textbf{0.37} {\small ($\pm$0.00)}& 6.49 {\small ($\pm$0.19)}\\
    \cline{2-7}
    & UMAP(4) & 64 & 0.78 {\small($\pm$0.05)}& 0.09 {\small($\pm$0.06)}&0.42 {\small($\pm$0.07)}& 15.70 {\small($\pm$0.45)}\\
    \cline{2-7}
    & Pool3D(2,2)& 256 & \textbf{\underline{0.87}} {\small($\pm$0.00)}& \textbf{\underline{0.33}} {\small($\pm$0.00)}& 0.43 {\small($\pm$0.00)}& 11.21 {\small($\pm$7.66)}\\
    \cline{1-7}
    \multirow{6}{=}{\parbox{2cm}{MRI+ \\ nnU-net}}&None&256& 0.96 {\small ($\pm$0.00)} & \textbf{\underline{1.00}} {\small ($\pm$0.00)} & \textbf{0.00} {\small($\pm$0.00)}& 6.78 {\small($\pm$0.10)}\\
    \cline{2-7}
    & PCA(8)& 256 & 0.97 {\small($\pm$0.00)}& \textbf{\underline{1.00}} {\small($\pm$0.00)}&\textbf{0.00} {\small($\pm$0.00)}& 1.16 {\small($\pm$0.07)}\\
    \cline{2-7}
    & t-SNE& 128 & 0.67 {\small($\pm$0.08)}&\textbf{\underline{1.00}} {\small($\pm$0.00)}&1.00 {\small($\pm$0.00)}& 4.76 {\small ($\pm$0.11)}\\
    \cline{2-7}
    & UMAP(2) & 2 & 0.96 {\small($\pm$0.04)}&\textbf{\underline{1.00}} {\small($\pm$0.00)}&0.13 {\small($\pm$0.16)}&14.91 {\small($\pm$2.08)}\\
    \cline{2-7}
    & Pool2D(2,2)& 256 & \textbf{0.98} {\small ($\pm$0.00)} & \textbf{\underline{1.00}} {\small ($\pm$0.00)} & \textbf{0.00} {\small ($\pm$0.00)} & \textbf{\underline{0.74}} {\small ($\pm$0.04)}\\
    \cline{1-7}
    \multirow{6}{=}{\parbox{2cm}{CT \\ nnU-net}}&None& 8 & 0.52 {\small($\pm$0.00)}& 0.13 {\small($\pm$0.00)}& 0.94 {\small($\pm$0.00)}& 37.64 {\small($\pm$0.67)}\\
    \cline{2-7}
    & PCA(8) & 4 & 0.55 {\small($\pm$0.00)}& 0.15 {\small($\pm$0.00)}& 0.97 {\small($\pm$0.00)}& \textbf{\underline{4.94}} {\small($\pm$0.04)}\\
    \cline{2-7}
    & t-SNE& 256 & 0.46 {\small($\pm$0.00)}& 0.19 {\small($\pm$0.00)}&0.95 {\small($\pm$0.01)}& 12.28 {\small($\pm$0.09)}\\
    \cline{2-7}
    & UMAP(4) & 256 & \textbf{\underline{0.65}} {\small($\pm$0.01)}& \textbf{\underline{0.24}} {\small($\pm$0.05)}& \textbf{\underline{0.88}} {\small($\pm$0.02)}& 199.93 {\small($\pm$1.96)}\\
    \cline{2-7}
    & Pool3D(2,2) & 4 & 0.54 {\small($\pm$0.00)}& 0.14 {\small($\pm$0.00)}& 0.96 {\small($\pm$0.00)}& 81.96 {\small($\pm$26.79)}\\
    \cline{1-7}
    \end{tabularx}
\end{table}

\subsubsection{Dimensionality Reduction}

Paired with MD, all dimensionality reduction techniques resulted in improvements in the AUROC (\emph{t}-tests, $p=.003$ UMAP/MRI+ UNETR, $p<.001$ all other tests; Table \ref{tab:results_md}).
On the MRI models, PCA achieved the best performance, outperforming average pooling by 0.14 ($\pm$0.06)\% AUROC and 535.11 ($\pm$903.70) seconds.
For CT nnU-net, UMAP achieved the best AUROC, outperforming average pooling by 0.09.
Figure \ref{fig:dsc_md} displays MDs computed on PCA-reduced features, along with the corresponding segmentations.
In this figure, higher MDs were associated with poor segmentation performance.

Paired with KNN, PCA, and average pooling resulted in AUROC improvements for all models (\emph{t}-tests, $p<.001$ all tests; Table \ref{tab:results_knn}).
Similar to MD, KNN on UMAP-reduced features achieved the highest AUROC for the CT nnU-net (\emph{t}-tests, $p<.001$ all tests).
In contrast to MD, KNN applied to average pooled features achieved the highest AUROCs for the MRI models (\emph{t}-tests, $p=.001$ UMAP/MRI UNETR, $p=.007$ UMAP/MRI+ UNETR, $p<.001$ all other tests except UMAP/MRI+ nnU-net).
On the MRI models, KNN outperformed MD when applied to average pooled features by 0.16 ($\pm$0.05) AUROC and 532.76 ($\pm$739.81) seconds.
Overall, KNN applied to average-pooled features slightly outperformed MD on PCA-reduced features by 0.02 ($\pm$0.00) AUROC for the MRI models (\emph{t}-tests, $p=.003$ MRI+ UNETR, $p<.001$ all other tests).

\begin{figure}[h]
    \centering
    \includegraphics[width=\textwidth]{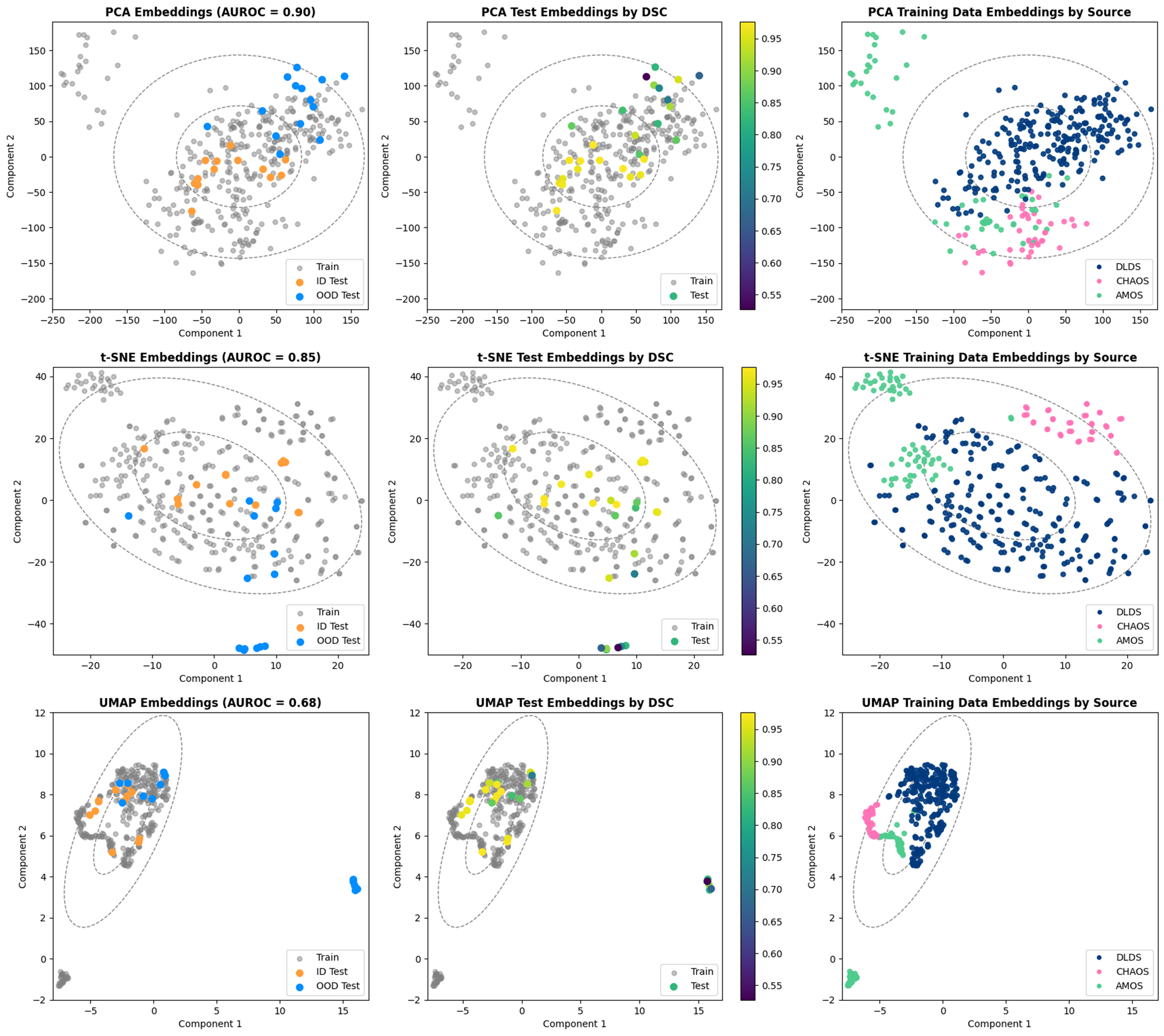}
    \caption{Visualization of 2D projections of MRI UNETR embeddings. 
    (Top Row) PCA projections. 
    (Middle Row) t-SNE projections. 
    (Bottom Row) UMAP projections. 
    (Left Column) Test projections split into ID and OOD by 95\% DSC.
    (Middle Column) Test projections by DSC.
    (Right Column) Projections for the training data by source. 
    The gray ellipses are the covariance ellipses (one and two standard deviations) for the training distribution.}
    \label{fig:embeddings}
\end{figure}

Figure \ref{fig:embeddings} visualizes the 2D embeddings produced by PCA, t-SNE, and UMAP for the MRI UNETR.
In addition, covariance ellipses generated by the training distribution are plotted, representing one and two standard deviations away from the mean training embedding.
PCA mapped most ID test images within one standard deviation of the mean training embedding (the image not within the first deviation contained a motion artifact).
On the other hand, most OOD test images were mapped outside of the first standard deviation. 
When test embeddings were visualized by their DSC, the three reduction techniques mapped the images with the lowest DSCs farthest from the mean training embedding.
Moreover, all three techniques clustered training embeddings by source.
The 26 images from the AMOS dataset mapped outside the second standard deviation by all techniques were deemed to be of low perceptual resolution by a physician.
These were the only images from the AMOS dataset whose axial dimension was larger than the sagittal dimension.
Sample images from both AMOS clusters are shown in Figure \ref{fig:AMOS} in Appendix \ref{sec:figs}.

\subsubsection{Comparison Methods}

On MRI$_{\text{Te}}$, MSP, MC Dropout, and ensembling outperformed MD and KNN (\emph{t}-tests on AUROC, $p<.001$ all tests), with MC Dropout perfectly differentiating between ID and OOD categories (Table \ref{tab:comp}).
Similarly, MSP outperformed MD and KNN on CT$_{\text{Te}}$ (\emph{t}-tests on AUROC, $p=.036$ MD, $p<.001$ KNN).
In contrast, MD and KNN outperformed the output-based methods on MRI+$_{\text{Te}}$ (\emph{t}-tests on AUROC, $p<.001$ all tests).

Although originally intended to improve the performance of MSP, 
both temperature scaling and energy scoring performed worse than MSP on the MRI$_{\text{Te}}$ and CT$_{\text{Te}}$ datasets (\emph{t}-tests on AUROC, $p<.001$ all tests).
Overall, MSP achieved the highest AUROCs for the MRI UNETR and CT nnU-net models (\emph{t}-tests, $p=.036$ MD/CT nnU-net, $p<.001$ all other tests), and KNN achieved the highest AUROCs for the MRI+ UNETR and nnU-net models (\emph{t}-tests, $p=.003$ MD/MRI+ UNETR, $p<.001$ all other tests).

\begin{table}[h]
    \centering
    \caption{
    OOD detection results for the best-performing configurations of the comparison methods: MSP, temperature scaling (TS) and energy scoring (Energy) with temperature $T$, MC Dropout (MCD), and ensembling (Ensemble).
    The results are averages ($\pm$SD) across 5 runs.
    Arrows denote whether a higher or lower value is better.
    Bold highlights the best performance per model, with underlined performances denoting statistical significance.
    Appendix \ref{sec:thres_comp} contains the results with varied thresholds, and Appendix \ref{sec:hyper_comp} contains the full hyperparameter searches.
    }
    \label{tab:comp}
    \begin{tabularx}{\textwidth}{|p{1.5cm}|l|c|c|c|c|c|}
    \cline{1-7}
    \textbf{Model} & \textbf{Method} & \textbf{T} & \textbf{AUROC} $\uparrow$ & \textbf{AUPRC} $\uparrow$ & \textbf{FPR90} $\downarrow$ & \textbf{Seconds} $\downarrow$\\
    \cline{1-7}
    \multirow{3}{*}{\parbox{2cm}{MRI \\UNETR}}&MSP &-& \textbf{\underline{0.96}} {\small($\pm$0.00)}& \textbf{\underline{0.97}} {\small($\pm$0.00)}& \textbf{\underline{0.00}} {\small($\pm$0.00)}& 8.78 {\small($\pm$0.11)}\\
    \cline{2-7}
    &TS &2 & 0.90 {\small($\pm$0.00)}& 0.93 {\small($\pm$0.00)}& 0.15 {\small($\pm$0.00)}& 9.25 {\small($\pm$0.11)}\\
    \cline{2-7}
    &Energy &3 &0.55 {\small($\pm$0.00)} &0.57 {\small($\pm$0.00)}& 0.69 {\small($\pm$0.00)} & \textbf{7.08} {\small($\pm$0.14)} \\
    \cline{1-7}
    \parbox{2cm}{\vspace{.1cm}MRI \\Dropout\vspace{.1cm}}& MCD &-&\textbf{\underline{1.00}} {\small($\pm$0.00)} & \textbf{\underline{1.00}} {\small($\pm$0.00)} & \textbf{\underline{0.00}} {\small($\pm$0.00)} & \textbf{\underline{14.56}} {\small($\pm$0.20)}\\
    \cline{1-7}
    \parbox{2cm}{\vspace{.1cm}MRI \\ Ensemble\vspace{.1cm}} & Ensemble &-& \textbf{\underline{0.96}} {\small($\pm$0.00)} & \textbf{\underline{0.96}} {\small($\pm$0.00)} & \textbf{\underline{0.14}} {\small($\pm$0.00)} & \textbf{\underline{14.02}} {\small($\pm$0.05)} \\
    \cline{1-7}
    \multirow{3}{*}{\parbox{2cm}{MRI+ \\UNETR}} & MSP &-& 0.57 {\small($\pm$0.00)} & \textbf{\underline{0.09}} {\small($\pm$0.00)} & \textbf{\underline{0.59}} {\small($\pm$0.00)} & 58.70 {\small($\pm$0.48)}\\
    \cline{2-7}
    &TS & 2& 0.47 {\small($\pm$0.00)} & 0.05 {\small($\pm$0.00)} & 0.78 {\small($\pm$0.00)} & 60.57 {\small($\pm$0.31)}\\
    \cline{2-7}
    &Energy &1000& \textbf{\underline{0.61}} {\small($\pm$0.00)} & 0.03 {\small($\pm$0.00)} & 0.71 {\small($\pm$0.00)} & \textbf{\underline{35.86}} {\small($\pm$0.41)}\\ 
    \cline{1-7}
    \multirow{3}{*}{\parbox{2cm}{MRI+ \\nnU-net}}& MSP&- & 0.45 {\small($\pm$0.00)} & 0.99 {\small($\pm$0.00)} & \textbf{\underline{1.00}} {\small($\pm$0.00)} & 1,114.34 {\small($\pm$0.51)}\\
    \cline{2-7}
    &TS &10 & 0.55 {\small($\pm$0.00)} & 0.99 {\small($\pm$0.00)} & \textbf{\underline{1.00}} {\small($\pm$0.00)} & 1,252.78 {\small($\pm$0.73)}\\
    \cline{2-7}
    & Energy &10 &\textbf{\underline{0.61}} {\small($\pm$0.00)} & \textbf{\underline{1.00}} {\small($\pm$0.00)} & \textbf{\underline{1.00}} {\small($\pm$0.00)} & \textbf{\underline{186.88}} {\small($\pm$0.47)}\\
    \cline{1-7}
    \multirow{3}{*}{\parbox{2cm}{CT \\nnU-net}}&MSP & -&\textbf{\underline{0.72}} {\small($\pm$0.00)} & \textbf{\underline{0.29}} {\small($\pm$0.00)} & \textbf{\underline{0.57}} {\small($\pm$0.00)} & 568.51 {\small($\pm$0.37)}\\
    \cline{2-7}
    &TS& 3 & 0.68 {\small($\pm$0.00)} & 0.26 {\small($\pm$0.00)} & 0.69 {\small($\pm$0.00)} & 699.07 {\small($\pm$0.37)}\\
    \cline{2-7}
    &Energy &2 & 0.67 {\small($\pm$0.00)} & 0.26 {\small($\pm$0.00)} & 0.75 {\small($\pm$0.00)} & \textbf{\underline{105.66}} {\small($\pm$0.60)} \\  
    \cline{1-7}
    \end{tabularx}
\end{table}

The OOD scores from KNN, MSP, and temperature scaling were significantly correlated with DSC across all models (Table \ref{tab:corr}).
MD was significantly correlated with DSC for all the MRI models.
Energy scoring was significantly correlated with DSC for only the nnU-nets.
MC Dropout and ensembling were significantly correlated with all segmentation metrics on MRI$_{\text{Te}}$, notably achieving PCCs of 0.96 and 0.97 with HD ($p<.001$ all correlations).
Considering only post-hoc detection methods, MSP achieved the best correlations with DSC for the MRI UNETR and CT nnU-net models (-0.77 and -0.28) and the best correlations with HD and NSD for MRI+ nnU-net (0.30 and -0.31; \emph{t}-tests, $p<.001$ all tests).
MD and temperature scaling achieved the best correlations with DSC for the MRI+ UNETR and MRI+ nnU-net models, respectively (-0.14 and -0.31; \emph{t}-tests, $p<.001$ all tests).

\begin{table}[h]
    \centering
    \caption{
    Correlation results for the best-performing configuration of all methods: MD and KNN with dimensionality-reduction techniques PCA($n$) and UMAP($n$) with $n$ components and $n_d$-dimensional average pooling with kernel size $k$ and stride $s$ Pool$n_d$D($k$,$s$), MSP, temperature scaling (TS) and energy scoring (Energy) with temperature $T$, MC Dropout (MCD), and ensembling (Ensemble).
    $^{*}$ represents that each of the five correlation coefficients that were averaged over were statistically significant.
    Arrows denote whether a higher or lower value is better.
    Bold highlights the best performance per model, with underlined performances denoting statistical significance.
    }
    \label{tab:corr}
    \begin{tabularx}{\textwidth}{|p{1.5cm}|l|c|c|c|c|}
    \cline{1-6}
    \textbf{Model} & \textbf{Method} & \textbf{Config} & \textbf{PCC [DSC]} $\downarrow$ & \textbf{PCC [HD]} $\uparrow$ & \textbf{PCC [NSD]} $\downarrow$\\
    \cline{1-6}
    \multirow{5}{*}{\parbox{2cm}{MRI\\ UNETR}}& MD &PCA(256) & -0.74* {\small($\pm$0.00)} & 0.09 {\small($\pm$0.00)} & \textbf{\underline{-0.76*}} {\small($\pm$0.00)}\\
    \cline{2-6}
    &KNN&\parbox{2cm}{\centering\vspace{.1cm}Pool2D(3,1) \\ K=256\vspace{.1cm}} & -0.72* {\small($\pm$0.00)} & 0.05 {\small($\pm$0.00)} & -0.73* {\small($\pm$0.00)}\\
    \cline{2-6}
    &MSP & - &\textbf{\underline{-0.77}}* {\small($\pm$0.00)} & 0.53* {\small($\pm$0.00)} & -0.70* {\small($\pm$0.00)} \\
    \cline{2-6}
    &TS & T=2&-0.69* {\small($\pm$0.00)} & \textbf{\underline{0.58}}* {\small($\pm$0.00)} & -0.64* {\small($\pm$0.00)} \\
    \cline{2-6}
    &Energy & T=3 &-0.22 {\small($\pm$0.00)} & 0.20 {\small($\pm$0.00)} & -0.24 {\small($\pm$0.00)}\\
    \cline{1-6}
    \parbox{2cm}{\vspace{.1cm}MRI \\ Dropout\vspace{.1cm}}& MCD & - &\textbf{\underline{-0.86}}* {\small($\pm$0.00)}& \textbf{\underline{0.96*}} {\small($\pm$0.00)} & \textbf{\underline{-0.67*}} {\small($\pm$0.00)}\\
    \cline{1-6}
    \parbox{2cm}{\vspace{.1cm}MRI \\ Ensemble\vspace{.1cm}} & Ensemble & - &\textbf{\underline{-0.82}}* {\small($\pm$0.00)} & \textbf{\underline{0.97}}* {\small($\pm$0.00)} & \textbf{\underline{-0.63*}} {\small($\pm$0.00)}\\
    \cline{1-6}
    \multirow{5}{*}{\parbox{2cm}{MRI+ \\UNETR}}& MD & PCA(32) & \textbf{\underline{-0.14*}} {\small($\pm$0.00)} & 0.05 {\small($\pm$0.00)} & 0.02 {\small($\pm$0.00)}\\
    \cline{2-6}
    &KNN& \parbox{2cm}{\centering\vspace{.1cm}Pool3D(2,2)\\K=256\vspace{.1cm}}&-0.13* {\small($\pm$0.00)}& 0.03 {\small($\pm$0.00)} & \textbf{\underline{-0.03}} {\small($\pm$0.00)}\\
    \cline{2-6}
    &MSP & - & -0.13* {\small($\pm$0.00)} & -0.01 {\small($\pm$0.00)}& 0.10 {\small($\pm$0.00)}\\
    \cline{2-6}
    &TS & T=2&-0.13* {\small($\pm$0.00)} & -0.02 {\small($\pm$0.00)} & 0.12 {\small($\pm$0.00)}\\
    \cline{2-6}
    &Energy  & T=1000&-0.04 {\small($\pm$0.00)} & \textbf{\underline{0.09}} {\small($\pm$0.00)} & \textbf{\underline{-0.03}} {\small($\pm$0.00)}\\      
    \cline{1-6}
    \multirow{5}{*}{\parbox{2cm}{MRI+ \\nnU-net}}&MD&PCA(8)& -0.20* {\small($\pm$0.00)}& 0.11* {\small($\pm$0.00)} & 0.12* {\small($\pm$0.00)}\\
    \cline{2-6}
    &KNN&\parbox{2cm}{\centering\vspace{.1cm}Pool2D(2,2)\\K=256\vspace{.1cm}} &-0.27* {\small($\pm$0.00)}& 0.22* {\small($\pm$0.00)}& 0.14* {\small($\pm$0.00)}\\
    \cline{2-6}
    &MSP & - & -0.13* {\small($\pm$0.00)}& \textbf{\underline{0.30}}* {\small($\pm$0.00)} & \textbf{\underline{-0.31}}* {\small($\pm$0.00)}\\
    \cline{2-6}
    &TS & T=10&\textbf{\underline{-0.31}}* {\small($\pm$0.00)} & 0.20* {\small($\pm$0.00)}& 0.04 {\small($\pm$0.00)}\\
    \cline{2-6}
    &Energy & T=10&-0.16* {\small($\pm$0.00)} & -0.02 {\small($\pm$0.00)} & 0.07 {\small($\pm$0.00)} \\  
    \cline{1-6}
    \multirow{5}{*}{\parbox{2cm}{CT \\nnU-net}}&MD& UMAP(128)&-0.11 {\small($\pm$0.03)} & 0.05 {\small($\pm$0.04)}& -0.10* {\small($\pm$0.02)}\\
    \cline{2-6}
    &KNN& \parbox{2cm}{\centering\vspace{.1cm}UMAP(4)\\K=256\vspace{.1cm}}&-0.21* {\small($\pm$0.01)} & \textbf{\underline{0.23}}* {\small($\pm$0.01)}& -0.16* {\small($\pm$0.01)}\\
    \cline{2-6}
    &MSP & - & \textbf{\underline{-0.28}}* {\small($\pm$0.00)} & \textbf{\underline{0.23}}* {\small($\pm$0.00)} & \textbf{\underline{-0.29}}* {\small($\pm$0.00)} \\
    \cline{2-6}
    &TS & T=3&-0.22* {\small($\pm$0.00)} & 0.20* {\small($\pm$0.00)}& -0.25* {\small($\pm$0.00)}\\
    \cline{2-6}
    &Energy & T=2&-0.20* {\small($\pm$0.00)} & 0.19* {\small($\pm$0.00)} & -0.23* {\small($\pm$0.00)}\\ 
    \cline{1-6}
    \end{tabularx}
\end{table}

\begin{figure}
    \centering
    \caption{OOD scores plotted against DSC for MRI$_{\text{Te}}$.
    Horizontal lines represent 95\% DSC.
    Vertical lines represent the 90\% TPR.}
    \includegraphics[width=.8\textwidth]{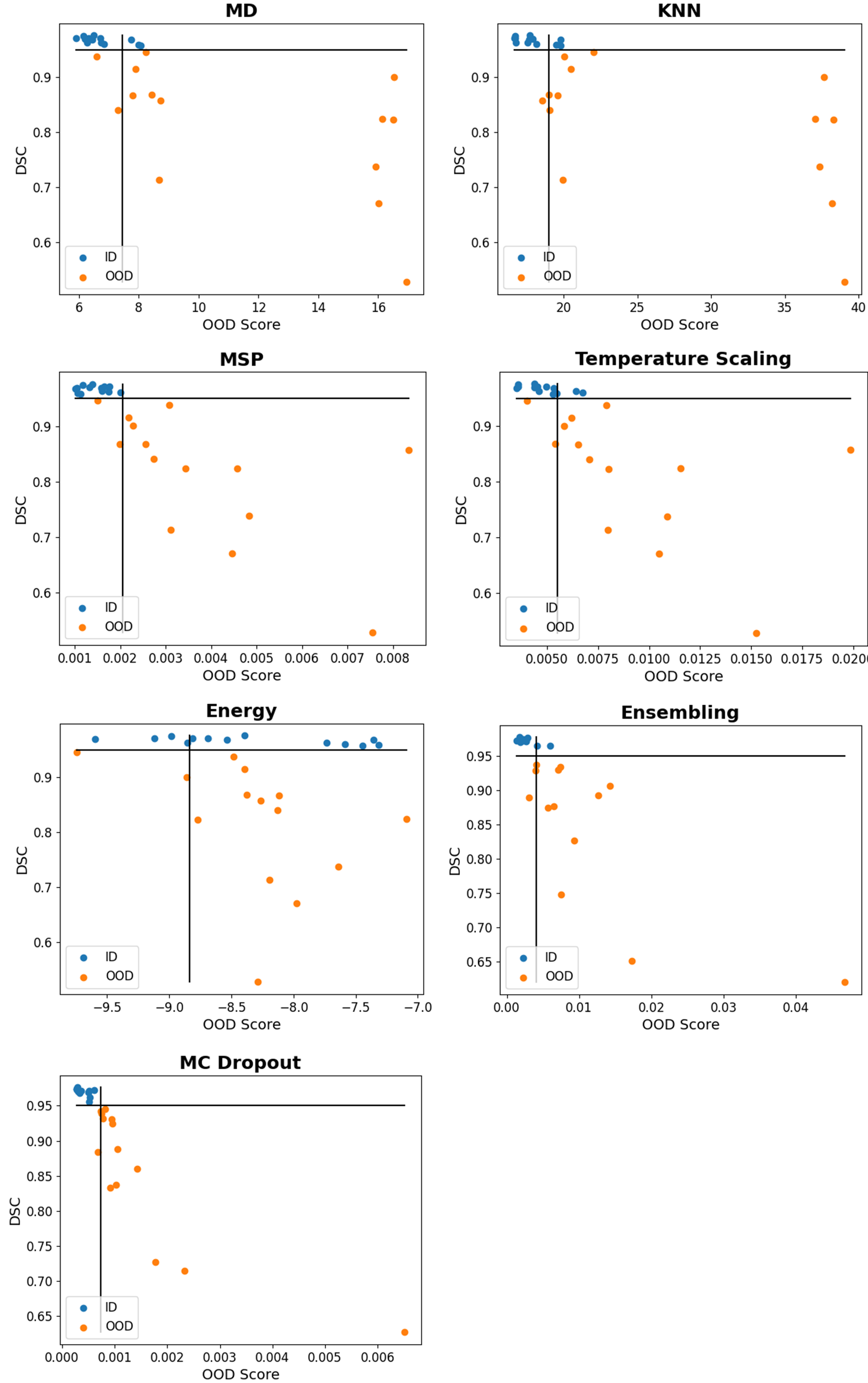}
    \label{fig:dsc95}
\end{figure}

Figure \ref{fig:dsc95} plots OOD scores against DSCs.
By moving the horizontal line vertically, one can visualize how the OOD detection performance would change if the DSC threshold were changed.
MSP, ensembling, and MC Dropout visually demonstrated the strongest negative linear relationship between OOD scores and DSC.
KNN and MD assigned noticeably higher OOD scores to six images with a wide range of DSCs.
These images came from the same patient who had a large tumor in the liver, resulting in missing liver segments (Figure \ref{fig:tumor}, Appendix D).
The training-based methods assigned the noticeably highest OOD score to a scan with an imaging artifact.
Instead of providing the intended further separation of softmax score distributions, temperature scaling and energy scoring visually pushed the distributions closer together.

\section{Discussion}

Our work provides several key takeaways.
First, MD is highly sensitive to the methodology used to reduce the feature space.
Past research reduced feature dimensionality with average pooling with fixed parameters \citep{Lee2018,Gonzalez2021}.
Our work demonstrates that this practice may not achieve the best results, considering average pooling was outperformed by PCA for the MRI data and UMAP for the CT data.
While PCA and UMAP demonstrated promise, the best dimensionality reduction technique and parameter configuration is likely dataset and architecture-dependent.
For example, while PCA and UMAP with only a few components performed well for liver segmentation, they may discard information important to the segmentation of smaller anatomical structures such as tumors.
Architecture components, such as automatic cropping on an image-by-image basis, may also affect the applicability of a technique.
Therefore, using a validation dataset to choose a dimensionality reduction technique and configuration could improve downstream OOD detection.
Our findings align with those of \cite{Ghosal2024}, whose work demonstrates that a careful and methodological selection of a subspace of features can improve feature-based OOD detection.
A fundamental difference between our approach and that of \cite{Gonzalez2021} is that we define OOD cases based on model performance.
This difference could explain the decline in MD performance observed with average pooling in our study.

Second, raw and reduced segmentation model features may not be Gaussian-distributed, challenging the suitability of MD to all OOD detection tasks.
This is especially true for medical imaging models, as the limited number of training images can lead to less clearly defined training distributions.
Moreover, the Gaussian assumption of MD fails to account for the potential multi-modality of medical imaging distributions arising from various factors such as differences across source datasets, acquisition parameters, contrast phases, disease states, artifacts, and stages of therapy.
In our work, a non-parametric approach, KNN, outperformed MD on raw features across all segmentation models and thresholds.
In addition, MD outperformed KNN on average pooled features from the MRI segmentation models across all thresholds.
Furthermore, our visualization of 2D embeddings highlighted significant gaps in the training distributions, with some training images positioned far from the central modes of the distributions.
Our findings corroborate those of \cite{Sun2022}, who demonstrated that parametric approaches are not always suitable for OOD detection.

Finally, the best OOD detection method for detecting poor segmentation performance may be task-dependent.
In \cite{Gonzalez2021}, MD outperformed MC Dropout and MSP when applied to lung lesion segmentation.
In \cite{GONZALEZ2022}, MD and MSP achieved perfect differentiation for hippocampus segmentation, outperforming MC Dropout.
For prostrate segmentation, on the other hand, MD and MC Dropout achieved perfect differentiation, outperforming MSP.
In our liver segmentation study, MC Dropout outperformed MD, with MC Dropout achieving perfect differentiation.
This suggests that MC Dropout may be better suited for tasks involving large and easily identifiable anatomical structures.
Furthermore, in our liver segmentation study, MSP marginally outperformed MD on the test datasets drawn from MD Anderson, whereas MD substantially outperformed MSP on the test dataset from Houston Methodist.
Moreover, on the MD Anderson test datasets, temperature scaling and energy scoring performed worse than MSP, a finding shared with \cite{Gonzalez2021,GONZALEZ2022}.
However, energy scoring achieved the best performance of the output-based methods on the test dataset from Houston Methodist.

This work has several limitations.
First, privacy must be considered when utilizing KNN, as the embeddings from all training images must be stored.
Second, this work solely focused on liver segmentation.
While this focus is advantageous for liver cancer research, the presented results may not extend to other anatomical structures.
Third, the OOD detection thresholds relied on DSC, whereas surface-based metrics may better estimate whether a contour is clinically acceptable \citep{Baroudi2023}.
Fourth, this work defined OOD detection based on model performance.
Although this definition is important to consider for patient safety, the presented results may not extend to other OOD definitions.
In addition, this definition caused the proportion of OOD test images to vary across models, limiting a direct comparison of OOD detection performance across these models.
Finally, due to the automatic cropping nature of the nnU-nets utilized, all nnU-net embeddings had to be average pooled across the dimension representing the number of patches. 
Therefore, the nnU-net results are not a true representation of the distances applied to raw, PCA-, t-SNE-, and UMAP-reduced features.

Our work has several potential applications.
First, a warning that the model likely failed could be added to automated segmentations with OOD scores above a specified threshold in a clinical setting. 
This would protect against automation bias, which would, in turn, protect patients whose scans have uncommon attributes.
Ensembling and MC Dropout may be well suited for a liver segmentation task if the computational resources are available due to their superior segmentation and OOD detection performance.
Second, detecting poor segmentation performance in retrospective studies where a large corpus of data is to be segmented.
As reviewing all autosegmentations would be infeasible, human evaluators would only need to review the autosegmentations associated with large OOD scores.
In this application, computational costs may outweigh performance.
Accordingly, KNN may be advantageous to utilize.
Third, the dimensionality techniques could provide a visualization tool for segmentation model creators to analyze how their model views their data.
For example, using PCA with two components highlighted the images of low perceptual quality in the AMOS dataset.
Lastly, this work could be used to diversify institutional training datasets by determining which images have the most utility to label.
The OOD scores of scans in unlabeled institutional databases would elucidate the most challenging cases and the cases that differ the most from the original training dataset.

This research provides several avenues for future work.
One of the biggest barriers to developing post-hoc OOD detection pipelines for medical imaging segmentation models is the number of choices one must consider when building their framework.
Considering only feature-based methods for a moment, one must determine if they are going to use the features directly \citep{Lee2018}, a spectral analysis of the features \citep{Karimi2023}, or pairwise feature correlations with Gram matrices \citep{Sastry2020}.
Then there are questions of which features should be used \cite{GONZALEZ2022,Anthony2023}, and if multiple are used, how to best aggregate them \citep{Lee2018}.
Once the features are chosen, one must determine how to properly reduce them to satisfy computational requirements and optimize performance \citep{Woodland2023,Ghosal2024}.
At this point, one should consider if parametric or non-parametric distances are the most appropriate for the reduced features \citep{Sun2022}.
These considerations open up a plethora of avenues for future work.
Research regarding each of these factors is of benefit to the field, in addition to large-scale application studies that demonstrate superior configurations for specific scenarios.
However, what the field is most lacking is the collaborative infrastructure to automate these decision processes for specific models and validation datasets.
In reality, the best configurations are most likely task-dependent, and most hospital systems developing segmentation models do not have the resources to perform such exhaustive searches.

\section{Conclusion}

In this work, MD was applied to dimensionality-reduced bottleneck features of a Swin UNETR trained for liver segmentation on T1-weighted MRIs. 
The resulting pipeline was able to embed entire 3D medical images into several components. 
These components were not only sufficient to cluster datasets drawn from different institutions but also could detect scans that the model performed poorly on with high performance and minimal computational cost (less than one second on CPUs).
We validated our methods on previously trained liver segmentation models and found that either PCA or UMAP improved performance over average pooling for all models.
Furthermore, we applied KNN to all models post hoc and found that it drastically outperformed the MD on raw and average pooled features: on a nnU-net trained on liver MRIs, it increased the AUROC to 96\% from 69\% and decreased the amount of time required to compute OOD scores for 352 MRIs from an hour to 7 seconds.

\acks{Research reported in this publication was supported in part by the Tumor Measurement Initiative through the MD Anderson Strategic Initiative Development Program (STRIDE), the Helen Black Image Guided Fund, the Image Guided Cancer Therapy Research Program at The University of Texas MD Anderson Cancer Center, a generous gift from the Apache Corporation, and the National Cancer Institute of the National Institutes of Health under award numbers R01CA221971, R01CA235564, R01CA195524, and P30CA016672. 
We'd like to thank Dr. Eugene J. Koay, Ph.D., for providing the liver MRI data from MD Anderson and Sarah Bronson - Scientific Editor at the Research Medical Library at MD Anderson - for editing sections of this article.}

\ethics{The work follows appropriate ethical standards in conducting research and writing the manuscript, following all applicable laws and regulations regarding the treatment of human subjects.
This retrospective study was approved by The University of Texas MD Anderson Cancer Center Institutional Review Board (PA18-0832). 
The requirement for written informed consent was waived for this retrospective analysis.}

\coi{We declare we have no conflicts of interest.}

\section*{Data Availability}
    The DLDS dataset is available at \url{https://zenodo.org/records/7774566}.
    The AMOS dataset can be accessed at \url{https://zenodo.org/records/7155725}.
    The CHAOS dataset can be downloaded from \url{https://zenodo.org/records/3431873}.
    The ATLAS dataset is hosted at \url{https://atlas-challenge.u-bourgogne.fr/}.
    The BTCV challenge dataset is available at \url{https://www.synapse.org/Synapse:syn3193805/wiki/217789}.
    The MD Anderson data may be made available upon request in compliance with institutional IRB requirements.

\section*{Code Availability}

Our code can be found at \url{https://github.com/mckellwoodland/dimen\_reduce\_mahal} \citep{Woodland2024}.

\bibliography{sample}

\clearpage
\appendix

\section{OOD detection results split by threshold}
\label{sec:thres}

\subsection{Mahalanobis Distance}
\label{sec:thres_md}

\begin{table}[h]
    \centering
    \caption{MD-based OOD detection of poor performance results.
    ID is a DSC $\geq$ 95\%.
    Only the best-performing configuration by AUROC is reported for each dimensionality reduction technique (Reduct): no reduction (None), PCA($n_p$) with $n_p$ components, t-SNE, UMAP($n_u$) with $n_u$ components, and $n_d$-dimensional average pooling with kernel size $k$ and stride $s$, Pool$n_d$D($k$,$s$).
    Seconds is the amount of time it took to calculate the test distances.
    The results are averages ($\pm$SD) across 5 runs.
    Arrows denote whether higher or lower is better.
    Bold highlights the best performance per model.
    }\label{tab:results}
    \begin{tabularx}{\textwidth}{|X|l|c|c|c|c|}
    \cline{1-6}
    \textbf{Model} & \textbf{Reduct} & \textbf{AUROC} $\uparrow$ & \textbf{AUPRC} $\uparrow$ & \textbf{FPR90} $\downarrow$ &\textbf{Seconds} $\downarrow$\\
    \cline{1-6}
    \multirow{6}{=}{\parbox{2cm}{MRI \\UNETR}}&None & 0.48 {\small ($\pm$0.00)}& 0.61 {\small ($\pm$0.00)} & 1.00 {\small ($\pm$0.00)} & 9,354.34 {\small ($\pm$48.53)}\\
    \cline{2-6}
    & PCA(256) & \textbf{0.93} {\small ($\pm$0.00)} & \textbf{0.94} {\small ($\pm$0.00)} & \textbf{0.23} {\small ($\pm$0.00)} & \textbf{2.82} {\small ($\pm$0.14)} \\
    \cline{2-6}
    & t-SNE & 0.70 {\small ($\pm$0.08)} & 0.72 {\small ($\pm$0.12)} & 0.71 {\small ($\pm$0.14)} & 4.70 {\small ($\pm$0.28)}\\
    \cline{2-6}
    & UMAP(2) & 0.77 {\small ($\pm$0.08)} & 0.79 {\small ($\pm$0.11)} & 0.57 {\small ($\pm$0.08)} & 10.44 {\small ($\pm$0.22)}\\
    \cline{2-6}
    & Pool2D(3,2)& 0.82 {\small ($\pm$0.00)} & 0.86 {\small ($\pm$0.00)} & 0.46 {\small ($\pm$0.00)} & 15.32 {\small ($\pm$8.92)}\\
    \cline{1-6}
    \multirow{6}{=}{\parbox{2cm}{MRI+ \\ UNETR}}& None& 0.46 {\small($\pm$0.00)} & \textbf{1.00} {\small($\pm$0.00)} & 1.00 {\small($\pm$0.00)} & 9809.40 {\small($\pm$57.20)}\\
    \cline{2-6}
    & PCA(16)& 0.82 {\small($\pm$0.00)} & \textbf{1.00} {\small($\pm$0.00)} & 1.00 {\small($\pm$0.00)} & \textbf{1.42} {\small($\pm$0.06)}\\
    \cline{2-6}
    & t-SNE& 0.92 {\small($\pm$0.00)} & \textbf{1.00} {\small($\pm$0.00)} & \textbf{0.00} {\small($\pm$0.00)} & 5.75 {\small($\pm$0.16)}\\
    \cline{2-6}
    & UMAP(16)& 0.91 {\small($\pm$0.03)} & \textbf{1.00} {\small($\pm$0.00)} & 0.20 {\small($\pm$0.40)} & 16.69 {\small(0.98)}\\
    \cline{2-6}
    & Pool3D(3,1)& \textbf{0.96} {\small($\pm$0.00)} & \textbf{1.00} {\small($\pm$0.00)} & \textbf{0.00} {\small($\pm$0.00)} & 78.77 {\small($\pm$0.20)} \\
    \cline{1-6}
    \multirow{6}{=}{\parbox{2cm}{MRI+ \\ nnU-net}}&None&0.69 {\small ($\pm$0.00)}&\textbf{1.00} {\small ($\pm$0.00)}&0.67 {\small ($\pm$0.00)}&4,125.96 {\small ($\pm$13.12)}\\
    \cline{2-6}
    & PCA(8)&\textbf{0.96} {\small ($\pm$0.00)}&\textbf{1.00} {\small ($\pm$0.00)}& \textbf{0.00} {\small ($\pm$0.00)} & \textbf{1.12} {\small ($\pm$0.04)}\\
    \cline{2-6}
    & t-SNE&0.70 {\small ($\pm$0.18)}&\textbf{1.00} {\small ($\pm$0.00)}&0.87 {\small ($\pm$0.16)}&4.72 {\small ($\pm$0.11)}\\
    \cline{2-6}
    & UMAP(16)&0.82 {\small ($\pm$0.08)}&\textbf{1.00} {\small ($\pm$0.00)}& 0.67 {\small ($\pm$0.03)}&19.07 {\small ($\pm$0.96)}\\
    \cline{2-6}
    & Pool2D(2,1)&0.85 {\small ($\pm$0.00)} &\textbf{1.00} {\small ($\pm$0.00)} &0.67 {\small ($\pm$0.00)}&1,579.73 {\small ($\pm$52.41)}\\
    \cline{1-6}
    \multirow{6}{=}{\parbox{2cm}{CT \\ nnU-net}}&None& 0.41 {\small ($\pm$0.00)} & 0.10 {\small ($\pm$0.00)} & 0.91 {\small ($\pm$0.00)} & 5,856.21 {\small ($\pm$63.04)}\\
    \cline{2-6}
    & PCA(32) & 0.56 {\small ($\pm$0.00)} & 0.17 {\small ($\pm$0.00)} & 0.92 {\small ($\pm$0.00)} & \textbf{8.17} {\small ($\pm$0.23)}\\
    \cline{2-6}
    & t-SNE& 0.59 {\small ($\pm$0.04)} & 0.20 {\small ($\pm$0.02)} & 0.80 {\small ($\pm$0.02)} & 13.75 {\small ($\pm$0.35)}\\
    \cline{2-6}
    & UMAP(128) & \textbf{0.68} {\small ($\pm$0.03)} & \textbf{0.22} {\small ($\pm$0.01)} & \textbf{0.74} {\small ($\pm$0.09)} & 288.42 {\small ($\pm$25.99)} \\
    \cline{2-6}
    & Pool2D(2,2)& 0.59 {\small ($\pm$0.00)} & 0.13 {\small ($\pm$0.00)} & 0.84 {\small ($\pm$0.00)} & 163.84 {\small ($\pm$19.54)}\\
    \cline{1-6}
    \end{tabularx}
\end{table}

\begin{table}[h]
    \centering
    \caption{MD-based OOD detection of poor performance results.
    ID is a DSC $\geq$ 80\%.
    Only the best-performing configuration by AUROC is reported for each dimensionality reduction technique (Reduct): no reduction (None), PCA($n_p$) with $n_p$ components, t-SNE, UMAP($n_u$) with $n_u$ components, and $n_d$-dimensional average pooling with kernel size $k$ and stride $s$, Pool$n_d$D($k$,$s$).
    Seconds is the amount of time it took to calculate the test distances.
    The results are averages ($\pm$SD) across 5 runs.
    Arrows denote whether higher or lower is better.
    Bold highlights the best performance per model.
    }\label{tab:results}
    \begin{tabularx}{\textwidth}{|X|l|c|c|c|c|}
    \cline{1-6}
    \textbf{Model} & \textbf{Reduct} & \textbf{AUROC} $\uparrow$ & \textbf{AUPRC} $\uparrow$ & \textbf{FPR90} $\downarrow$ &\textbf{Seconds} $\downarrow$\\
    \cline{1-6}
    \multirow{6}{=}{\parbox{2cm}{MRI \\UNETR}}&None & 0.48 {\small($\pm$0.00)} & 0.21 {\small($\pm$0.00)}& 0.83 {\small($\pm$0.00)} & 9349.73 {\small($\pm$10.85)}\\
    \cline{2-6}
    & PCA(128) & 0.92 {\small($\pm$0.00)} & 0.68 {\small($\pm$0.00)} & \textbf{0.13} {\small($\pm$0.00)}& 1.95 {\small($\pm$0.12)}\\
    \cline{2-6}
    & t-SNE & 0.85 {\small($\pm$0.00)} & 0.49 {\small($\pm$0.00)} & 0.26 {\small($\pm$0.00)} & 4.35 {\small($\pm$0.13)}\\
    \cline{2-6}
    & UMAP(8) & \textbf{0.93} {\small($\pm$0.03)} & \textbf{0.73} {\small($\pm$0.07)} & \textbf{0.13} {\small($\pm$0.00)} & 5.61 {\small($\pm$0.14)}\\
    \cline{2-6}
    & Pool3D(4,1)& 0.87 {\small($\pm$0.00)} &  0.50 {\small($\pm$0.00)} & 0.22 {\small($\pm$0.00)} & \textbf{1.72} {\small($\pm$0.06)}\\
    \cline{1-6}
    \multirow{6}{=}{\parbox{2cm}{MRI+ \\ UNETR}}& None&0.53 {\small ($\pm$0.00)}&0.03 {\small ($\pm$0.00)}& 0.78 {\small ($\pm$0.00)} & 10,070.78 {\small ($\pm$141.69)}\\
    \cline{2-6}
    & PCA(32)& \textbf{0.85} {\small ($\pm$0.01)} & \textbf{0.13} {\small ($\pm$0.00)} & \textbf{0.34} {\small ($\pm$0.02)} & \textbf{1.86} {\small ($\pm$0.32)}\\
    \cline{2-6}
    & t-SNE& 0.66 {\small ($\pm$0.00)} & 0.03 {\small ($\pm$0.00)} & 0.45 {\small ($\pm$0.00)} & 5.77 {\small ($\pm$0.06)}\\
    \cline{2-6}
    & UMAP(2)& 0.68 {\small ($\pm$0.07)} & 0.05 {\small ($\pm$0.01)} & 0.49 {\small ($\pm$0.06)} & 21.29 {\small ($\pm$0.43)}\\
    \cline{2-6}
    & Pool2D(4,1) & 0.64 {\small ($\pm$0.00)} & 0.04 {\small ($\pm$0.00)} & 0.64 {\small ($\pm$0.00)} & 16.08 {\small ($\pm$13.41)}\\
    \cline{1-6}
    \multirow{6}{=}{\parbox{2cm}{MRI+ \\ nnU-net}}&None& 0.58 {\small($\pm$0.00)} & 0.03 {\small($\pm$0.00)} & 0.70 {\small($\pm$0.00)} & 4,100.82 {\small($\pm$4.00)}\\
    \cline{2-6}
    & PCA(8)& 0.74 {\small($\pm$0.00)} & 0.08 {\small($\pm$0.00)} & 0.52 {\small($\pm$0.00)} & \textbf{1.17} {\small($\pm$0.04)}\\
    \cline{2-6}
    & t-SNE& 0.35 {\small($\pm$0.00)} & 0.02 {\small($\pm$0.00)} & 0.90 {\small($\pm$0.00)} & 4.68 {\small($\pm$0.06)} \\
    \cline{2-6}
    & UMAP(4)& \textbf{0.78} {\small($\pm$0.00)} & 0.06 {\small($\pm$0.02)} & \textbf{0.45} {\small($\pm$0.07)} & 13.43 {\small($\pm$0.24)}\\
    \cline{2-6}
    & Pool3D(2,2)& 0.68 {\small($\pm$0.00)} & \textbf{0.30} {\small($\pm$0.00)} & 0.71 {\small($\pm$0.00)} & 6.99 {\small($\pm$0.18)} \\
    \cline{1-6}
    \multirow{6}{=}{\parbox{2cm}{CT \\ nnU-net}}&None& - & - & - & - \\
    \cline{2-6}
    & PCA & - & - & - & - \\
    \cline{2-6}
    & t-SNE& - & - & - & - \\
    \cline{2-6}
    & UMAP & - & - & - & - \\
    \cline{2-6}
    & Pool& - & - & - & - \\
    \cline{1-6}
    \end{tabularx}
\end{table}

\begin{table}[h]
    \centering
    \caption{MD-based OOD detection of poor performance results.
    ID is a DSC $\geq$ the median DSC.
    Only the best-performing configuration by AUROC is reported for each dimensionality reduction technique (Reduct): no reduction (None), PCA($n_p$) with $n_p$ components, t-SNE, UMAP($n_u$) with $n_u$ components, and $n_d$-dimensional average pooling with kernel size $k$ and stride $s$, Pool$n_d$D($k$,$s$).
    Seconds is the amount of time it took to calculate the test distances.
    The results are averages ($\pm$SD) across 5 runs.
    Arrows denote whether higher or lower is better.
    Bold highlights the best performance per model.
    }\label{tab:results}
    \begin{tabularx}{\textwidth}{|X|l|c|c|c|c|}
    \cline{1-6}
    \textbf{Model} & \textbf{Reduct} & \textbf{AUROC} $\uparrow$ & \textbf{AUPRC} $\uparrow$ & \textbf{FPR90} $\downarrow$ &\textbf{Seconds} $\downarrow$\\
    \cline{1-6}
    \multirow{6}{=}{\parbox{2cm}{MRI \\UNETR}}&None & 0.48 {\small($\pm$0.00)}& 0.61 {\small($\pm$0.00)} & 1.00 {\small($\pm$0.00)} & 9,283.76 {\small($\pm$8.18)} \\
    \cline{2-6}
    & PCA(256) & \textbf{0.93} {\small($\pm$0.00)} & \textbf{0.94} {\small($\pm$0.00)} & \textbf{0.23} {\small($\pm$0.00)} & \textbf{2.93} {\small($\pm$0.14)} \\
    \cline{2-6}
    & t-SNE & 0.60 {\small($\pm$0.00)} & 0.57 {\small($\pm$0.00)} & 0.77 {\small($\pm$0.00)} & 4.42 {\small($\pm$0.09)}\\
    \cline{2-6}
    & UMAP(8) & 0.74 {\small($\pm$0.05)} & 0.83 {\small($\pm$0.04)} & 0.74 {\small($\pm$0.10)} & 5.79 {\small($\pm$0.22)}\\
    \cline{2-6}
    & Pool2D(3,2)& 0.82 {\small($\pm$0.00)} & 0.86 {\small($\pm$0.00)} & 0.46 {\small($\pm$0.00)}& 4.70 {\small($\pm$0.14)}\\
    \cline{1-6}
    \multirow{6}{=}{\parbox{2cm}{MRI+ \\ UNETR}}& None& 0.50 {\small($\pm$0.00)} & 0.52 {\small($\pm$0.00)} & 0.90 {\small($\pm$0.00)} & 9,777.69 {\small($\pm$12.41)}\\
    \cline{2-6}
    & PCA(32)& 0.54 {\small($\pm$0.00)} & 0.58 {\small($\pm$0.00)} & 0.86 {\small($\pm$0.00)} & \textbf{1.66} {\small($\pm$0.15)}\\
    \cline{2-6}
    & t-SNE& 0.52 {\small($\pm$0.00)} & 0.54 {\small($\pm$0.00)} & 0.90 {\small($\pm$0.00)} & 5.73 {\small($\pm$0.16)}\\
    \cline{2-6}
    & UMAP(32)& 0.55 {\small($\pm$0.00)} & 0.55 {\small($\pm$0.01)} & \textbf{0.84} {\small($\pm$0.02)} & 17.01 {\small($\pm$0.57)}\\
    \cline{2-6}
    & Pool2D(4,1)& \textbf{0.57} {\small($\pm$0.00)} & \textbf{0.61} {\small($\pm$0.00)} & \textbf{0.84} {\small($\pm$0.00)} & 9.38 {\small($\pm$0.14)}\\
    \cline{1-6}
    \multirow{6}{=}{\parbox{2cm}{MRI+ \\ nnU-net}}&None& 0.51 {\small($\pm$0.00)} & 0.51 {\small($\pm$0.00)} & 0.88 {\small($\pm$0.00)} & 4,091.49 {\small($\pm$5.61)}\\
    \cline{2-6}
    & PCA(32) & \textbf{0.62} {\small($\pm$0.00)} & \textbf{0.62} {\small($\pm$0.00)} & \textbf{0.80} {\small($\pm$0.00)} & \textbf{1.39} {\small($\pm$0.09)} \\
    \cline{2-6}
    & t-SNE& 0.59 {\small($\pm$0.00)} & 0.60 {\small($\pm$0.00)} & 0.82 {\small($\pm$0.00)} & 4.65 {\small($\pm$0.07)} \\
    \cline{2-6}
    & UMAP(64) & 0.58 {\small($\pm$0.02)} & 0.58 {\small($\pm$0.01)} & 0.86 {\small($\pm$0.03)} & 14.16 {\small($\pm$0.16)} \\
    \cline{2-6}
    & Pool3D(2,2)& 0.57 {\small($\pm$0.00)} & 0.58 {\small($\pm$0.00)} & 0.86 {\small($\pm$0.00)} & 6.86 {\small($\pm$0.11)}\\
    \cline{1-6}
    \multirow{6}{=}{\parbox{2cm}{CT \\ nnU-net}}&None& 0.49 {\small($\pm$0.00)} & 0.49 {\small($\pm$0.00)} & 0.90 {\small($\pm$0.00)} & 5,388.67 {\small($\pm$47.96)}\\
    \cline{2-6}
    & PCA(2) & 0.56 {\small($\pm$0.00)} & 0.56 {\small($\pm$0.00)} & 0.88 {\small($\pm$0.00)} & \textbf{4.79} {\small($\pm$0.06)} \\
    \cline{2-6}
    & t-SNE& \textbf{0.63} {\small($\pm$0.00)} & \textbf{0.66} {\small($\pm$0.00)} & \textbf{0.85} {\small($\pm$0.00)} & 12.09 {\small($\pm$0.17)} \\
    \cline{2-6}
    & UMAP(32) & 0.57 {\small($\pm$0.03)} & 0.58 {\small($\pm$0.03)} & \textbf{0.85} {\small($\pm$0.04)} & 199.15 {\small($\pm$1.56)} \\
    \cline{2-6}
    & Pool2D(4,1)& 0.57 {\small($\pm$0.00)} & 0.58 {\small($\pm$0.00)} & \textbf{0.85} {\small($\pm$0.00)} & 130.59 {\small($\pm$1.80)}\\
    \cline{1-6}
    \end{tabularx}
\end{table}

\clearpage

\subsection{K-Nearest Neighbors}
\label{sec:thres_knn}

\begin{table}[h]
    \centering
    \caption{KNN-based OOD detection of poor performance results.
    ID is a DSC $\geq$ 95\%.
    Only the best-performing configuration by AUROC is reported for each dimensionality reduction technique (Reduct): no reduction (None), PCA($n_p$) with $n_p$ components, t-SNE, UMAP($n_u$) with $n_u$ components, and $n_d$-dimensional average pooling with kernel size $k$ and stride $s$, Pool$n_d$D($k$,$s$).
    Seconds is the amount of time it took to calculate the test distances.
    The results are averages ($\pm$SD) across 5 runs.
    Arrows denote whether higher or lower is better.
    Bold highlights the best performance per model.
    }\label{tab:results}
    \begin{tabularx}{\textwidth}{|X|l|c|c|c|c|c|}
    \cline{1-7}
    \textbf{Model} & \textbf{Reduction} & \textbf{K} & \textbf{AUROC} $\uparrow$ & \textbf{AUPRC} $\uparrow$ & \textbf{FPR90} $\downarrow$ &\textbf{Seconds} $\downarrow$\\
    \cline{1-7}
    \multirow{6}{=}{\parbox{2cm}{MRI \\UNETR}}&None& 256 & 0.87 {\small ($\pm$0.00)} & {\small 0.88 ($\pm$0.00)} & 0.31 {\small ($\pm$0.00)} & \textbf{0.78} {\small ($\pm$0.00)}\\
    \cline{2-7}
    & PCA(2) & 256 & 0.90 {\small ($\pm$0.00)} & 0.92 {\small ($\pm$0.00)} & 0.31 {\small ($\pm$0.00)} & 0.95 {\small ($\pm$0.00)}\\
    \cline{2-7}
    & t-SNE & 256 & 0.77 {\small ($\pm$0.05)} & 0.83 {\small ($\pm$0.04)} & 0.74 {\small ($\pm$0.06)} & 4.48 {\small ($\pm$0.07)} \\
    \cline{2-7}
    & UMAP(32) & 256 & 0.83 {\small ($\pm$0.05)} & 0.85 {\small ($\pm$0.04)} & 0.51 {\small ($\pm$0.08)} & 6.70 {\small ($\pm$0.15)}\\
    \cline{2-7}
    & Pool2D(3,1) & 256 & \textbf{0.94} {\small ($\pm$0.00)} & \textbf{0.95} {\small ($\pm$0.00)} & \textbf{0.23} {\small ($\pm$0.00)} & 0.90 {\small ($\pm$0.00) }\\
    \cline{1-7}
    \multirow{6}{=}{\parbox{2cm}{MRI+ \\ UNETR}}& None& 32 & 0.86 {\small($\pm$0.00)} & \textbf{1.00} {\small($\pm$0.00)} & 1.00 {\small($\pm$0.00)} & 9.27 {\small($\pm$0.15)}\\
    \cline{2-7}
    & PCA(16) & 8 & 0.88 {\small($\pm$0.02)} & \textbf{1.00} {\small($\pm$0.00)} & 0.40 {\small($\pm$0.49)} & 1.52 {\small($\pm$0.06)}\\
    \cline{2-7}
    & t-SNE& 256 & 0.94 {\small($\pm$0.00)} & \textbf{1.00} {\small($\pm$0.00)} & \textbf{0.00} {\small($\pm$0.00)} & 5.67 {\small($\pm$0.08)}\\
    \cline{2-7}
    & UMAP(2) & 64 & 0.89 {\small($\pm$0.02)} & \textbf{1.00} {\small($\pm$0.00)} & 0.60 {\small($\pm$0.49)} & 16.47 {\small($\pm$0.28)}\\
    \cline{2-7}
    & Pool2D(3,2) & 4 & \textbf{0.97} {\small($\pm$0.00)} & \textbf{1.00} {\small($\pm$0.00)} & \textbf{0.00} {\small($\pm$0.00)} & \textbf{1.30} {\small($\pm$0.03)}\\
    \cline{1-7}
    \multirow{6}{=}{\parbox{2cm}{MRI+ \\ nnU-net}}&None& 256 & 0.96 {\small ($\pm$0.00)} & \textbf{1.00} {\small ($\pm$0.00)} & \textbf{0.00} {\small($\pm$0.00)}& 6.78 {\small($\pm$0.10)}\\
    \cline{2-7}
    & PCA(8) & 256 & 0.97 {\small($\pm$0.00)}& \textbf{1.00} {\small($\pm$0.00)}&\textbf{0.00} {\small($\pm$0.00)}& 1.16 {\small($\pm$0.07)}\\
    \cline{2-7}
    & t-SNE & 128 & 0.67 {\small($\pm$0.08)}&\textbf{1.00} {\small($\pm$0.00)}&1.00 {\small($\pm$0.00)}& 4.76 {\small ($\pm$0.11)}\\
    \cline{2-7}
    & UMAP(2) & 2 & 0.96 {\small($\pm$0.04)}&\textbf{1.00} {\small($\pm$0.00)}&0.80 {\small($\pm$0.27)}&14.18 {\small($\pm$0.55)}\\
    \cline{2-7}
    & Pool2D(2,2)& 256 & \textbf{0.98} {\small ($\pm$0.00)} & \textbf{1.00} {\small ($\pm$0.00)} & \textbf{0.00} {\small ($\pm$0.00)} & \textbf{0.74} {\small ($\pm$0.04)}\\
    \cline{1-7}
    \multirow{6}{=}{\parbox{2cm}{CT \\ nnU-net}}&None& 8 & 0.52 {\small($\pm$0.00)}& 0.13 {\small($\pm$0.00)}& 0.94 {\small($\pm$0.00)}& 37.64 {\small($\pm$0.67)}\\
    \cline{2-7}
    & PCA(8) & 4 & 0.55 {\small($\pm$0.00)}& 0.15 {\small($\pm$0.00)}& 0.97 {\small($\pm$0.00)}& \textbf{4.94} {\small($\pm$0.04)}\\
    \cline{2-7}
    & t-SNE& 256 & 0.46 {\small($\pm$0.00)}& 0.19 {\small($\pm$0.00)}&0.95 {\small($\pm$0.01)}& 12.28 {\small($\pm$0.09)}\\
    \cline{2-7}
    & UMAP(4) & 256 & \textbf{0.65} {\small($\pm$0.01)}& \textbf{0.24} {\small($\pm$0.05)}& \textbf{0.88} {\small($\pm$0.02)}& 199.93 {\small($\pm$1.96)}\\
    \cline{2-7}
    & Pool3D(2,2)& 4 & 0.54 {\small($\pm$0.00)}& 0.14 {\small($\pm$0.00)}& 0.96 {\small($\pm$0.00)}& 81.96 {\small($\pm$26.79)}\\
    \cline{1-7}
    \end{tabularx}
\end{table}

\begin{table}[h]
    \centering
    \caption{KNN-based OOD detection of poor performance results.
    ID is a DSC $\geq$ 80\%.
    Only the best-performing configuration by AUROC is reported for each dimensionality reduction technique (Reduct): no reduction (None), PCA($n_p$) with $n_p$ components, t-SNE, UMAP($n_u$) with $n_u$ components, and $n_d$-dimensional average pooling with kernel size $k$ and stride $s$, Pool$n_d$D($k$,$s$).
    Seconds is the amount of time it took to calculate the test distances.
    The results are averages ($\pm$SD) across 5 runs.
    Arrows denote whether higher or lower is better.
    Bold highlights the best performance per model.
    }\label{tab:results}
    \begin{tabularx}{\textwidth}{|X|l|c|c|c|c|c|}
    \cline{1-7}
    \textbf{Model} & \textbf{Reduct} & \textbf{K} & \textbf{AUROC} $\uparrow$ & \textbf{AUPRC} $\uparrow$ & \textbf{FPR90} $\downarrow$ &\textbf{Seconds} $\downarrow$\\
    \cline{1-7}
    \multirow{6}{=}{\parbox{2cm}{MRI \\UNETR}}&None & 256 & 0.87 {\small($\pm$0.00)} & 0.62 {\small($\pm$0.00)} & \textbf{0.13} {\small($\pm$0.00)} & \textbf{0.79} {\small($\pm$0.02)}\\
    \cline{2-7}
    & PCA(2) & 128 & \textbf{0.92} {\small($\pm$0.00)} & 0.67 {\small($\pm$0.00)} & \textbf{0.13} {\small($\pm$0.00)} & 0.93 {\small($\pm$0.05)}\\
    \cline{2-7}
    & t-SNE & 128 & 0.83 {\small($\pm$0.09)} & 0.44 {\small($\pm$0.11)} & 0.31 {\small($\pm$0.17)} & 4.36 {\small($\pm$0.07)}\\
    \cline{2-7}
    & UMAP(2) & 256 & 0.90 {\small($\pm$0.05)} & 0.57 {\small($\pm$0.18)} & 0.16 {\small($\pm$0.08)} & 5.68 {\small($\pm$0.10)}\\
    \cline{2-7}
    & Pool2D(3,2)& 256 & \textbf{0.92} {\small($\pm$0.00)} & \textbf{0.69} {\small($\pm$0.00)} & \textbf{0.13} {\small($\pm$0.00)} & 0.84 {($\pm$0.08)}\\
    \cline{1-7}
    \multirow{6}{=}{\parbox{2cm}{MRI+ \\ UNETR}}&None& 256 & 0.76 {\small ($\pm$0.00)}& 0.25 {\small ($\pm$0.00)} & 0.59 {\small ($\pm$0.00)}& 9.44 {\small ($\pm$0.16)}\\
    \cline{2-7}
    & PCA(32)& 64 & 0.84 {\small ($\pm$0.01)}& 0.17 {\small ($\pm$0.02)}& 0.40 {\small ($\pm$0.01)}& \textbf{1.57} {\small ($\pm$0.06)}\\
    \cline{2-7}
    & t-SNE & 64 & 0.70 {\small ($\pm$0.00)} & 0.04 {\small ($\pm$0.00)} & \textbf{0.37} {\small ($\pm$0.00)}& 6.49 {\small ($\pm$0.19)}\\
    \cline{2-7}
    & UMAP(4)& 64 & 0.78 {\small($\pm$0.05)}& 0.09 {\small($\pm$0.06)}&0.42 {\small($\pm$0.07)}& 15.70 {\small($\pm$0.45)}\\
    \cline{2-7}
    & Pool3D(2,2)& 256 & \textbf{0.87} {\small($\pm$0.00)}& \textbf{0.33} {\small($\pm$0.00)}& 0.43 {\small($\pm$0.00)}& 11.21 {\small($\pm$7.66)}\\
    \cline{1-7}
    \multirow{6}{=}{\parbox{2cm}{MRI+ \\ nnU-net}}&None& 256 & 0.72 {\small($\pm$0.00)} & \textbf{0.09} {\small($\pm$0.00)} & 0.67 {\small($\pm$0.00)} & 6.44 {\small($\pm$0.09)}\\
    \cline{2-7}
    & PCA(8)& 256 & 0.81 {\small($\pm$0.00)} & 0.06 {\small($\pm$0.00)} & 0.36 {\small($\pm$0.00)} & 1.12 {\small($\pm$0.07)}\\
    \cline{2-7}
    & t-SNE& 2 & 0.74 {\small($\pm$0.10)} & 0.05 {\small($\pm$0.02)} & 0.44 {\small($\pm$0.20)} & 4.80 {\small($\pm$0.12)}\\
    \cline{2-7}
    & UMAP(128) & 64 & \textbf{0.82} {\small($\pm$0.00)} & 0.06 {\small($\pm$0.01)} & 0.39 {\small($\pm$0.02)} & 14.61 {\small($\pm$0.25)}\\
    \cline{2-7}
    & Pool3D(4,1)&64& 0.76 {\small($\pm$0.00)} & 0.08 {\small($\pm$0.00)} & \textbf{0.34} {\small($\pm$0.00)} & \textbf{0.77} {\small($\pm$0.06)} \\
    \cline{1-7}
    \multirow{6}{=}{\parbox{2cm}{CT \\ nnU-net}}&None&-& - & - & - & - \\
    \cline{2-7}
    & PCA &-& - & - & - & - \\
    \cline{2-7}
    & t-SNE&-& - & - & - & - \\
    \cline{2-7}
    & UMAP &-& - & - & - & - \\
    \cline{2-7}
    & Pool&-& - & - & - & - \\
    \cline{1-7}
    \end{tabularx}
\end{table}

\begin{table}[h]
    \centering
    \caption{KNN-based OOD detection of poor performance results.
    ID is a DSC $\geq$ the median DSC.
    Only the best-performing configuration by AUROC is reported for each dimensionality reduction technique (Reduct): no reduction (None), PCA($n_p$) with $n_p$ components, t-SNE, UMAP($n_u$) with $n_u$ components, and $n_d$-dimensional average pooling with kernel size $k$ and stride $s$, Pool$n_d$D($k$,$s$).
    Seconds is the amount of time it took to calculate the test distances.
    The results are averages ($\pm$SD) across 5 runs.
    Arrows denote whether higher or lower is better.
    Bold highlights the best performance per model.
    }\label{tab:results}
    \begin{tabularx}{\textwidth}{|X|l|c|c|c|c|c|}
    \cline{1-7}
    \textbf{Model} & \textbf{Reduct} & \textbf{K} & \textbf{AUROC} $\uparrow$ & \textbf{AUPRC} $\uparrow$ & \textbf{FPR90} $\downarrow$ &\textbf{Seconds} $\downarrow$\\
    \cline{1-7}
    \multirow{6}{=}{\parbox{2cm}{MRI \\UNETR}}& None & 256 & 0.87 {\small($\pm$0.00)} & 0.88 {\small($\pm$0.00)} & 0.31 {\small($\pm$0.00)} & \textbf{0.78} {\small($\pm$0.00)} \\
    \cline{2-7}
    & PCA(2) & 256 & 0.90 {\small($\pm$0.00)} & 0.92 {\small($\pm$0.00)} & 0.31 {\small($\pm$0.00)} & 0.98 {\small($\pm$0.08)} \\
    \cline{2-7}
    & t-SNE & 256 & 0.77 {\small($\pm$0.05)} & 0.83 {\small($\pm$0.04)} & 0.74 {\small($\pm$0.06)} & 4.37 {\small($\pm$0.11)}\\
    \cline{2-7}
    & UMAP(128) & 256 & 0.82 {\small($\pm$0.04)} & 0.86 {\small($\pm$0.04)} & 0.63 {\small($\pm$0.16)} & 6.61 {\small($\pm$0.43)} \\
    \cline{2-7}
    & Pool2D(3,1)& 256 & \textbf{0.94} {\small($\pm$0.00)} & \textbf{0.95} {\small($\pm$0.00)} & \textbf{0.23} {\small($\pm$0.00)} & 0.89 {\small($\pm$0.01)}\\
    \cline{1-7}
    \multirow{6}{=}{\parbox{2cm}{MRI+ \\ UNETR}}& None & 32 & 0.86 {\small($\pm$0.00)} & \textbf{1.00} {\small($\pm$0.00)} & 1.00 {\small($\pm$0.00)} & 9.01 {\small($\pm$0.07)} \\
    \cline{2-7}
    & PCA(16) & 8 & 0.88 {\small($\pm$0.02)} & \textbf{1.00} {\small($\pm$0.00)} & 0.40 {\small($\pm$0.49)} & \textbf{1.46} {\small($\pm$0.09)}\\
    \cline{2-7}
    & t-SNE& 256 & 0.94 {\small($\pm$0.00)} & \textbf{1.00} {\small($\pm$0.00)} & \textbf{0.00} {\small($\pm$0.00)}& 5.67 {\small($\pm$0.04)}\\
    \cline{2-7}
    & UMAP(2) & 64 & 0.90 {\small($\pm$0.02)} & \textbf{1.00} {\small($\pm$0.00)} & 0.40 {\small($\pm$0.49)} & 16.75 {\small($\pm$0.26)}\\
    \cline{2-7}
    & Pool2D(3,2) & 4 & \textbf{0.97} {\small($\pm$0.00)} & \textbf{1.00} {\small($\pm$0.00)} & \textbf{0.00} {\small($\pm$0.00)} & 1.29 {\small($\pm$0.01)}\\
    \cline{1-7}
    \multirow{6}{=}{\parbox{2cm}{MRI+ \\ nnU-net}}&None&256& 0.96 {\small($\pm$0.00)} & \textbf{1.00} {\small($\pm$0.00)} & \textbf{0.00} {\small($\pm$0.00)} & 6.76 {\small($\pm$0.00)} \\
    \cline{2-7}
    & PCA(8)&256& 0.97 {\small($\pm$0.00)} & \textbf{1.00} {\small($\pm$0.00)} & \textbf{0.00} {\small($\pm$0.00)} & 1.12 {\small($\pm$0.04)} \\
    \cline{2-7}
    & t-SNE& 128& 0.67 {\small($\pm$0.09)} & \textbf{1.00} {\small($\pm$0.00)} & 1.00 {\small($\pm$0.00)} & 4.73 {\small($\pm$0.11)} \\
    \cline{2-7}
    & UMAP(2)&2& 0.96 {\small($\pm$0.02)} & \textbf{1.00} {\small($\pm$0.00)} & 0.20 {\small($\pm$0.16)} & 14.82 {\small($\pm$1.60)}\\
    \cline{2-7}
    & Pool2D(2,2)&256& \textbf{0.98} {\small($\pm$0.00)} & \textbf{1.00} {\small($\pm$0.00)} & \textbf{0.00} {\small($\pm$0.00)} & \textbf{0.77} {\small($\pm$0.03)}\\
    \cline{1-7}
    \multirow{6}{=}{\parbox{2cm}{CT \\ nnU-net}}&None&8& 0.52 {\small($\pm$0.00)} & 0.13 {\small($\pm$0.00)} & 0.94 {\small($\pm$0.00)} & 37.15 {\small($\pm$0.28)}\\
    \cline{2-7}
    & PCA(8) & 4& 0.55 {\small($\pm$0.00)} & 0.15 {\small($\pm$0.00)} & 0.97 {\small($\pm$0.00)} & \textbf{5.03} {\small($\pm$0.08)} \\
    \cline{2-7}
    & t-SNE& 256& 0.46 {\small($\pm$0.00)} & 0.19 {\small($\pm$0.00)} & 0.95 {\small($\pm$0.01)} & 12.07 {\small($\pm$0.12)} \\
    \cline{2-7}
    & UMAP(2) & 256& \textbf{0.66} {\small($\pm$0.01)} & \textbf{0.23} {\small($\pm$0.05)} & \textbf{0.88} {\small($\pm$0.02)} & 206.25 {\small($\pm$1.34)}\\
    \cline{2-7}
    & Pool3D(2,1) & 4 & 0.54 {\small($\pm$0.00)} & 0.14 {\small($\pm$0.00)} & 0.96 {\small($\pm$0.00)} & 23.84 {\small($\pm$0.15)}\\
    \cline{1-7}
    \end{tabularx}
\end{table}

\clearpage

\subsection{Comparison Methods}
\label{sec:thres_comp}

\begin{table}[h]
    \centering
    \caption{
    OOD detection results for the best-performing configurations of the comparison methods: MSP, temperature scaling (TS) and energy scoring (Energy) with temperature $T$, MC Dropout (MCD), and ensembling (Ensemble).
    ID is a DSC $\geq$ 95\%.
    Seconds is the amount of time it took to calculate the test distances.
    Only the best-performing configuration by AUROC is reported for each method.
    The results are averages ($\pm$SD) across 5 runs.
    Arrows denote whether a higher or lower value is better.
    Bold highlights the best performance per model.
    }
    \label{tab:more_results}
    \begin{tabularx}{\textwidth}{|p{1.5cm}|l|c|c|c|c|c|}
    \cline{1-7}
    \textbf{Model} & \textbf{Method} & \textbf{T} & \textbf{AUROC} $\uparrow$ & \textbf{AUPRC} $\uparrow$ & \textbf{FPR90} $\downarrow$ & \textbf{Seconds} $\downarrow$\\
    \cline{1-7}
    \multirow{3}{*}{\parbox{2cm}{MRI \\UNETR}}&MSP & - & \textbf{0.96} {\small($\pm$0.00)}& \textbf{0.97} {\small($\pm$0.00)}& \textbf{0.00} {\small($\pm$0.00)}& 8.78 {\small($\pm$0.11)}\\
    \cline{2-7}
    &TS & 2 & 0.90 {\small($\pm$0.00)} & 0.93 {\small($\pm$0.00)} & 0.15 {\small($\pm$0.00)}& 9.25 {\small($\pm$0.11)}\\
    \cline{2-7}
    &Energy & 3 & 0.55 {\small($\pm$0.00)} &0.57 {\small($\pm$0.00)}& 0.69 {\small($\pm$0.00)} & \textbf{7.08} {\small($\pm$0.14)} \\
    \cline{1-7}
    \parbox{2cm}{\vspace{.1cm}Dropout \\UNETR\vspace{.1cm}}& MCD & - &\textbf{1.00} {\small($\pm$0.00)} & \textbf{1.00} {\small($\pm$0.00)} & \textbf{0.00} {\small($\pm$0.00)} & \textbf{14.56} {\small($\pm$0.20)}\\
    \cline{1-7}
    \parbox{2cm}{\vspace{.1cm}Ensemble \\UNETR\vspace{.1cm}} & Ensemble & - & \textbf{0.96} {\small($\pm$0.00)} & \textbf{0.96} {\small($\pm$0.00)} & \textbf{0.14} {\small($\pm$0.00)} & \textbf{14.02} {\small($\pm$0.05)} \\
    \cline{1-7}
    \multirow{3}{*}{\parbox{2cm}{MRI+ \\UNETR}} & MSP & - & 0.91 {\small ($\pm$0.00)} & \textbf{1.00} {\small ($\pm$0.00)} & \textbf{0.00} {\small ($\pm$0.00)} & 59.18 {\small ($\pm$0.46)}\\
    \cline{2-7}
    &TS& 2 & \textbf{0.93} {\small($\pm$0.00)} & \textbf{1.00} {\small($\pm$0.00)} &  \textbf{0.00} {\small($\pm$0.00)}& 60.57 {\small($\pm$0.43)}\\
    \cline{2-7}
    &Energy & 1 & 0.89 {\small($\pm$0.00)} & \textbf{1.00} {\small($\pm$0.00)} & 1.00 {\small($\pm$0.00)} & \textbf{36.53} {\small($\pm$1.82)}\\  
    \cline{1-7}
    \multirow{3}{*}{\parbox{2cm}{MRI+ \\nnU-net}}& MSP & - & 0.45 {\small($\pm$0.00)} & 0.99 {\small($\pm$0.00)} & \textbf{1.00} {\small($\pm$0.00)} & 1,114.34 {\small($\pm$0.51)}\\
    \cline{2-7}
    &TS& 10 & 0.55 {\small($\pm$0.00)} & 0.99 {\small($\pm$0.00)} & \textbf{1.00} {\small($\pm$0.00)} & 1,252.78 {\small($\pm$0.73)}\\
    \cline{2-7}
    & Energy & 10 & \textbf{0.61} {\small($\pm$0.00)} & \textbf{1.00} {\small($\pm$0.00)} & \textbf{1.00} {\small($\pm$0.00)} & \textbf{186.88} {\small($\pm$0.47)}\\
    \cline{1-7}
    \multirow{3}{*}{\parbox{2cm}{CT \\nnU-net}}&MSP & -& \textbf{0.72} {\small($\pm$0.00)} & \textbf{0.29} {\small($\pm$0.00)} & \textbf{0.57} {\small($\pm$0.00)} & 568.51 {\small($\pm$0.37)}\\
    \cline{2-7}
    &TS& 3 & 0.68 {\small($\pm$0.00)} & 0.26 {\small($\pm$0.00)} & 0.69 {\small($\pm$0.00)} & 699.07 {\small($\pm$0.37)}\\
    \cline{2-7}
    &Energy & 2 & 0.67 {\small($\pm$0.00)} & 0.26 {\small($\pm$0.00)} & 0.75 {\small($\pm$0.00)} & \textbf{105.66} {\small($\pm$0.60)} \\   
    \cline{1-7}
    \end{tabularx}
\end{table}

\begin{table}[h]
    \centering
    \caption{
    OOD detection results for the best-performing configurations of the comparison methods: MSP, temperature scaling (TS) and energy scoring (Energy) with temperature $T$, MC Dropout (MCD), and ensembling (Ensemble).
    ID is a DSC $\geq$ 80\%.
    Seconds is the amount of time it took to calculate the test distances.
    The results are averages ($\pm$SD) across 5 runs.
    Arrows denote whether a higher or lower value is better.
    Bold highlights the best performance per model.
    }
    \label{tab:more_results}
    \begin{tabularx}{\textwidth}{|p{1.5cm}|l|c|c|c|c|c|}
    \cline{1-7}
    \textbf{Model} & \textbf{Method} & \textbf{T} & \textbf{AUROC} $\uparrow$ & \textbf{AUPRC} $\uparrow$ & \textbf{FPR90} $\downarrow$ & \textbf{Seconds} $\downarrow$\\
    \cline{1-7}
    \multirow{3}{*}{\parbox{2cm}{MRI \\UNETR}}& MSP & -&\textbf{0.92} {\small($\pm$0.00)} & \textbf{0.58} {\small($\pm$0.00)} & \textbf{0.09} {\small($\pm$0.00)} & 8.84 {\small($\pm$0.14)}\\
    \cline{2-7}
    &TS& 2& 0.91 {\small($\pm$0.00)} & 0.54 {\small($\pm$0.00)} & \textbf{0.09} {\small($\pm$0.00)} & 9.10 {\small($\pm$0.14)}\\
    \cline{2-7}
    &Energy & 1000& 0.72 {\small($\pm$0.00)} & 0.26 {\small($\pm$0.00)} & 0.35 {\small($\pm$0.00)} & \textbf{6.70} {\small($\pm$0.18)}\\
    \cline{1-7}
    \parbox{2cm}{\vspace{.1cm}MRI \\ Dropout\vspace{.1cm}}& MCD & - & \textbf{1.00} {\small($\pm$0.00)} & \textbf{1.00} {\small($\pm$0.00)} & \textbf{0.00} {\small($\pm$0.00)} & \textbf{14.70} {\small($\pm$0.23)}\\
    \cline{1-7}
    \parbox{2cm}{\vspace{.1cm}MRI \\ Ensemble\vspace{.1cm}} & Ensemble & - & \textbf{0.96} {\small($\pm$0.00)} & \textbf{0.83} {\small($\pm$0.00)} & \textbf{0.08} {\small($\pm$0.00)} & \textbf{14.16} {\small($\pm$0.20)} \\
    \cline{1-7}
    \multirow{3}{*}{\parbox{2cm}{MRI+ \\UNETR}} & MSP & - & 0.57 {\small($\pm$0.00)} & \textbf{0.09} {\small($\pm$0.00)} & \textbf{0.59} {\small($\pm$0.00)} & 58.70 {\small($\pm$0.48)}\\
    \cline{2-7}
    & TS &2& 0.47 {\small($\pm$0.00)} & 0.05 {\small($\pm$0.00)} & 0.78 {\small($\pm$0.00)} & 60.57 {\small($\pm$0.31)}\\
    \cline{2-7}
    &Energy & 1000& \textbf{0.61} {\small($\pm$0.00)} & 0.03 {\small($\pm$0.00)} & 0.71 {\small($\pm$0.00)} & \textbf{35.86} {\small($\pm$0.41)}\\   
    \cline{1-7}
    \multirow{3}{*}{\parbox{2cm}{MRI+ \\nnU-net}}&MSP & - & 0.87 {\small($\pm$0.00)} & 0.10 {\small($\pm$0.00)} & 0.22 {\small($\pm$0.00)} & 1,115.75 {\small($\pm$0.60)} \\
    \cline{2-7}
    & TS & 3& \textbf{0.92} {\small($\pm$0.00)} & \textbf{0.15} {\small($\pm$0.00)} & \textbf{0.14} {\small($\pm$0.00)} & 1,252.07 {\small($\pm$0.22)} \\  
    \cline{2-7}
    &Energy & 1& 0.68 {\small($\pm$0.00)} & 0.09 {\small($\pm$0.00)} & 0.75 {\small($\pm$0.00)} & \textbf{185.96} {\small($\pm$0.39)} \\
    \cline{1-7}
    \multirow{3}{*}{\parbox{2cm}{CT \\nnU-net}}&MSP & - & - & - & - & - \\
    \cline{2-7}
    & TS & - & - & - & - & - \\
    \cline{2-7}
    &Energy & - & - & - & - & - \\    
    \cline{1-7}
    \end{tabularx}
\end{table}

\begin{table}[h]
    \centering
    \caption{
    OOD detection results for the best-performing configurations of the comparison methods: MSP, temperature scaling (TS) and energy scoring (Energy) with temperature $T$, MC Dropout (MCD), and ensembling (Ensemble).
    ID is a DSC $\geq$ the median DSC.
    Seconds is the amount of time it took to calculate the test distances.
    The results are averages ($\pm$SD) across 5 runs.
    Arrows denote whether a higher or lower value is better.
    Bold highlights the best performance per model.
    }
    \label{tab:more_results}
    \begin{tabularx}{\textwidth}{|X|l|c|c|c|c|c|}
    \cline{1-7}
    \textbf{Model} & \textbf{Method} & \textbf{T} & \textbf{AUROC} $\uparrow$ & \textbf{AUPRC} $\uparrow$ & \textbf{FPR90} $\downarrow$ & \textbf{Seconds} $\downarrow$\\
    \cline{1-7}
    \multirow{3}{*}{\parbox{2cm}{MRI \\UNETR}}&MSP & - & \textbf{0.96} {\small($\pm$0.00)} & \textbf{0.97} {\small($\pm$0.00)} & \textbf{0.00} {\small($\pm$0.00)} & 8.73 {\small($\pm$0.12)}\\
    \cline{2-7}
    & TS & 2& 0.90 {\small($\pm$0.00)} & 0.93 {\small($\pm$0.00)} & 0.15 {\small($\pm$0.00)} & 9.14 {\small($\pm$0.05)}\\
    \cline{2-7}
    &Energy & 4& 0.64 {\small($\pm$0.00)} & 0.56 {\small($\pm$0.00)} & 0.60 {\small($\pm$0.00)} & \textbf{6.95} {\small($\pm$0.26)} \\
    \cline{1-7}
    \parbox{2cm}{\vspace{.1cm}MRI \\ Dropout\vspace{.1cm}}& MCD&- & \textbf{1.00} {\small($\pm$0.00)} & \textbf{1.00} {\small($\pm$0.00)} & \textbf{0.00} {\small($\pm$0.00)} & \textbf{14.57} {\small($\pm$0.14)}\\
    \cline{1-7}
    \parbox{2cm}{\vspace{.1cm}MRI \\ Ensemble\vspace{.1cm}} & Ensemble & - & \textbf{0.96} {\small($\pm$0.00)} & \textbf{0.96} {\small($\pm$0.00)} & \textbf{0.14} {\small($\pm$0.00)} & \textbf{13.93} {\small($\pm$0.09)}\\
    \cline{1-7}
    \multirow{3}{*}{\parbox{2cm}{MRI+ \\UNETR}} & MSP & - & 0.60 {\small($\pm$0.00)} & 0.62 {\small($\pm$0.00)}& \textbf{0.85} {\small($\pm$0.00)}& 58.31 {\small($\pm$0.21)}\\
    \cline{2-7}
    & TS & 5& \textbf{0.63} {\small($\pm$0.00)} & \textbf{0.65} {\small($\pm$0.00)}& \textbf{0.85} {\small($\pm$0.00)}& 60.77 {\small($\pm$0.29)}\\
    \cline{2-7}
    &Energy & 2& 0.62 {\small($\pm$0.00)} & 0.63 {\small($\pm$0.00)} & \textbf{0.85} {\small($\pm$0.00)} & \textbf{42.34} {\small($\pm$0.26)} \\ 
    \cline{1-7}
    \multirow{3}{*}{\parbox{2cm}{MRI+ \\nnU-net}}&MSP & - &0.51 {\small($\pm$0.00)}& 0.56 {\small($\pm$0.00)}&0.99 {\small($\pm$0.00)}&1,114.55 {\small($\pm$0.17)}\\
    \cline{2-7}
    &TS& 5& \textbf{0.64} {\small($\pm$0.00)} & \textbf{0.66} {\small($\pm$0.00)} & 0.85 {\small($\pm$0.00)} & 1,252.77 {\small($\pm$0.19)}\\
    \cline{2-7}
    &Energy & 1& 0.63 {\small($\pm$0.00)} & 0.64 {\small($\pm$0.00)} & \textbf{0.78} {\small($\pm$0.00)} & \textbf{190.26} {\small($\pm$0.42)} \\
    \cline{1-7}
    \multirow{3}{*}{\parbox{2cm}{CT \\nnU-net}}&MSP & - & \textbf{0.67} {\small($\pm$0.00)} & \textbf{0.71} {\small($\pm$0.00)} & \textbf{0.91} {\small($\pm$0.00)} & 567.86 {\small($\pm$0.33)}\\
    \cline{2-7}
    &TS & 3& 0.66 {\small($\pm$0.00)} & 0.70 {\small($\pm$0.00)} & \textbf{0.91} {\small($\pm$0.00)} & 699.59 {\small($\pm$0.24)}\\
    \cline{2-7}
    &Energy & 1& 0.65 {\small($\pm$0.00)} & 0.70 {\small($\pm$0.00)} & 0.93 {\small($\pm$0.00)} & \textbf{100.77} {\small($\pm$0.84)} \\   
    \cline{1-7}
    \end{tabularx}
\end{table}

\clearpage

\clearpage
\section{Hyperparameter searches}
\label{sec:hyperparam}

\subsection{Mahalanobis Distance}
\label{sec:hyper_md}

\begin{table}[h!]
    \centering
    \label{tab:mri_unetr_search}
    \caption{
    MD hyperparameter searches for the MRI UNETR (95\% threshold).
    Dimensionality reduction techniques include PCA($n_p$) with $n_p$ components, UMAP($n_u$) with $n_u$ components, and $n_d$-dimensional average pooling with kernel size $k$ and stride $s$, Pool$n_d$D($k$,$s$).
    Seconds is the amount of time it took to calculate the test distances.
    The results are averages ($\pm$SD) across 5 runs.
    Arrows denote whether higher or lower is better.
    Bold highlights the best performance per technique.
    }
    \begin{tabular}{|l|c|c|c|c|}
    \hline
    \textbf{Experiment} & \textbf{AUROC} $\uparrow$ & \textbf{AUPRC} $\uparrow$ & \textbf{FPR90} $\downarrow$ & \textbf{Seconds} $\downarrow$ \\
    \hline
    PCA(2) & 0.90 ($\pm$0.00) & 0.93 ($\pm$0.00) & 0.38 ($\pm$0.00) & 0.95 ($\pm$0.06)\\
    \hline
    PCA(4) & 0.70 ($\pm$0.00) & 0.66 ($\pm$0.00) & 0.46 ($\pm$0.00) &1.02 ($\pm$0.04)\\
    \hline
    PCA(8) & 0.73 ($\pm$0.00) & 0.74 ($\pm$0.00) & 0.54 ($\pm$0.00) & 1.02 ($\pm$0.10)\\
    \hline
    PCA(16) & 0.87 ($\pm$0.00) & 0.87 ($\pm$0.00) & \textbf{0.23} ($\pm$0.00) & 1.13 ($\pm$0.11) \\
    \hline
    PCA(32) & 0.85 ($\pm$0.01) & 0.88 ($\pm$0.01) & 0.34 ($\pm$0.08) & 1.18 ($\pm$0.08) \\
    \hline
    PCA(64) & 0.85 ($\pm$0.01) & 0.87 ($\pm$0.02) & \textbf{0.23} ($\pm$0.00) & 1.31 ($\pm$0.18)\\
    \hline
    PCA(128) & 0.91 ($\pm$0.00) & 0.93 ($\pm$0.00) & \textbf{0.23} ($\pm$0.00) & 1.99 ($\pm$0.09)\\
    \hline
    PCA(256) & \textbf{0.93} ($\pm$0.00) & \textbf{0.94} ($\pm$0.00) & \textbf{0.23} ($\pm$0.00) & 2.82 ($\pm$0.14)\\
    \hlineB{5}
    UMAP(2) & \textbf{0.77} ($\pm$0.08) & 0.79 ($\pm$0.11) & \textbf{0.57} ($\pm$0.08) & 10.44 ($\pm$0.22)\\
    \hline
    UMAP(4) & 0.75 ($\pm$0.04) & 0.80 ($\pm$0.07) & 0.66 ($\pm$0.16) & 10.63 ($\pm$0.39)\\
    \hline
    UMAP(8) & 0.74 ($\pm$0.05) & \textbf{0.82} ($\pm$0.03) & 0.72 ($\pm$0.17) &  10.57 ($\pm$0.16)\\
    \hline
    UMAP(16) & 0.65 ($\pm$0.03) & 0.77 ($\pm$0.02) & 0.91 ($\pm$0.09) & 10.76 ($\pm$0.29)\\
    \hline
    UMAP(32) & 0.66 ($\pm$0.03) & 0.77 ($\pm$0.02) & 0.91 ($\pm$0.06) & 10.64 ($\pm$0.38)\\
    \hline
    UMAP(64) & 0.63 ($\pm$0.03) & 0.75 ($\pm$0.03) & 0.88 ($\pm$0.06) & 10.76 ($\pm$0.33)\\
    \hline
    UMAP(128) & 0.65 ($\pm$0.03) & 0.77 ($\pm$0.01) & 0.88 ($\pm$0.01) & 10.99 ($\pm$0.45)\\
    \hline
    UMAP(256) & 0.63 ($\pm$0.03) & 0.76 ($\pm$0.02) & 0.88 ($\pm$0.08) & 11.17 ($\pm$0.28) \\
    \hlineB{5}
    Pool2D(2,1) & 0.71 ($\pm$0.00) & 0.74 ($\pm$0.00) & 0.54 ($\pm$0.00) & 1,446.90 ($\pm$11.62) \\
    \hline
    Pool2D(2,2) & 0.63 ($\pm$0.00) & 0.69 ($\pm$0.00) & 0.85 ($\pm$0.00) & 187.00 ($\pm$10.68)\\
    \hline
    Pool2D(3,1) & 0.72 ($\pm$0.00) & 0.72 ($\pm$0.00) & 0.54 ($\pm$0.00) & 145.02 ($\pm$0.24)\\
    \hline
    Pool2D(3,2) & \textbf{0.82} ($\pm$0.00) & \textbf{0.86} ($\pm$0.00) & \textbf{0.46} ($\pm$0.00) & 15.32 ($\pm$8.92)\\
    \hline
    Pool2D(4,1) & 0.73 ($\pm$0.00) & 0.78 ($\pm$0.00) & 0.77 ($\pm$0.00) & 11.95 ($\pm$5.92)\\
    \hline
    Pool3D(2,1) & 0.70 ($\pm$0.00) & 0.78 ($\pm$0.00) & 0.92 ($\pm$0.00) & 1,109.29 ($\pm$12.88)\\
    \hline
    Pool3D(2,2) & 0.60 ($\pm$0.00) & 0.69 ($\pm$0.00) & 0.77 ($\pm$0.00) & 18.69 ($\pm$0.25)\\
    \hline
    Pool3D(3,1) & 0.75 ($\pm$0.00) & 0.82 ($\pm$0.00) & 0.85 ($\pm$0.00) & 74.33 ($\pm$15.52)\\
    \hline
    Pool3D(3,2) & 0.54 ($\pm$0.00) & 0.58 ($\pm$0.00) & 0.85 ($\pm$0.00) & 1.10 ($\pm$0.04)\\
    \hline
    Pool3D(4,1) & 0.70 ($\pm$0.00) & 0.74 ($\pm$0.00)& 0.54 ($\pm$0.00)& 1.89 ($\pm$0.16)\\
    \hline
    \end{tabular}
\end{table}

\begin{table}
    \centering
    \caption{
    MD hyperparameter searches for the MRI+ UNETR (80\% threshold).
    Dimensionality reduction techniques include PCA($n_p$) with $n_p$ components, UMAP($n_u$) with $n_u$ components, and $n_d$-dimensional average pooling with kernel size $k$ and stride $s$, Pool$n_d$D($k$,$s$).
    Seconds is the amount of time it took to calculate the test distances.
    The results are averages ($\pm$SD) across 5 runs.
    Arrows denote whether higher or lower is better.
    Bold highlights the best performance per technique.
    }
    \begin{tabular}{|l|c|c|c|c|}
    \cline{1-5}
    \textbf{Experiment} & \textbf{AUROC} $\uparrow$ & \textbf{AUPRC} $\uparrow$ & \textbf{FPR90} $\downarrow$ & \textbf{Seconds} $\downarrow$ \\
    \cline{1-5}
    PCA(2) & 0.60 ($\pm$0.00)&0.03 ($\pm$0.00)&0.57 ($\pm$0.00)& 1.59 ($\pm$0.03)\\
    \hline
    PCA(4) & 0.57 ($\pm$0.00) & 0.03 ($\pm$0.00) & 0.62 ($\pm$0.00) & 1.72 ($\pm$0.09)\\
    \hline
    PCA(8) & 0.80 ($\pm$0.00) & \textbf{0.20} ($\pm$0.00) & 0.46 ($\pm$0.00) & 1.56 ($\pm$0.03)\\
    \hline
    PCA(16) & 0.84 ($\pm$0.01) & 0.10 ($\pm$0.01) & 0.35 ($\pm$0.02) & 1.60 ($\pm$0.23)\\
    \hline
    PCA(32) & \textbf{0.85} ($\pm$0.01) & 0.13 ($\pm$0.00) & 0.34 ($\pm$0.02) & 1.86 ($\pm$0.32)\\
    \hline
    PCA(64) & 0.80 ($\pm$0.01) & 0.06 ($\pm$0.00) & 0.36 ($\pm$0.02) & 1.71 ($\pm$0.05)\\
    \hline
    PCA(128) & 0.75 ($\pm$0.02) & 0.04 ($\pm$0.00) & 0.37 ($\pm$0.01) & 2.88 ($\pm$0.13)\\
    \hline
    PCA(256) & 0.75 ($\pm$0.01) & 0.04 ($\pm$0.00) & \textbf{0.32} ($\pm$0.01) & 3.57 ($\pm$0.10)\\
    \hlineB{5}
    UMAP(2) & \textbf{0.68} ($\pm$0.07) & \textbf{0.05} ($\pm$0.01) & \textbf{0.49} ($\pm$0.06) & 21.29 ($\pm$0.43) \\
    \hline
    UMAP(4) & 0.64 ($\pm$0.07) & 0.04 ($\pm$0.00) & 0.50 ($\pm$0.07) & 21.37 ($\pm$0.44)\\
    \hline
    UMAP(8)& 0.63 ($\pm$0.02) & 0.03 ($\pm$0.00) & 0.51 ($\pm$0.05) & 21.39 ($\pm$0.39)\\
    \hline
    UMAP(16)& 0.60 ($\pm$0.04) & 0.03 ($\pm$0.01) & 0.54 ($\pm$0.05) & 21.07 ($\pm$0.48)\\
    \hline
    UMAP(32)& 0.57 ($\pm$0.04) & 0.03 ($\pm$0.00) & 0.66 ($\pm$0.03) & 21.35 ($\pm$0.64)\\
    \hline
    UMAP(64)& 0.53 ($\pm$0.03) & 0.03 ($\pm$0.01) & 0.64 ($\pm$0.05) & 21.73 ($\pm$0.77)\\
    \hline
    UMAP(128)& 0.54 ($\pm$0.04) & 0.03 ($\pm$0.00) & 0.61 ($\pm$0.04) & 21.90 ($\pm$0.71)\\
    \hline
    UMAP(256)& 0.55 ($\pm$0.04) & 0.03 ($\pm$0.00) & 0.60 ($\pm$0.03) & 22.47 ($\pm$0.51)\\
    \hlineB{5}
    Pool2D(2,1)& 0.60 ($\pm$0.00) & 0.03 ($\pm$0.00) & 0.81 ($\pm$0.00) & 1,602.81 ($\pm$48.97)\\
    \hline
    Pool2D(2,2)& 0.62 ($\pm$0.00) & 0.03 ($\pm$0.00) & 0.63 ($\pm$0.00) & 218.32 ($\pm$33.26)\\
    \hline
    Pool2D(3,1)& 0.32 ($\pm$0.00) & 0.02 ($\pm$0.00) & 0.96 ($\pm$0.00) & 208.96 ($\pm$24.75) \\
    \hline
    Pool2D(3,2)& 0.58 ($\pm$0.00) & 0.03 ($\pm$0.00) & 0.73 ($\pm$0.00) & 36.33 ($\pm$8.93)\\
    \hline
    Pool2D(4,1)& \textbf{0.64} ($\pm$0.00) & 0.04 ($\pm$0.00) & 0.64 ($\pm$0.00) & 16.08 ($\pm$13.41)\\
    \hline
    Pool3D(2,1)& 0.57 ($\pm$0.00) & 0.02 ($\pm$0.00) & \textbf{0.61} ($\pm$0.00) & 1,338.82 ($\pm$99.92) \\
    \hline
    Pool3D(2,2)& 0.59 ($\pm$0.00) & \textbf{0.05} ($\pm$0.00) & 0.88 ($\pm$0.00) & 68.26 ($\pm$7.43)\\
    \hline
    Pool3D(3,1)& 0.39 ($\pm$0.00) & 0.02 ($\pm$0.00) & 0.99 ($\pm$0.00) & 117.52 ($\pm$9.36) \\
    \hline
    Pool3D(3,2)& 0.59 ($\pm$0.00) & 0.03 ($\pm$0.00) & 0.66 ($\pm$0.00) & 29.52 ($\pm$6.92)\\
    \hline
    Pool3D(4,1)& 0.62 ($\pm$0.00) & 0.03 ($\pm$0.00) & 0.62 ($\pm$0.00) & 39.30 ($\pm$0.49)\\
    \cline{1-5}
    \end{tabular}
\end{table}

\begin{table}
    \centering
    \caption{
    MD hyperparameter searches for the MRI+ nnU-net (95\% threshold).
    Dimensionality reduction techniques include PCA($n_p$) with $n_p$ components, UMAP($n_u$) with $n_u$ components, and $n_d$-dimensional average pooling with kernel size $k$ and stride $s$, Pool$n_d$D($k$,$s$).
    Seconds is the amount of time it took to calculate the test distances.
    The results are averages ($\pm$SD) across 5 runs.
    Arrows denote whether higher or lower is better.
    Bold highlights the best performance per technique.
    }
    \begin{tabular}{|l|c|c|c|c|}
    \hline
    \textbf{Experiment} & \textbf{AUROC} $\uparrow$ & \textbf{AUPRC} $\uparrow$ & \textbf{FPR90} $\downarrow$ & \textbf{Seconds} $\downarrow$ \\
    \hline
    PCA(2)&0.88 ($\pm$0.00)&\textbf{1.00} ($\pm$0.00)&0.67 ($\pm$0.00)&1.04 ($\pm$0.04)\\
    \hline
    PCA(4)&0.65 ($\pm$0.00) &\textbf{1.00} ($\pm$0.00) & 1.00 ($\pm$0.00) & 1.17 ($\pm$0.06)\\
    \hline
    PCA(8)& \textbf{0.96} ($\pm$0.00) & \textbf{1.00} ($\pm$0.00) & \textbf{0.00} ($\pm$0.00) & 1.12 ($\pm$0.04)\\
    \hline
    PCA(16)& 0.91 ($\pm$0.00) & \textbf{1.00} ($\pm$0.00) & \textbf{0.00} ($\pm$0.00) & 1.16 ($\pm$0.03)\\
    \hline
    PCA(32)&0.89 ($\pm$0.00)& \textbf{1.00} ($\pm$0.00) & 0.33 ($\pm$0.00) & 1.22 ($\pm$0.03)\\
    \hline
    PCA(64)&0.81 ($\pm$0.01) & \textbf{1.00} ($\pm$0.00) & 1.00 ($\pm$0.00) & 1.28 ($\pm$0.08)\\
    \hline
    PCA(128)& 0.85 ($\pm$0.01)& \textbf{1.00} ($\pm$0.00)&0.67 ($\pm$0.21)&1.66 ($\pm$0.05)\\
    \hline
    PCA(256)& 0.84 ($\pm$0.01) & \textbf{1.00} ($\pm$0.00) & 0.93 ($\pm$0.13) & 2.27 ($\pm$0.08)\\
    \hlineB{5}
    UMAP(2)& 0.60 ($\pm$0.07) & 0.99 ($\pm$0.00) & 1.00 ($\pm$0.00) & 18.34 ($\pm$0.71)\\
    \hline
    UMAP(4)& 0.68 ($\pm$0.08) & \textbf{1.00} ($\pm$0.00) & 0.87 ($\pm$0.27) & 18.69 ($\pm$0.47)\\
    \hline
    UMAP(8)& 0.77 ($\pm$0.07) & \textbf{1.00} ($\pm$0.00) & 0.73 ($\pm$0.33)& 19.27 ($\pm$0.94)\\
    \hline
    UMAP(16)& \textbf{0.82} ($\pm$0.08) & \textbf{1.00} ($\pm$0.00) & \textbf{0.67} ($\pm$0.03)& 19.07 ($\pm$0.96)\\
    \hline
    UMAP(32)&0.66 ($\pm$0.09) & \textbf{1.00} ($\pm$0.00) & 0.93 ($\pm$0.13)& 18.74 ($\pm$0.63)\\
    \hline
    UMAP(64)& 0.55 ($\pm$0.11)& \textbf{1.00} ($\pm$0.00)& 0.93 ($\pm$0.13)& 18.45 ($\pm$0.32)\\
    \hline
    UMAP(128)&0.64 ($\pm$0.03)&\textbf{1.00} ($\pm$0.00)&1.00 ($\pm$0.00)& 19.20 ($\pm$0.85)\\
    \hline
    UMAP(256)& 0.67 ($\pm$0.11) & \textbf{1.00} ($\pm$0.00) & 1.00 ($\pm$0.00) &19.85 ($\pm$0.99)\\
    \hlineB{5}
    Pool2D(2,1)&\textbf{0.85} ($\pm$0.00)& \textbf{1.00} ($\pm$0.00)& 0.67 ($\pm$0.00)& 1,579.73 ($\pm$52.41)\\
    \hline
    Pool2D(2,2)&0.49 ($\pm$0.00)&0.99 ($\pm$0.00)&1.00 ($\pm$0.00)& 32.25 ($\pm$0.16)\\
    \hline
    Pool2D(3,1)&0.39 ($\pm$0.00)& 0.99 ($\pm$0.00)&1.00 ($\pm$0.00)& 372.05 ($\pm$7.89)\\
    \hline
    Pool2D(3,2)& 0.20 ($\pm$0.00) & 0.98 ($\pm$0.00)& 1.00 ($\pm$0.00)& 47.42 ($\pm$11.16)\\
    \hline
    Pool2D(4,1)& 0.72 ($\pm$0.00) & \textbf{1.00} ($\pm$0.00)&\textbf{0.33} ($\pm$0.00)&103.14 ($\pm$20.14)\\
    \hline
    Pool3D(2,1)& 0.36 ($\pm$0.00) & 0.99 ($\pm$0.00) & 1.00 ($\pm$0.00)& 1,052.74 ($\pm$52.81)\\
    \hline
    Pool3D(2,2)& 0.81 ($\pm$0.00) & \textbf{1.00} ($\pm$0.00) & 0.67 ($\pm$0.00) & 35.73 ($\pm$8.67)\\
    \hline
    Pool3D(3,1)& 0.27 ($\pm$0.00) & 0.99 ($\pm$0.00) & 1.00 ($\pm$0.00) & 145.59 ($\pm$13.95)\\
    \hline
    Pool3D(3,2)& 0.84 ($\pm$0.00) & \textbf{1.00} ($\pm$0.00) & 0.67 ($\pm$0.00) & 24.64 ($\pm$11.49)\\
    \hline
    Pool3D(4,1)& 0.66 ($\pm$0.00) & \textbf{1.00} ($\pm$0.00) & 1.00 ($\pm$0.00) & 36.02 ($\pm$11.78)\\
    \hline
    \end{tabular}
\end{table}

\begin{table}
    \centering
    \caption{
    MD hyperparameter searches for the CT nnU-net (95\% threshold).
    Dimensionality reduction techniques include PCA($n_p$) with $n_p$ components, UMAP($n_u$) with $n_u$ components, and $n_d$-dimensional average pooling with kernel size $k$ and stride $s$, Pool$n_d$D($k$,$s$).
    Seconds is the amount of time it took to calculate the test distances.
    The results are averages ($\pm$SD) across 5 runs.
    Arrows denote whether higher or lower is better.
    Bold highlights the best performance per technique.
    }
    \begin{tabular}{|l|c|c|c|c|}
    \hline
    \textbf{Experiment} & \textbf{AUROC} $\uparrow$ & \textbf{AUPRC} $\uparrow$ & \textbf{FPR90} $\downarrow$ & \textbf{Seconds} $\downarrow$ \\
    \hline
    PCA(2)& 0.45 ($\pm$0.00) & 0.10 ($\pm$0.00) & 0.95 ($\pm$0.00) & 6.45 ($\pm$0.12)\\
    \hline
    PCA(4)& 0.51 ($\pm$0.00) & 0.16 ($\pm$0.00) & 0.91 ($\pm$0.00) & 6.64 ($\pm$0.11)\\
    \hline
    PCA(8)&0.55 ($\pm$0.00) & 0.13 ($\pm$0.00) & \textbf{0.85} ($\pm$0.00) & 7.22 ($\pm$0.07)\\
    \hline
    PCA(16)&0.51 ($\pm$0.00) & 0.12 ($\pm$0.00) & 0.96 ($\pm$0.00) & 7.79 ($\pm$0.15)\\
    \hline
    PCA(32)& \textbf{0.56} ($\pm$0.00) & \textbf{0.17} ($\pm$0.00) & 0.92 ($\pm$0.00) & 8.17 ($\pm$0.23)\\
    \hline
    PCA(64)& \textbf{0.56} ($\pm$0.00) & 0.16 ($\pm$0.00) & 0.93 ($\pm$0.00) & 8.57 ($\pm$0.24)\\
    \hline
    PCA(128)& \textbf{0.56} ($\pm$0.00) & 0.15 ($\pm$0.00) & 0.97 ($\pm$0.00) & 9.53 ($\pm$0.36)\\
    \hline
    PCA(256)& 0.54 ($\pm$0.00) & 0.13 ($\pm$0.00) & 0.98 ($\pm$0.00) & 40.72 ($\pm$2.68)\\
    \hlineB{5}
    UMAP(2)& 0.58 ($\pm$0.07) & 0.17 ($\pm$0.05) & 0.86 ($\pm$0.11) & 204.88 ($\pm$1.76)\\
    \hline
    UMAP(4)& 0.52 ($\pm$0.02) & 0.17 ($\pm$0.05) & 0.93 ($\pm$0.02) & 208.06 ($\pm$6.61)\\
    \hline
    UMAP(8)& 0.57 ($\pm$0.03) & 0.16 ($\pm$0.05) & 0.87 ($\pm$0.07)& 216.04 ($\pm$8.45)\\
    \hline
    UMAP(16)& 0.63 ($\pm$0.02) & 0.27 ($\pm$0.01) & 0.77 ($\pm$0.02) & 221.97 ($\pm$14.13)\\
    \hline
    UMAP(32)& 0.67 ($\pm$0.01) & \textbf{0.29} ($\pm$0.03) & 0.81 ($\pm$0.08) & 268.08 ($\pm$25.20)\\
    \hline
    UMAP(64)& 0.67 ($\pm$0.02) & 0.26 ($\pm$0.02) & 0.78 ($\pm$0.04) & 305.58 ($\pm$46.07)\\
    \hline
    UMAP(128)&\textbf{0.68} ($\pm$0.03) & 0.22 ($\pm$0.01) & 0.74 ($\pm$0.09) & 288.42 ($\pm$25.99) \\
    \hline
    UMAP(256)& 0.62 ($\pm$0.03) & 0.17 ($\pm$0.04) & \textbf{0.72} ($\pm$0.10) & 351.84 ($\pm$18.14)\\
    \hlineB{5}
    Pool2D(2,1)& 0.46 ($\pm$ 0.00) & 0.12 ($\pm$0.00) & 0.92 ($\pm$0.00) & 1,886.94 ($\pm$51.54)\\
    \hline
    Pool2D(2,2)& \textbf{0.59} ($\pm$0.00) & \textbf{0.13} ($\pm$0.00) & 0.84 ($\pm$0.00) & 163.84 ($\pm$19.54)\\
    \hline
    Pool2D(3,1)& 0.47 ($\pm$0.00) & 0.11 ($\pm$0.00) & 0.94 ($\pm$0.00) & 654.53 ($\pm$34.88) \\
    \hline
    Pool2D(3,2)& 0.46 ($\pm$0.00) & 0.10 ($\pm$0.00) & 0.92 ($\pm$0.00) & 77.30 ($\pm$39.40) \\
    \hline
    Pool2D(4,1)& 0.57 ($\pm$0.00) & \textbf{0.13} ($\pm$0.00) & 0.73 ($\pm$0.00) & 198.10 ($\pm$50.88)\\
    \hline
    Pool3D(2,1)& 0.54 ($\pm$0.00) & 0.12 ($\pm$0.00) & 0.88 ($\pm$0.00) & 1214.52 ($\pm$77.92)\\
    \hline
    Pool3D(2,2)& 0.55 ($\pm$0.00) & 0.11 ($\pm$0.00) & 0.75 ($\pm$0.00) & 106.38 ($\pm$27.31)\\
    \hline
    Pool3D(3,1)& 0.49 ($\pm$0.00) & \textbf{0.13} ($\pm$0.00) & 0.91 ($\pm$0.00) & 298.91 ($\pm$15.05)\\
    \hline
    Pool3D(3,2)& 0.50 ($\pm$0.00) & 0.10 ($\pm$0.00) & 0.90 ($\pm$0.00) & 93.39 ($\pm$15.41)\\
    \hline
    Pool3D(4,1)& 0.57 ($\pm$0.00) & \textbf{0.13} ($\pm$0.00) & \textbf{0.69} ($\pm$0.00) & 125.90 ($\pm$13.68)\\
    \hline
    \end{tabular}
\end{table}
\clearpage

\subsection{K-Nearest Neighbors}
\label{sec:hyper_knn}

\begin{table}[h!]
    \centering
    \label{tab:mri_unetr_search}
    \caption{KNN hyperparameter searches for MRI UNETR (95\% threshold).
    Dimensionality reduction techniques include PCA($n_p$) with $n_p$ components, UMAP($n_u$) with $n_u$ components, and $n_d$-dimensional average pooling with kernel size $k$ and stride $s$, Pool$n_d$D($k$,$s$).
    Only the best-performing configuration of $k$ is reported, with the logs containing all results available at \url{https://github.com/mckellwoodland/dimen_reduce_mahal/tree/main/logs}.
    Seconds is the amount of time it took to calculate the test distances.
    The results are averages ($\pm$SD) across 5 runs.
    Arrows denote whether higher or lower is better.
    Bold highlights the best performance per technique.}
    \begin{tabular}{|l|c|c|c|c|c|}
    \hline
    \textbf{Experiment} & \textbf{K} & \textbf{AUROC} $\uparrow$ & \textbf{AUPRC} $\uparrow$ & \textbf{FPR90} $\downarrow$ & \textbf{Seconds} $\downarrow$ \\
    \hline
    PCA(2) & 256 & \textbf{0.90} ($\pm$0.00) & \textbf{0.92} ($\pm$0.00) & 0.31 ($\pm$0.00) & 0.95 ($\pm$0.05)\\
    \hline
    PCA(4)&256&0.84 ($\pm$0.00) & 0.88 ($\pm$0.00) & 0.46 ($\pm$0.00) & 0.93 ($\pm$0.04)\\
    \hline
    PCA(8)&256&0.84 ($\pm$0.00) & 0.86 ($\pm$0.00) & 0.43 ($\pm$0.04) & 0.91 ($\pm$0.02)\\
    \hline
    PCA(16)&256&0.86 ($\pm$0.00) & 0.88 ($\pm$0.00) & 0.31 ($\pm$0.00) & 1.04 ($\pm$0.08)\\
    \hline
    PCA(32)&256&0.86 ($\pm$0.00) & 0.88 ($\pm$0.00) & 0.31 ($\pm$0.00) & 1.24 ($\pm$0.19)\\
    \hline
    PCA(64)&256&0.87 ($\pm$0.00) & 0.88 ($\pm$0.00) & \textbf{0.23} ($\pm$0.00) & 1.25 ($\pm$0.04)\\
    \hline
    PCA(128)&256 &0.87 ($\pm$0.00) & 0.88 ($\pm$0.00) & \textbf{0.23} ($\pm$0.00) & 1.86 ($\pm$0.03)\\
    \hline
    PCA(256)&8&0.88 ($\pm$0.00) & 0.89 ($\pm$0.00) & 0.26 ($\pm$0.04) & 2.64 ($\pm$0.08)\\
    \hlineB{5}
    UMAP(2)&256&0.75 ($\pm$0.03)&0.79 ($\pm$0.07)& 0.82 ($\pm$0.08) & 5.67 ($\pm$0.13)\\
    \hline
    UMAP(4)&256&0.82 ($\pm$0.04)&0.84 ($\pm$0.04)&0.59 ($\pm$0.15)& 6.41 ($\pm$0.20)\\
    \hline
    UMAP(8)&256&0.82 ($\pm$0.03) & 0.82 ($\pm$0.06) & \textbf{0.46} ($\pm$0.10) & 6.15 ($\pm$0.16)\\
    \hline
    UMAP(16)&256 &0.79 ($\pm$0.03) & 0.82 ($\pm$0.04) & 0.57 ($\pm$0.08) & 5.72 ($\pm$0.15)\\
    \hline
    UMAP(32) & 256 & \textbf{0.83} ($\pm$0.05) & \textbf{0.85} ($\pm$0.04) & 0.51 ($\pm$0.08) & 6.70 ($\pm$0.15)\\
    \hline
    UMAP(64)&256&0.80 ($\pm$0.05)&0.84 ($\pm$0.05)&0.65 ($\pm$0.13)& 6.60 ($\pm$0.21)\\
    \hline
    UMAP(128)&256 &0.77 ($\pm$0.07) & 0.81 ($\pm$0.06) & 0.69 ($\pm$0.17) & 6.80 ($\pm$0.13)\\
    \hline
    UMAP(256)&256&0.80 ($\pm$0.04) & 0.84 ($\pm$0.03) & 0.74 ($\pm$0.17) & 7.21 ($\pm$0.12)\\
    \hlineB{5}
    Pool2D(2,1) & 256 & \textbf{0.98} ($\pm$0.00) & \textbf{1.00} ($\pm$0.00) & \textbf{0.00} ($\pm$0.00) & 4.90 ($\pm$0.06)\\
    \hline
    Pool2D(2,2) & 256 & \textbf{0.98} ($\pm$0.00) & \textbf{1.00} ($\pm$0.00) & \textbf{0.00} ($\pm$0.00) & 0.74 ($\pm$0.04)\\
    \hline
    Pool2D(3,1)&256 &0.96 ($\pm$0.00) & \textbf{1.00} ($\pm$0.00) & 0.33 ($\pm$0.00) & 2.75 ($\pm$0.03)\\
    \hline
    Pool2D(3,2)&256 &0.94 ($\pm$0.00) & \textbf{1.00} ($\pm$0.00) & 0.33 ($\pm$0.00) & 9.54 ($\pm$8.16)\\
    \hline
    Pool2D(4,1)&256 &0.91 ($\pm$0.00) & \textbf{1.00} ($\pm$0.00) & 0.33 ($\pm$0.00) & 1.09 ($\pm$0.01)\\
    \hline
    Pool3D(2,1)& 256 &0.96 ($\pm$0.00) & \textbf{1.00} ($\pm$0.00) & \textbf{0.00} ($\pm$0.00) & 4.27 ($\pm$0.08)\\
    \hline
    Pool3D(2,2)&256 &0.85 ($\pm$0.00) & \textbf{1.00} ($\pm$0.00) & 1.00 ($\pm$0.00) & 0.36 ($\pm$0.01)\\
    \hline
    Pool3D(3,1)&256&0.94 ($\pm$0.00)& \textbf{1.00} ($\pm$0.00) & \textbf{0.00} ($\pm$0.00) & 1.76 ($\pm$0.08)\\
    \hline
    Pool3D(3,2) & 256 & 0.92 ($\pm$0.00) &\textbf{1.00} ($\pm$0.00) & 0.33 ($\pm$0.00) & 0.36 ($\pm$0.00)\\
    \hline
    Pool3D(4,1)&256 &0.89 ($\pm$0.00) & \textbf{1.00} ($\pm$0.00) & 0.33 ($\pm$0.00) & 0.64 ($\pm$0.02)\\
    \hline
    \end{tabular}
\end{table}

\begin{table}
    \centering
    \caption{KNN hyperparameter searches for MRI+ UNETR (80\% threshold).
    Dimensionality reduction techniques include PCA($n_p$) with $n_p$ components, UMAP($n_u$) with $n_u$ components, and $n_d$-dimensional average pooling with kernel size $k$ and stride $s$, Pool$n_d$D($k$,$s$).
    Only the best-performing configuration of $k$ is reported, with the logs containing all results available at \url{https://github.com/mckellwoodland/dimen_reduce_mahal/tree/main/logs}.
    Seconds is the amount of time it took to calculate the test distances.
    The results are averages ($\pm$SD) across 5 runs.
    Arrows denote whether higher or lower is better.
    Bold highlights the best performance per technique.}
    \begin{tabular}{|l|c|c|c|c|c|}
    \hline
    \textbf{Experiment} & \textbf{K} & \textbf{AUROC} $\uparrow$ & \textbf{AUPRC} $\uparrow$ & \textbf{FPR90} $\downarrow$ & \textbf{Seconds} $\downarrow$ \\
    \hline
    PCA(2)&128&0.62 ($\pm$0.00)&0.03 ($\pm$0.00)&0.49 ($\pm$0.00)&1.43 ($\pm$0.10)\\
    \hline
    PCA(4)&128&0.63 ($\pm$0.00)&0.03 ($\pm$0.00)&0.54 ($\pm$0.00)&1.35 ($\pm$0.03)\\
    \hline
    PCA(8)&128&0.78 ($\pm$0.00)&0.06 ($\pm$0.00)&0.43 ($\pm$0.00)&1.42 ($\pm$0.04)\\
    \hline
    PCA(16)&64&0.82 ($\pm$0.01)&0.09 ($\pm$0.01)&0.42 ($\pm$0.01)&1.44 ($\pm$0.03)\\
    \hline
    PCA(32)&64&\textbf{0.84} ($\pm$0.01)&\textbf{0.17} ($\pm$0.02)&\textbf{0.40} ($\pm$0.01)&1.57 ($\pm$0.06)\\
    \hline
    PCA(64)&128&0.83 ($\pm$0.01)&0.11 ($\pm$0.01)&0.42 ($\pm$0.01) & 1.67 ($\pm$0.10)\\
    \hline
    PCA(128)&128&0.83 ($\pm$0.00)&0.13 ($\pm$0.01)&0.42 ($\pm$0.01)&2.66 ($\pm$0.11)\\
    \hline
    PCA(256)&64&0.83 ($\pm$0.00)&0.14 ($\pm$0.01)&0.42 ($\pm$0.01)&4.26 ($\pm$1.19)\\
    \hlineB{5}
    UMAP(2)&128&0.69 ($\pm$0.09) & 0.07 ($\pm$0.06) & 0.46 ($\pm$0.08) & 17.26 ($\pm$0.61)\\
    \hline
    UMAP(4)&64&\textbf{0.78} ($\pm$0.05)& 0.09 ($\pm$0.06) & 0.42 ($\pm$0.07) & 15.70 ($\pm$0.45)\\
    \hline
    UMAP(8)&128&0.73 ($\pm$0.09)&0.06 ($\pm$0.03)&0.46 ($\pm$0.06)&17.80 ($\pm$0.26)\\
    \hline
    UMAP(16)&128&0.73 ($\pm$0.05)&0.08 ($\pm$0.06)&0.46 ($\pm$0.03)&17.26 ($\pm$0.77)\\
    \hline
    UMAP(32)&128&0.73 ($\pm$0.07)&0.06 ($\pm$0.04)&0.46 ($\pm$0.05)&17.23 ($\pm$0.25)\\
    \hline
    UMAP(64)&256&0.74 ($\pm$0.07)&0.05 ($\pm$0.01)&0.42 ($\pm$0.06)&17.43 ($\pm$0.88)\\
    \hline
    UMAP(128)&256&0.74 ($\pm$0.05)&0.05 ($\pm$0.02)&0.44 ($\pm$0.03)&18.50 ($\pm$0.86)\\
    \hline
    UMAP(256)&64&0.75 ($\pm$0.06) &\textbf{0.12} ($\pm$0.16) &\textbf{0.39} ($\pm$0.03)&17.27 ($\pm$0.50)\\
    \hlineB{5}
    Pool2D(2,1)&128&0.86 ($\pm$0.00)&0.19 ($\pm$0.00)&0.40 ($\pm$0.00)&6.28 ($\pm$0.05)\\
    \hline
    Pool2D(2,2)&256 & 0.85 ($\pm$0.00) &0.27 ($\pm$0.00) & 0.43 ($\pm$0.00) & 13.53 ($\pm$7.27)\\
    \hline
    Pool2D(3,1)&64&0.86 ($\pm$0.00)&0.15 ($\pm$0.00)&0.31 ($\pm$0.00)&2.81 ($\pm$0.06)\\
    \hline
    Pool2D(3,2)&256&0.76 ($\pm$0.00)&0.07 ($\pm$0.00)&0.58 ($\pm$0.00)&1.34 ($\pm$0.06)\\
    \hline
    Pool2D(4,1)&32&0.85 ($\pm$0.00)&0.27 ($\pm$0.00)&0.40 ($\pm$0.00)&9.09 ($\pm$5.19)\\
    \hline
    Pool3D(2,1)&256&\textbf{0.87} ($\pm$0.00)& 0.30 ($\pm$0.00) & 0.38 ($\pm$0.00) & 23.34 ($\pm$2.74)\\
    \hline
    Pool3D(2,2)&256&\textbf{0.87} ($\pm$0.00) & \textbf{0.33} ($\pm$0.00) &0.43 ($\pm$0.00) & 11.21 ($\pm$7.66)\\
    \hline
    Pool3D(3,1)&64&\textbf{0.87} ($\pm$0.00) & 0.19 ($\pm$0.00) &\textbf{0.30} ($\pm$0.00) & 2.30 ($\pm$0.10)\\
    \hline
    Pool3D(3,2)&64&0.78 ($\pm$0.00) & 0.06 ($\pm$0.00) &0.41 ($\pm$0.00) & 11.14 ($\pm$6.26)\\
    \hline
    Pool3D(4,1)&8&0.84 ($\pm$0.00) & 0.26 ($\pm$0.00) & 0.41 ($\pm$0.00) &1.24 ($\pm$0.07)\\
    \hline
    \end{tabular}
\end{table}

\begin{table}
    \centering
    \caption{KNN hyperparameter searches for MRI+ nnU-net (95\% threshold).
    Dimensionality reduction techniques include PCA($n_p$) with $n_p$ components, UMAP($n_u$) with $n_u$ components, and $n_d$-dimensional average pooling with kernel size $k$ and stride $s$, Pool$n_d$D($k$,$s$).
    Only the best-performing configuration of $k$ is reported, with the logs containing all results available at \url{https://github.com/mckellwoodland/dimen_reduce_mahal/tree/main/logs}.
    Seconds is the amount of time it took to calculate the test distances.
    The results are averages ($\pm$SD) across 5 runs.
    Arrows denote whether higher or lower is better.
    Bold highlights the best performance per technique.}
    \begin{tabular}{|l|c|c|c|c|c|}
    \hline
    \textbf{Experiment} & \textbf{K} & \textbf{AUROC} $\uparrow$ & \textbf{AUPRC} $\uparrow$ & \textbf{FPR90} $\downarrow$ & \textbf{Seconds} $\downarrow$ \\
    \hline
    PCA(2)&256&0.94 ($\pm$0.00)&\textbf{1.00} ($\pm$0.00) & \textbf{0.00} ($\pm$0.00) & 0.99 ($\pm$0.02)\\
    \hline
    PCA(4)&256&0.75 ($\pm$0.00) & \textbf{1.00} ($\pm$0.00) & 1.00 ($\pm$0.00) & 1.04 ($\pm$0.02)\\
    \hline
    PCA(8)&256&0.97 ($\pm$0.00) & \textbf{1.00} ($\pm$0.00) & \textbf{0.00} ($\pm$0.00) & 1.16 ($\pm$0.07)\\
    \hline
    PCA(16)&256&\textbf{0.98} ($\pm$0.00) & \textbf{1.00} ($\pm$0.00) & 0.33 ($\pm$0.00) & 1.16 ($\pm$0.08)\\
    \hline
    PCA(32)&256&0.95 ($\pm$0.00) & \textbf{1.00} ($\pm$0.00) & \textbf{0.00} ($\pm$0.00) & 1.25 ($\pm$0.04)\\
    \hline
    PCA(64)&256&0.95 ($\pm$0.00)&\textbf{1.00} ($\pm$0.00)&\textbf{0.00} ($\pm$0.00)&1.24 ($\pm$0.07)\\
    \hline
    PCA(128)&256&0.96 ($\pm$0.00) & \textbf{1.00} ($\pm$0.00) & \textbf{0.00} ($\pm$0.00) & 1.71 ($\pm$0.07)\\
    \hline
    PCA(256)&256&0.95 ($\pm$0.00) & \textbf{1.00} ($\pm$0.00) & \textbf{0.00} ($\pm$0.00) & 2.89 ($\pm$0.41)\\
    \hlineB{5}
    UMAP(2)&2&\textbf{0.96} ($\pm$0.04) & \textbf{1.00} ($\pm$0.00) & \textbf{0.13} ($\pm$0.16) & 14.91 ($\pm$2.08)\\
    \hline
    UMAP(4)&2&0.94 ($\pm$0.02) & \textbf{1.00} ($\pm$0.00) & 0.33 ($\pm$0.00) & 15.15 ($\pm$2.30)\\
    \hline
    UMAP(8)&2&0.94 ($\pm$0.03) & \textbf{1.00} ($\pm$0.00) & 0.40 ($\pm$0.33) & 15.22 ($\pm$1.57)\\
    \hline
    UMAP(16)&2 &0.94 ($\pm$0.02) & \textbf{1.00} ($\pm$0.00) &0.26 ($\pm$0.13) & 14.89 ($\pm$2.38)\\
    \hline
    UMAP(32)&2&0.89 ($\pm$0.05) & \textbf{1.00} ($\pm$0.00) & 0.40 ($\pm$0.33) & 14.64 ($\pm$1.92)\\
    \hline
    UMAP(64)&2&0.92 ($\pm$0.02) & \textbf{1.00} ($\pm$0.00) & 0.26 ($\pm$0.13) & 14.82 ($\pm$1.92)\\
    \hline
    UMAP(128)&2&0.92 ($\pm$0.02) & \textbf{1.00} ($\pm$0.00) & 0.33 ($\pm$0.00) & 15.08 ($\pm$1.74)\\
    \hline
    UMAP(256)&2&0.91 ($\pm$0.04) & \textbf{1.00} ($\pm$0.00) & 0.40 ($\pm$0.14) & 15.75 ($\pm$1.83)\\
    \hlineB{5}
    Pool2D(2,1)&256&\textbf{0.98} ($\pm$0.00) & \textbf{1.00} ($\pm$0.00) & \textbf{0.00} ($\pm$0.00) & 4.90 ($\pm$0.06)\\
    \hline
    Pool2D(2,2)&256&\textbf{0.98} ($\pm$0.00) & \textbf{1.00} ($\pm$0.00) & \textbf{0.00} ($\pm$0.00) & 0.74 ($\pm$0.04)\\
    \hline
    Pool2D(3,1)&256&0.96 ($\pm$0.00) & \textbf{1.00} ($\pm$0.00) & 0.33 ($\pm$0.00) & 2.75 ($\pm$0.03)\\
    \hline
    Pool2D(3,2)&256&0.94 ($\pm$0.00) & \textbf{1.00} ($\pm$0.00) & 0.33 ($\pm$0.00) & 9.54 ($\pm$8.16)\\
    \hline
    Pool2D(4,1)&256&0.91 ($\pm$0.00) & \textbf{1.00} ($\pm$0.00) & 0.33 ($\pm$0.00) & 1.09 ($\pm$0.01)\\
    \hline
    Pool3D(2,1)&256&0.96 ($\pm$0.00) & \textbf{1.00} ($\pm$0.00) & \textbf{0.00} ($\pm$0.00) & 4.27 ($\pm$0.08)\\
    \hline
    Pool3D(2,2)&256&0.85 ($\pm$0.00) & \textbf{1.00} ($\pm$0.00) & 1.00 ($\pm$0.00) &0.36 ($\pm$0.01)\\
    \hline
    Pool3D(3,1)&256&0.94 ($\pm$0.00) & \textbf{1.00} ($\pm$0.00) & \textbf{0.00} ($\pm$0.00) & 1.76 ($\pm$0.08)\\
    \hline
    Pool3D(3,2)&256&0.92 ($\pm$0.00)&\textbf{1.00} ($\pm$0.00)&0.33 ($\pm$0.00)&0.36 ($\pm$0.00)\\
    \hline
    Pool3D(4,1)&256&0.89 ($\pm$0.00) &\textbf{1.00} ($\pm$0.00) & 0.33 ($\pm$0.00) & 0.64 ($\pm$0.02)\\
    \hline
    \end{tabular}
\end{table}

\begin{table}
    \centering
    \caption{KNN hyperparameter searches for CT nnU-net (95\% threshold).
    Dimensionality reduction techniques include PCA($n_p$) with $n_p$ components, UMAP($n_u$) with $n_u$ components, and $n_d$-dimensional average pooling with kernel size $k$ and stride $s$, Pool$n_d$D($k$,$s$).
    Only the best-performing configuration of $k$ is reported, with the logs containing all results available at \url{https://github.com/mckellwoodland/dimen_reduce_mahal/tree/main/logs}.
    Seconds is the amount of time it took to calculate the test distances.
    The results are averages ($\pm$SD) across 5 runs.
    Arrows denote whether higher or lower is better.
    Bold highlights the best performance per technique.}
    \begin{tabular}{|l|c|c|c|c|c|}
    \hline
    \textbf{Experiment} & \textbf{K} & \textbf{AUROC} $\uparrow$ & \textbf{AUPRC} $\uparrow$ & \textbf{FPR90} $\downarrow$ & \textbf{Seconds} $\downarrow$ \\
    \hline
    PCA(2)&8&0.42 ($\pm$0.00) & 0.09 ($\pm$0.00) & 0.93 ($\pm$0.00) & 4.95 ($\pm$0.12)\\
    \hline
    PCA(4)&256&0.47 ($\pm$0.00) & 0.11 ($\pm$0.00) & \textbf{0.91} ($\pm$0.00) & 5.91 ($\pm$0.84)\\
    \hline
    PCA(8)&4&\textbf{0.55} ($\pm$0.00) & \textbf{0.15} ($\pm$0.00) & 0.97 ($\pm$0.00) & 4.94 ($\pm$0.04)\\
    \hline
    PCA(16)&4&0.52 ($\pm$0.00) & 0.12 ($\pm$0.00) & 0.96 ($\pm$0.00) & 5.00 ($\pm$0.04)\\
    \hline
    PCA(32)&4&0.52 ($\pm$0.00) & 0.13 ($\pm$0.00) & 0.97 ($\pm$0.00) & 5.20 ($\pm$0.07)\\
    \hline
    PCA(64)&8&0.53 ($\pm$0.00) & 0.13 ($\pm$0.00) & 0.96 ($\pm$0.00) & 5.60 ($\pm$0.10)\\
    \hline
    PCA(128)&4&0.53 ($\pm$0.00) & 0.13 ($\pm$0.00) & 0.96 ($\pm$0.00) & 5.98 ($\pm$0.04)\\
    \hline
    PCA(256)&4&0.53 ($\pm$0.00) & 0.13 ($\pm$0.00) & 0.96 ($\pm$0.00) & 24.75 ($\pm$14.29)\\
    \hlineB{5}
    UMAP(2)&256&0.64 ($\pm$0.02) & \textbf{0.24} ($\pm$0.04) & 0.88 ($\pm$0.03) & 200.58 ($\pm$1.93)\\
    \hline
    UMAP(4)&256&\textbf{0.65} ($\pm$0.01) & \textbf{0.24} ($\pm$0.05) & 0.88 ($\pm$0.02) & 199.93 ($\pm$1.96)\\
    \hline
    UMAP(8)&256&\textbf{0.65} ($\pm$0.01) & 0.22 ($\pm$0.04) & 0.87 ($\pm$0.02) & 198.33 ($\pm$2.34)\\
    \hline
    UMAP(16)&256 &\textbf{0.65} ($\pm$0.00) & 0.23 ($\pm$0.05) & \textbf{0.86} ($\pm$0.01) & 204.01 ($\pm$1.27)\\
    \hline
    UMAP(32)&256&0.63 ($\pm$0.01) & 0.19 ($\pm$0.04) & 0.88 ($\pm$0.02) & 195.59 ($\pm$0.55)\\
    \hline
    UMAP(64)&256&\textbf{0.65} ($\pm$0.01) & 0.22 ($\pm$0.03) & 0.88 ($\pm$0.01) & 197.23 ($\pm$0.68)\\
    \hline
    UMAP(128)&256&0.64 ($\pm$0.01) & 0.18 ($\pm$0.01) & \textbf{0.86} ($\pm$0.03) & 199.48 ($\pm$1.40)\\
    \hline
    UMAP(256)&256&0.64 ($\pm$0.00) & 0.19 ($\pm$0.02) & 0.87 ($\pm$0.03) & 200.73 ($\pm$0.65)\\
    \hlineB{5}
    Pool2D(2,1)&4&0.53 ($\pm$0.00) & \textbf{0.15} ($\pm$0.00) & 0.95 ($\pm$0.00) & 97.91 ($\pm$9.21)\\
    \hline
    Pool2D(2,2)&32&0.52 ($\pm$0.00) & \textbf{0.15} ($\pm$0.00) & 0.95 ($\pm$0.00) & 33.17 ($\pm$6.84)\\
    \hline
    Pool2D(3,1)&2&0.52 ($\pm$0.00) & 0.14 ($\pm$0.00) & \textbf{0.93} ($\pm$0.00) & 104.41 ($\pm$14.69)\\
    \hline
    Pool2D(3,2)&4&0.51 ($\pm$0.00) & 0.12 ($\pm$0.00) & 0.95 ($\pm$0.00) & 84.06 ($\pm$16.41)\\
    \hline
    Pool2D(4,1)&4&0.52 ($\pm$0.00) & 0.11 ($\pm$0.00) &0.95 ($\pm$0.00) & 42.95 ($\pm$16.59)\\
    \hline
    Pool3D(2,1)&4&\textbf{0.54} ($\pm$0.00) & 0.14 ($\pm$0.00) & 0.96 ($\pm$0.00) & 98.68 ($\pm$39.50)\\
    \hline
    Pool3D(2,2)&4&\textbf{0.54} ($\pm$0.00) & 0.14 ($\pm$0.00) & 0.96 ($\pm$0.00) & 81.96 ($\pm$26.79)\\
    \hline
    Pool3D(3,1)&4&0.53 ($\pm$0.00) & 0.14 ($\pm$0.00) & 0.94 ($\pm$0.00) & 81.41 ($\pm$28.32)\\
    \hline
    Pool3D(3,2)&4&0.53 ($\pm$0.00) & 0.13 ($\pm$0.00) & 0.95 ($\pm$0.00) & 75.00 ($\pm$14.21)\\
    \hline
    Pool3D(4,1)&4&0.53 ($\pm$0.00) & 0.13 ($\pm$0.00) & \textbf{0.93} ($\pm$0.00) & 84.91 ($\pm$11.64)\\
    \hline
    \end{tabular}
\end{table}

\clearpage
\subsection{Temperature Scaling and Energy Scoring}
\label{sec:hyper_comp}
\begin{table}[h]
    \centering
    \caption{
    Temperature scaling (TS) and energy scoring (Energy) hyperparameter searches for the UNETRs. 
    Seconds is the amount of time it took to calculate the test distances.
    The results are averages ($\pm$SD) across 5 runs.
    Arrows denote whether higher or lower is better.
    Bold denotes the best performance per method and model.
    }
    \begin{tabularx}{\textwidth}{|X|l|c|c|c|c|c|}
    \hline
    \textbf{Model} & \textbf{Method} & \textbf{T} & \textbf{AUROC} $\uparrow$ & \textbf{AUPRC} $\uparrow$ & \textbf{FPR90} $\downarrow$ & \textbf{Seconds} $\downarrow$ \\
    \hline
    \multirow{16}{*}{\parbox{2cm}{MRI\\UNETR}} & \multirow{8}{*}{\parbox{1cm}{TS}}&2& \textbf{0.90} {\small($\pm$0.00)} & \textbf{0.93} {\small($\pm$0.00)} & \textbf{0.15} {\small($\pm$0.00)}& 9.25 {\small($\pm$0.11)}\\
    \cline{3-7}
    &&3&0.72 {\small($\pm$0.00)} & 0.77 {\small($\pm$0.00)} & 0.38 {\small($\pm$0.00)} & 9.22 {\small($\pm$0.04)}\\
    \cline{3-7}
    &&4&0.67 {\small($\pm$0.00)}& 0.71 {\small($\pm$0.00)}& 0.46 {\small($\pm$0.00)}& 9.21 {\small($\pm$0.10)}\\
    \cline{3-7}
    &&5&0.61 {\small($\pm$0.00)}& 0.64 {\small($\pm$0.00)}& 0.54 {\small($\pm$0.00)}& 9.08 {\small($\pm$0.06)}\\
    \cline{3-7}
    &&10&0.57 {\small($\pm$0.00)}& 0.58 {\small($\pm$0.00)}& 0.62 {\small($\pm$0.00)}& 9.17 {\small($\pm$0.06)}\\
    \cline{3-7}
    &&100&0.55 {\small($\pm$0.00)}& 0.57 {\small($\pm$0.00)}& 0.69 {\small($\pm$0.00)}& 9.22 {\small($\pm$0.10)}\\
    \cline{3-7}
    &&1000&0.55 {\small($\pm$0.00)}& 0.57 {\small($\pm$0.00)}& 0.69 {\small($\pm$0.00)}& 9.20 {\small($\pm$0.07)}\\
    \cline{2-7}
    &\multirow{8}{*}{Energy}&1&\textbf{0.55} {\small($\pm$0.00)} & \textbf{0.57} {\small($\pm$0.00)}& \textbf{0.69} {\small($\pm$0.00)} & 7.32 {\small($\pm$0.28)}\\
    \cline{3-7}
    &&2& \textbf{0.55} {\small($\pm$0.00)} & \textbf{0.57} {\small($\pm$0.00)} & \textbf{0.69} {\small($\pm$0.00)} & 7.25 {\small($\pm$0.17)}\\
    \cline{3-7}
    &&3&\textbf{0.55} {\small($\pm$0.00)} &\textbf{0.57} {\small($\pm$0.00)}& \textbf{0.69} {\small($\pm$0.00)} & \textbf{7.08} {\small($\pm$0.14)} \\
    \cline{3-7}
    &&4&\textbf{0.55} {\small($\pm$0.00)} &\textbf{0.57} {\small($\pm$0.00)}& \textbf{0.69} {\small($\pm$0.00)} & 7.21 {\small($\pm$0.12)}\\
    \cline{3-7}
    &&5&\textbf{0.55} {\small($\pm$0.00)} &\textbf{0.57} {\small($\pm$0.00)}& \textbf{0.69} {\small($\pm$0.00)} & 7.38 {\small($\pm$0.22)}\\
    \cline{3-7}
    &&10& 0.54 {\small($\pm$0.00)}&\textbf{0.57} {\small($\pm$0.00)}& \textbf{0.69} {\small($\pm$0.00)} & 7.11 {\small($\pm$0.09)}\\
    \cline{3-7}
    &&100& 0.54 {\small($\pm$0.00)}&\textbf{0.57} {\small($\pm$0.00)}& \textbf{0.69} {\small($\pm$0.00)} & 7.28 {\small($\pm$0.11)}\\
    \cline{3-7}
    &&1000& 0.54 {\small($\pm$0.00)}&\textbf{0.57} {\small($\pm$0.00)}& \textbf{0.69} {\small($\pm$0.00)} & 7.19 {\small($\pm$0.14)}\\
    \cline{1-7}
    \multirow{16}{*}{\parbox{2cm}{MRI+\\UNETR}} & \multirow{8}{*}{\parbox{1cm}{TS}}&2&\textbf{0.47} {\small($\pm$0.00)} & \textbf{0.05} {\small($\pm$0.00)} & \textbf{0.78} {\small($\pm$0.00)} & 60.57 {\small($\pm$0.31)}\\
    \cline{3-7}
    &&3&0.44 {\small($\pm$0.00)} & 0.04 {\small($\pm$0.00)} & 0.86 {\small($\pm$0.00)} & 60.87 {\small($\pm$0.27)}\\
    \cline{3-7}
    &&4& 0.42 {\small($\pm$0.00)} & 0.04 {\small($\pm$0.00)} & 0.88 {\small($\pm$0.00)} & 60.55 {\small($\pm$0.27)}\\
    \cline{3-7}
    &&5& 0.42 {\small($\pm$0.00)} & 0.03 {\small($\pm$0.00)} & 0.90 {\small($\pm$0.00)} & 60.56 {\small($\pm$0.16)} \\
    \cline{3-7}
    &&10& 0.42 {\small($\pm$0.00)} & 0.03 {\small($\pm$0.00)} & 0.89 {\small($\pm$0.00)} & 60.15 {\small($\pm$0.12)}\\
    \cline{3-7}
    &&100& 0.44 {\small($\pm$0.00)} & 0.04 {\small($\pm$0.00)} & 0.88 {\small($\pm$0.00)} & 60.79 {\small($\pm$0.75)}\\
    \cline{3-7}
    &&1000& 0.45 {\small($\pm$0.00)} & 0.04 {\small($\pm$0.00)}& 0.88 {\small($\pm$0.00)}& 60.67 {\small($\pm$0.18)}\\
    \cline{2-7}
    &\multirow{8}{*}{Energy}&1& 0.44 {\small($\pm$0.00)} & 0.03 {\small($\pm$0.00)} & 0.89 {\small($\pm$0.00)} & 35.39 {\small($\pm$1.46)} \\
    \cline{3-7}
    &&2& 0.43 {\small($\pm$0.00)} & 0.03 {\small($\pm$0.00)} & 0.90 {\small($\pm$0.00)} & \textbf{34.82} {\small($\pm$0.38)}\\
    \cline{3-7}
    &&3& 0.44 {\small($\pm$0.00)} & 0.03 {\small($\pm$0.00)} & 0.89 {\small($\pm$0.00)} & 34.94 {\small($\pm$0.23)}\\
    \cline{3-7}
    &&4& 0.46 {\small($\pm$0.00)} & 0.03 {\small($\pm$0.00)} & 0.88 {\small($\pm$0.00)} & 35.14 {\small($\pm$0.58)}\\
    \cline{3-7}
    &&5& 0.47 {\small($\pm$0.00)} & 0.04 {\small($\pm$0.00)} & 0.87 {\small($\pm$0.00)} & 35.61 {\small($\pm$0.40)}\\
    \cline{3-7}
    &&10& 0.51 {\small($\pm$0.00)} & 0.05 {\small($\pm$0.00)} & 0.85 {\small($\pm$0.00)} & 36.23 {\small($\pm$0.11)}\\
    \cline{3-7}
    &&100& 0.54 {\small($\pm$0.00)} & \textbf{0.06} {\small($\pm$0.00)} & 0.79 {\small($\pm$0.00)} & 35.92 {\small($\pm$0.63)}\\
    \cline{3-7}
    &&1000& \textbf{0.61} {\small($\pm$0.00)} & 0.03 {\small($\pm$0.00)} & \textbf{0.71} {\small($\pm$0.00)} & 35.86 {\small($\pm$0.41)}\\
    \hline
    \end{tabularx}
\end{table}

\begin{table}[h]
    \centering
    \caption{
    Temperature scaling (TS) and energy scoring (Energy) hyperparameter searches for the nnU-nets. 
    NaN signifies that the calculation did not produce a number due to computational instabilities.
    Seconds is the amount of time it took to calculate the test distances.
    The results are averages ($\pm$SD) across 5 runs.
    Arrows denote whether higher or lower is better.
    Bold denotes the best performance per method and model.
    }
    \begin{tabularx}{\textwidth}{|X|l|c|c|c|c|c|}
    \hline
    \textbf{Model} & \textbf{Method} & \textbf{T} & \textbf{AUROC} $\uparrow$ & \textbf{AUPRC} $\uparrow$ & \textbf{FPR90} $\downarrow$ & \textbf{Seconds} $\downarrow$ \\
    \hline
    \multirow{16}{*}{\parbox{2cm}{MRI+\\nnU-net}} & \multirow{8}{*}{\parbox{1cm}{TS}}&2& 0.45 {\small($\pm$0.00)} & \textbf{0.99} {\small($\pm$0.00)} & \textbf{1.00} {\small($\pm$0.00)} & 1,254.10 {\small($\pm$0.29)}\\
    \cline{3-7}
    &&3& 0.50 {\small($\pm$0.00)} & \textbf{0.99} {\small($\pm$0.00)} & \textbf{1.00} {\small($\pm$0.00)}  & 1,252.72 {\small($\pm$0.52)}\\
    \cline{3-7}
    &&4& 0.53 {\small($\pm$0.00)} & \textbf{0.99} {\small($\pm$0.00)} & \textbf{1.00} {\small($\pm$0.00)}  & 1,253.29 {\small($\pm$0.62)}\\
    \cline{3-7}
    &&5& 0.53 {\small($\pm$0.00)} & \textbf{0.99} {\small($\pm$0.00)} & \textbf{1.00} {\small($\pm$0.00)}  & 1,252.59 {\small($\pm$0.76)}\\
    \cline{3-7}
    &&10& \textbf{0.55} {\small($\pm$0.00)} & \textbf{0.99} {\small($\pm$0.00)} & \textbf{1.00} {\small($\pm$0.00)}  & 1,252.78 {\small($\pm$0.73)}\\
    \cline{3-7}
    &&100& 0.50 {\small($\pm$0.00)} & \textbf{0.99} {\small($\pm$0.00)} & \textbf{1.00} {\small($\pm$0.00)}  & 1,253.07 {\small($\pm$0.80)}\\
    \cline{3-7}
    &&1000& 0.50 {\small($\pm$0.00)} & \textbf{0.99} {\small($\pm$0.00)} & \textbf{1.00} {\small($\pm$0.00)} & \textbf{1,204.30} {\small($\pm$0.48)}\\
    \cline{2-7}
    &\multirow{8}{*}{Energy}&1& 0.58 {\small($\pm$0.00)} & \textbf{1.00} {\small($\pm$0.00)} & \textbf{1.00} {\small($\pm$0.00)} & 186.32 {\small($\pm$0.49)}\\
    \cline{3-7}
    &&2& 0.58 {\small($\pm$0.00)} & \textbf{1.00} {\small($\pm$0.00)} & \textbf{1.00} {\small($\pm$0.00)} & \textbf{185.73} {\small($\pm$0.76)} \\
    \cline{3-7}
    &&3& 0.59 {\small($\pm$0.00)} & \textbf{1.00} {\small($\pm$0.00)} & \textbf{1.00} {\small($\pm$0.00)} & 186.50 {\small($\pm$0.49)}\\
    \cline{3-7}
    &&4& 0.59 {\small($\pm$0.00)} & \textbf{1.00} {\small($\pm$0.00)} & \textbf{1.00} {\small($\pm$0.00)} & 186.55 {\small($\pm$0.76)}\\
    \cline{3-7}
    &&5& 0.60 {\small($\pm$0.00)} & \textbf{1.00} {\small($\pm$0.00)} & \textbf{1.00} {\small($\pm$0.00)} & 186.62 {\small($\pm$0.19)}\\
    \cline{3-7}
    &&10& \textbf{0.61} {\small($\pm$0.00)} & \textbf{1.00} {\small($\pm$0.00)} & \textbf{1.00} {\small($\pm$0.00)} & 186.88 {\small($\pm$0.47)}\\
    \cline{3-7}
    &&100& NaN & NaN & NaN & NaN\\
    \cline{3-7}
    &&1000& NaN & NaN & NaN & NaN \\
    \hline
    \multirow{16}{*}{\parbox{2cm}{CT\\nnU-net}} & \multirow{8}{*}{\parbox{1cm}{TS}}&2& \textbf{0.68} {\small($\pm$0.00)} & 0.25 {\small($\pm$0.00)}& 0.66 {\small($\pm$0.00)}& 699.65 {\small($\pm$0.83)} \\
    \cline{3-7}
    &&3& \textbf{0.68} {\small($\pm$0.00)}& \textbf{0.26} {\small($\pm$0.00)}& 0.69 {\small($\pm$0.00)}& 699.07 {\small($\pm$0.37)}\\
    \cline{3-7}
    &&4& \textbf{0.68} {\small($\pm$0.00)}& 0.25 {\small($\pm$0.00)}& 0.69 {\small($\pm$0.00)}& 699.38 {\small($\pm$0.60)}\\
    \cline{3-7}
    &&5& 0.67 {\small($\pm$0.00)}& 0.25 {\small($\pm$0.00)}& 0.69 {\small($\pm$0.00)}& 699.45 {\small($\pm$0.39)} \\
    \cline{3-7}
    &&10& 0.67 {\small($\pm$0.00)}& 0.24 {\small($\pm$0.00)}& 0.70 {\small($\pm$0.00)}& 699.68 {\small($\pm$0.42)}\\
    \cline{3-7}
    &&100& 0.67 {\small($\pm$0.00)}& 0.15 {\small($\pm$0.00)}& \textbf{0.36} {\small($\pm$0.00)}& 700.37 {\small($\pm$0.41)}\\
    \cline{3-7}
    &&1000& 0.57 {\small($\pm$0.00)} & 0.10 {\small($\pm$0.00)} & 1.00 {\small($\pm$0.00)} & \textbf{670.35} {\small($\pm$0.69)}\\
    \cline{2-7}
    &\multirow{8}{*}{Energy}&1& 0.66 {\small($\pm$0.00)} & \textbf{0.26} {\small($\pm$0.00)} & \textbf{0.72} {\small($\pm$0.00)} & 105.88 {\small($\pm$0.90)} \\
    \cline{3-7}
    &&2& \textbf{0.67} {\small($\pm$0.00)} & \textbf{0.26} {\small($\pm$0.00)} & 0.75 {\small($\pm$0.00)} & \textbf{105.66} {\small($\pm$0.60)} \\
    \cline{3-7}
    &&3& \textbf{0.67} {\small($\pm$0.00)} & 0.25 {\small($\pm$0.00)} & 0.75 {\small($\pm$0.00)} & 106.26 {\small($\pm$0.52)} \\
    \cline{3-7}
    &&4& 0.66 {\small($\pm$0.00)} & 0.24 {\small($\pm$0.00)} & 0.75 {\small($\pm$0.00)} & 106.06 {\small($\pm$0.68)} \\
    \cline{3-7}
    &&5& 0.66 {\small($\pm$0.00)} & 0.23 {\small($\pm$0.00)} & 0.75 {\small($\pm$0.00)} & 106.15 {\small($\pm$0.61)} \\
    \cline{3-7}
    &&10& 0.66 {\small($\pm$0.00)} & 0.17 {\small($\pm$0.00)} & 0.76 {\small($\pm$0.00)} & 106.12 {\small($\pm$0.56)} \\
    \cline{3-7}
    &&100& NaN & NaN & NaN & NaN \\
    \cline{3-7}
    &&1000& 0.40 {\small($\pm$0.00)} & 0.08 {\small($\pm$0.00)} & 1.00 {\small($\pm$0.00)} & 105.69 {\small($\pm$0.62)} \\
    \hline
    \end{tabularx}
\end{table}

\clearpage

\section{Additional Figures}
\label{sec:figs}

\begin{figure}[h]
    \centering
    \includegraphics[width=\textwidth]{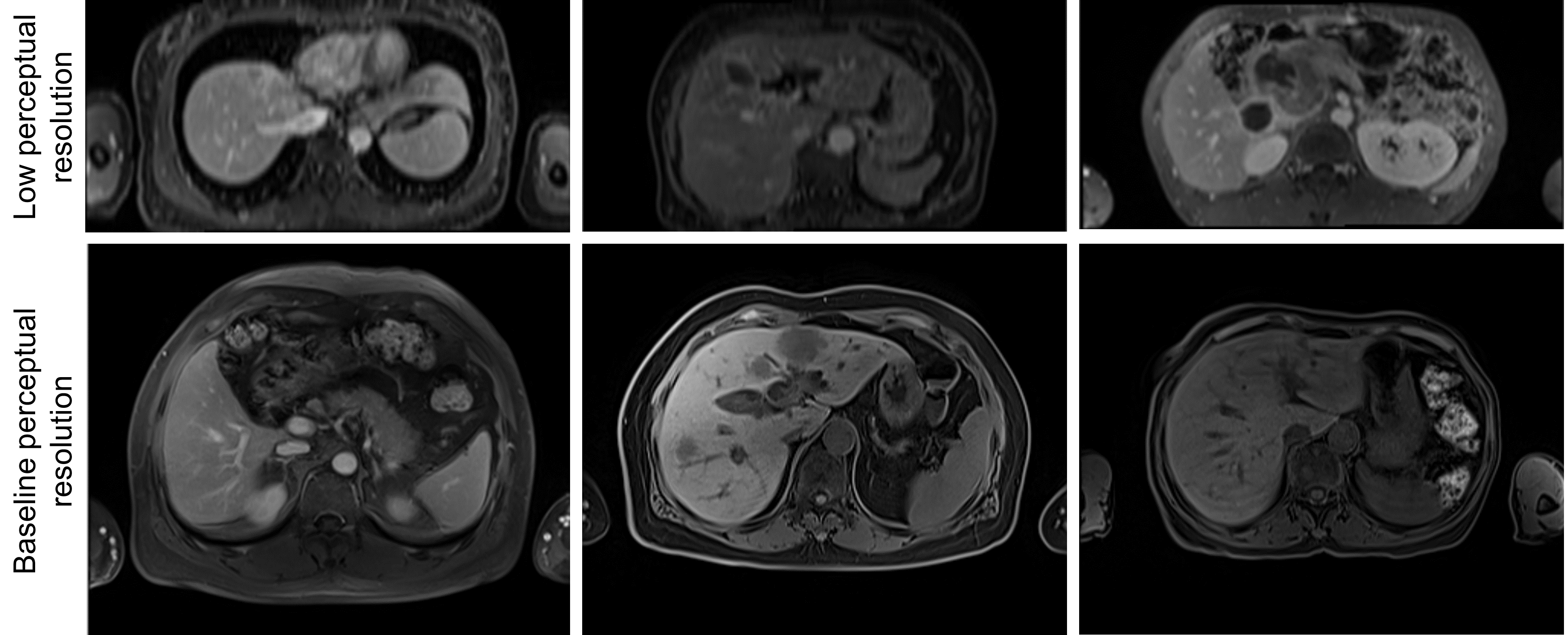}
    \caption{Sample images from the AMOS dataset \citep{Ji2022data} that represent the baseline and low perceptual resolution images that were clustered separately by the dimensionality reduction techniques.}
    \label{fig:AMOS}
\end{figure}

\begin{figure}[h]
    \centering
    \includegraphics[width=\textwidth]{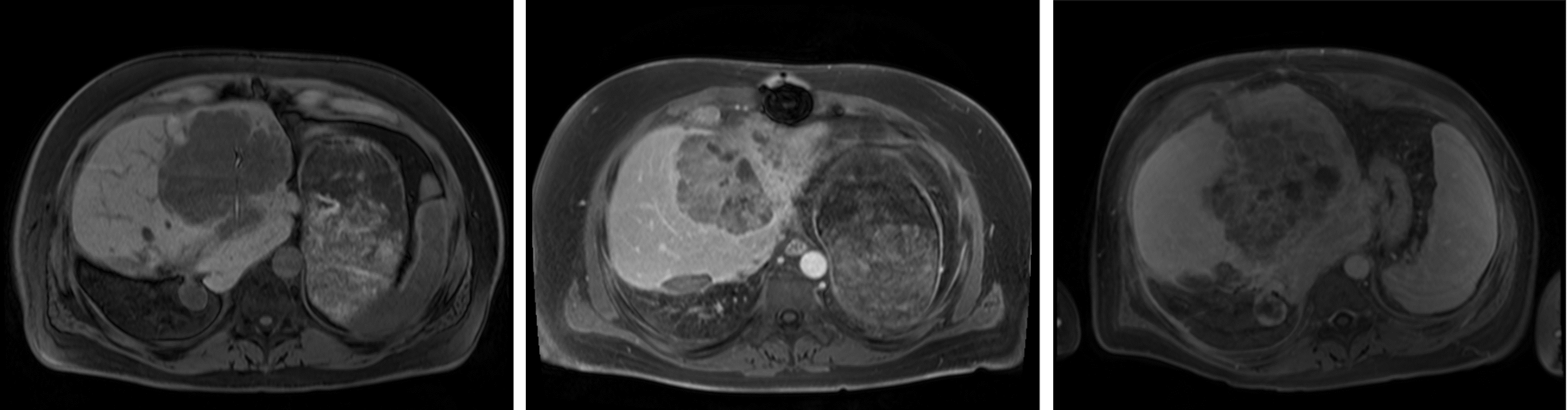}
    \caption{Sample slices from different scans from the same patient with a large tumor.}
    \label{fig:tumor}
\end{figure}

\end{document}